\documentclass[runningheads]{llncs}

\usepackage{eccv}

\usepackage{eccvabbrv}

\usepackage{graphicx}
\usepackage{booktabs}

\usepackage[accsupp]{axessibility}  %

\usepackage{tabularx} %
\usepackage{booktabs} %
\usepackage{makecell} %
\usepackage{multirow} %
\usepackage{colortbl}
\usepackage{arydshln}
\usepackage{graphicx}
\usepackage{pifont} %
\usepackage{xcolor} %
\usepackage{gensymb}
\usepackage{enumitem} %

\usepackage[capbesideposition=outside,capbesidesep=quad]{floatrow}
\newcommand{\ours}{Gaussian-SLAM\xspace}
\newcommand{\boldparagraph}[1]{\vspace{2pt}\noindent{\bf #1}}

\colorlet{colorFst}{Green!25}       %
\colorlet{colorSnd}{SpringGreen!45} %
\colorlet{colorTrd}{Yellow!30}      %
\colorlet{colorLow}{darkgray!30}    %
\newcommand{\fs}{\cellcolor{colorFst}\bf}   %
\newcommand{\nd}{\cellcolor{colorSnd}}      %
\newcommand{\rd}{\cellcolor{colorTrd}}      %

\newcommand{\greencheck}{{\color{PineGreen}\checkmark}}
\newcommand{\redx}{{\color{red}\ding{55}}}

\usepackage[pagebackref,breaklinks,colorlinks,citecolor=eccvblue]{hyperref}

\usepackage{orcidlink}
\usepackage[export]{adjustbox}
\newcolumntype{Y}{>{\centering\arraybackslash}X}

\begin{document}

\title{Gaussian-SLAM: Photo-realistic Dense SLAM with Gaussian Splatting} 

\titlerunning{Gaussian-SLAM: Photo-realistic Dense SLAM}

\author{Vladimir Yugay\inst{1} \and
Yue Li \inst{1} \and
Theo Gevers\inst{1} \and
Martin R. Oswald\inst{1}}

\authorrunning{V.~Yugay et al.}

\institute{University of Amsterdam, Netherlands \\
\url{https://vladimiryugay.github.io/gaussian_slam} \\}

\maketitle

\begin{abstract}
We present a dense simultaneous localization and mapping (SLAM) method that uses 3D Gaussians as a scene representation. Our approach enables interactive-time reconstruction and photo-realistic rendering from real-world single-camera RGBD videos. To this end, we propose a novel effective strategy for seeding new Gaussians for newly explored areas and their effective online optimization that is independent of the scene size and thus scalable to larger scenes. This is achieved by organizing the scene into sub-maps which are independently optimized and do not need to be kept in memory. We further accomplish frame-to-model camera tracking by minimizing photometric and geometric losses between the input and rendered frames. The Gaussian representation allows for high-quality photo-realistic real-time rendering of real-world scenes. Evaluation on synthetic and real-world datasets demonstrates competitive or superior performance in mapping, tracking, and rendering compared to existing neural dense SLAM methods.

\end{abstract}

\section{Introduction}
Simultaneous localization and mapping (SLAM) has been an active research topic for the past two decades~\cite{fuentes2015visual,kazerouni2022survey}. 
A major byproduct of that journey is the investigation of various scene representations to either push the tracking performance and mapping capabilities or to adapt it for more complex downstream tasks like path planning or semantic understanding.
Specifically, earlier works focus on tracking using various scene representations like feature point clouds~\cite{klein2007parallel,davison2007monoslam,Mur-Artal2017ORB-SLAM2:Cameras}, surfels~\cite{whelan2015elasticfusion,schops2019bad}, depth maps~\cite{stuhmer2010real,newcombe2011dtam}, or implicit representations~\cite{newcombe2011kinectfusion,niessner2013voxel_hashing,dai2017bundlefusion}.
Later works focus more on the map quality and density.
With the advent of powerful neural scene representations like neural radiance fields~\cite{Mildenhall2020NeRF:Synthesis} that allow for high fidelity view-synthesis, a rapidly growing body of dense neural SLAM methods~\cite{Sucar2021IMAP:Real-Time,huang2021di,zhu2022nice,mahdi2022eslam,tang2023mips,wang2023co,sandstrom2023point,zhang2023go} has been developed.
Despite their impressive gains in scene representation quality, these methods are still limited to small synthetic scenes and their re-rendering results are far from photo-realistic.

Recently, a novel scene representation based on Gaussian splatting~\cite{kerbl3Dgaussians} has been shown to deliver on-par rendering performance with NeRFs while being an order of magnitude faster in rendering and optimization. Moreover, this scene representation is directly interpretable and can be directly manipulated which is desirable for many downstream tasks.
With these advantages, the Gaussian splatting representation lends itself to be applied in an online SLAM system with real-time demands and opens the door to photo-realistic dense SLAM.

\begin{figure}
\newcommand{\imwidth}{0.4\textwidth}
\newcommand{\imheight}{3.5cm}
\centering
    \raisebox{0.5\height}{\makebox[0.01\textwidth]{\rotatebox{90}{\makecell{\scriptsize ESLAM~\cite{mahdi2022eslam}}}}}
    \includegraphics[width=\imwidth,height=\imheight]{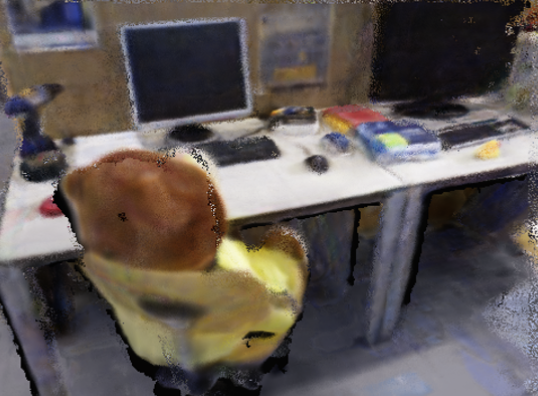}
    \includegraphics[width=\imwidth,height=\imheight]{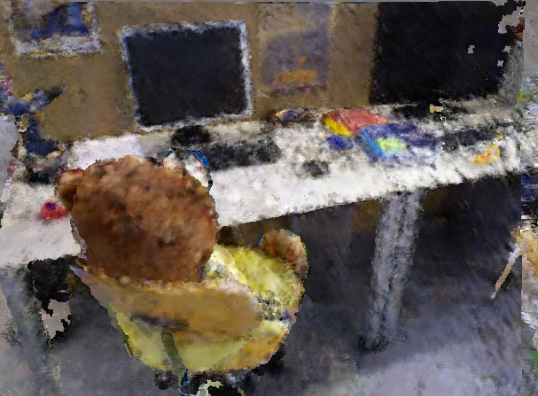} 
    \raisebox{0.25\height}{\makebox[0.01\textwidth]{\rotatebox{90}{\makecell{\scriptsize Point-SLAM~\cite{sandstrom2023point}}}}}
    \\
    \raisebox{0.1\height}{\makebox[0.01\textwidth]{\rotatebox{90}{\makecell{\scriptsize \ours (Ours)}}}}    
    \includegraphics[width=\imwidth,height=\imheight]{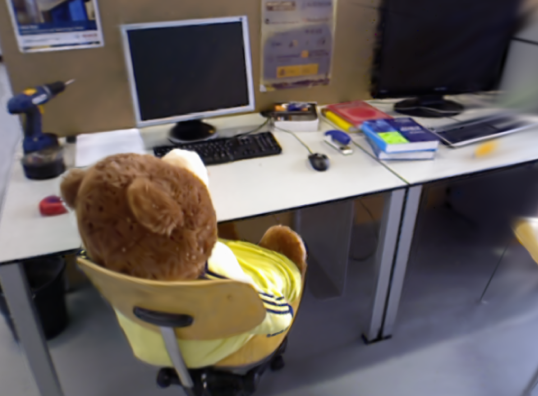}
    \includegraphics[width=\imwidth,height=\imheight]{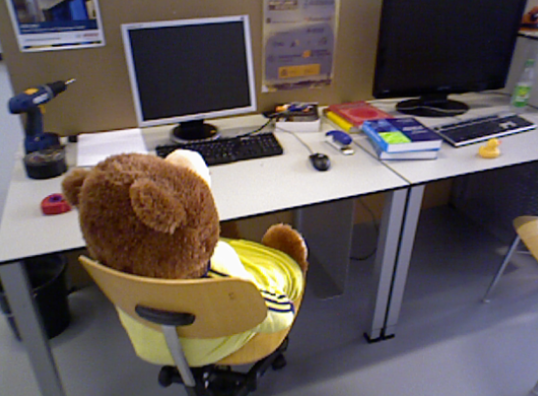} 
    \raisebox{0.5\height}{\makebox[0.01\textwidth]{\rotatebox{90}{\makecell{\scriptsize Ground Truth}}}}    
\caption{\textbf{Rendering Results of \ours.} Embedded into a dense SLAM pipeline, the 3D Gaussian-based scene representation allows for fast, photo-realistic rendering of scene views. This leads to high-quality rendering, especially on real-world data like this TUM-RGBD~\cite{Sturm2012ASystems} frame that contains many high-frequency details that other methods struggle to capture.}
\label{fig:teaser}
\end{figure}

In this paper, we introduce \ours, a dense RGBD SLAM system using 3D Gaussians to build a scene representation that allows for mapping, tracking, and photo-realistic re-rendering at interactive runtimes.
An example of the high-fidelity rendering output of \ours is depicted in Fig.~\ref{fig:teaser}.
In summary, our \textbf{contributions} include:
\begin{itemize}[itemsep=0pt,topsep=2pt,leftmargin=10pt,label=$\bullet$]
  \item A dense RGBD SLAM approach that uses 3D Gaussians to construct a scene representation allowing SOTA rendering results on real-world scenes.
  \item An extension of Gaussian splatting that better encodes geometry and allows reconstruction beyond radiance fields in a single-camera setup.
  \item An online optimization method for Gaussian splats that processes the map as sub-maps and introduces efficient seeding and optimization strategies.
  \item A frame-to-model tracker with the Gaussian splatting scene representation via photometric and geometric error minimization.
\end{itemize}

\noindent
All source code and data will be made publicly available.

\section{Related Work} \label{sec:rel}
\boldparagraph{Dense Visual SLAM and Online Mapping.}
The seminal work of Curless and Levoy~\cite{curless1996volumetric} set the stage for a variety of 3D reconstruction methods using truncated signed distance functions (TSDF). 
A line of works was built upon it improving speed~\cite{newcombe2011kinectfusion} through efficient implementation and volume integration, scalability through voxel hashing~\cite{niessner2013voxel_hashing,kahler2015hierarchical,Oleynikova2017voxblox} and octree data structure~\cite{steinbrucker2013large}, as well as tracking with sparse image features~\cite{BylowOlssonKahl} and loop closures~\cite{newcombe2011dtam,schops2019bad,cao2018real,dai2017bundlefusion}. 
Tackling the problem of unreliable depth maps, RoutedFusion~\cite{Weder2020RoutedFusion} introduced a learning-based fusion network for updating the TSDF in volumetric grids. This concept was further evolved by NeuralFusion~\cite{weder2021neuralfusion} and DI-Fusion~\cite{huang2021di}, which adopt implicit learning for scene representation, enhancing their robustness against outliers. 
Recent research has successfully achieved dense online reconstruction using solely RGB cameras~\cite{murez2020atlas,choe2021volumefusion,bovzivc2021transformerfusion,stier2021vortx,sun2021neuralrecon,sayed2022simplerecon,li2023dense}, bypassing the need for depth data.

Recently, test-time optimization methods have become popular due to their ability to adapt to unseen scenes on the fly. Continuous Neural Mapping~\cite{yan2021continual}, for instance, employs a continual mapping strategy from a series of depth maps to learn scene representation. Inspired by Neural Radiance Fields~\cite{Mildenhall2020NeRF:Synthesis}, there has been immense progress in dense surface reconstruction~\cite{Oechsle2021UNISURF:Reconstruction,wang2022neuris} and accurate pose estimation~\cite{Rosinol2022NeRF-SLAM:Fields,Lin2021BARF:Fields,wang2021nerf,bian2022nope}.
These efforts have led to the development of comprehensive dense SLAM systems~\cite{yang2022vox,zhu2022nice,zhu2023nicer,Sucar2021IMAP:Real-Time,mahdi2022eslam,zhang2023go,sandstrom2023uncle}, showing a trend in the pursuit of precise and reliable visual SLAM. A comprehensive survey on online RGBD reconstruction can be found in~\cite{zollhofer2018state}. 

While the latest neural methods show impressive rendering capabilities on synthetic data, they struggle when applied to real-world data. Further, these methods are not yet practical for real-world applications due to computation requirements, slow speed, and the inability to effectively incorporate pose updates, as the neural representations rely on positional encoding. In contrast, our method shows impressive performance on real-world data, has competitive tracking and runtime, and uses a scene representation that naturally allows pose updates.

\boldparagraph{Scene Representations for SLAM.}
The majority of dense 3D scene representations for SLAM are grid-based, point-based, network-based, or hybrid. Among these, grid-based techniques are perhaps the most extensively researched. 
They further divide into methods using dense grids~\cite{zhu2022nice,newcombe2011kinectfusion,weder2021neuralfusion,Weder2020RoutedFusion,curless1996volumetric,sun2021neuralrecon,bovzivc2021transformerfusion,li2022bnv,choi2015robust,whelan2015elasticfusion,zhou2013dense,zhou2013elastic,whelan2012kintinuous}, hierarchical octrees~\cite{yang2022vox,steinbrucker2013large,marniok2017efficient,chen2013scalable,liu2023self, Liu2020NeuralSparseVoxelFields} and voxel hashing~\cite{niessner2013voxel_hashing,kahler2015hierarchical,dai2017bundlefusion,wang2022neuris,muller2022instant} for efficient memory management. Grids offer the advantage of simple and quick neighborhood lookups and context integration. However, a key limitation is the need to predefine grid resolution, which is not easily adjustable during reconstruction. This can result in inefficient memory usage in empty areas while failing to capture finer details due to resolution constraints. 

Point-based approaches address some of the grid-related challenges and have been effectively utilized in 3D reconstruction~\cite{whelan2015elasticfusion,schops2019bad,cao2018real,chung2022orbeez,Kahler2015infiniTAM,keller2013real,cho2021sp,zhang2020dense}. Unlike grid resolution, the density of points in these methods does not have to be predetermined and can naturally vary throughout the scene. 
Moreover, point sets can be efficiently concentrated around surfaces, not spending memory on modeling empty space. The trade-off for this adaptability is the complexity of finding neighboring points, as point sets lack structured connectivity. 
In dense SLAM, this challenge can be mitigated by transforming the 3D neighborhood search into a 2D problem via projection onto keyframes~\cite{whelan2015elasticfusion,schops2019bad}, or by organizing points within a grid structure for expedited searching~\cite{xu2022point}.

Network-based methods for dense 3D reconstruction provide a continuous scene representation by implicitly modeling it with coordinate-based networks~\cite{azinovic2022neural,Sucar2021IMAP:Real-Time,wang2022neuris,ortiz2022isdf,yan2021continual,yang2022fd, mescheder2019occupancy,li2023dense,zhang2023go,wang2023co}. 
This representation can capture high-quality maps and textures. However, they are generally unsuitable for online scene reconstruction due to their inability to update local scene regions and to scale for larger scenes. More recently, a hybrid representation combining the advantages of point-based and neural-based was proposed \cite{sandstrom2023point}. 
While addressing some of the issues of both representations it struggles with real-world scenes, and cannot seamlessly integrate trajectory updates in the scene representation.

Outside these three primary categories, some studies have explored alternative representations like surfels~\cite{mihajlovic2021deepsurfels,gao2023surfelnerf} and neural planes~\cite{mahdi2022eslam,peng2020convolutional}. Parameterized surface elements are generally
not great at modeling a flexible shape template while
feature planes struggle with scene reconstructions containing multiple surfaces, due to their overly compressed representation.
Recently, Kerbl~\etal~\cite{kerbl3Dgaussians} proposed to represent a scene with 3D Gaussians. 
The Gaussian parameters are optimized via differential rendering with multi-view supervision. 
While being very efficient and achieving impressive rendering results, this representation is tailored for fully-observed multi-view environments and does not encode geometry well. Concurrent to our work, \cite{yang2023deformable,wu20234d,yang2023realtime} focus on dynamic scene reconstruction, and ~\cite{luiten2023dynamic} on tracking. However, they are all offline methods and do not suit single-camera dense SLAM setups.

Concurrently, several methods \cite{keetha2023splatam, matsuki2023gaussian, yan2024gsslam, huang2023photoslam} have used Gaussian Splatting \cite{kerbl3Dgaussians} for SLAM. While most splatting-based methods use gradient-based map densification similar to \cite{kerbl3Dgaussians}, we follow a more controlled approach with exact thresholding by utilizing fast nearest-neighbor search and alpha masking. Further, unlike all other concurrent work, our mapping pipeline does not require holding all the 3D Gaussians in the GPU memory, allowing our method to scale and not slow down as more areas are covered. Moreover, while in concurrent works the 3D Gaussians are very densely seeded, our color gradient and masking-based seeding strategy allows for sparser seeding while preserving SOTA rendering quality. Finally, in contrast to \cite{matsuki2023gaussian, yan2024gsslam, huang2023photoslam}, our tracking does not rely on explicitly computed camera pose derivatives and is implemented in PyTorch.

\section{Method} \label{sec:methods}
\begin{figure*}[t]
  \centering
  \includegraphics[width=\textwidth]{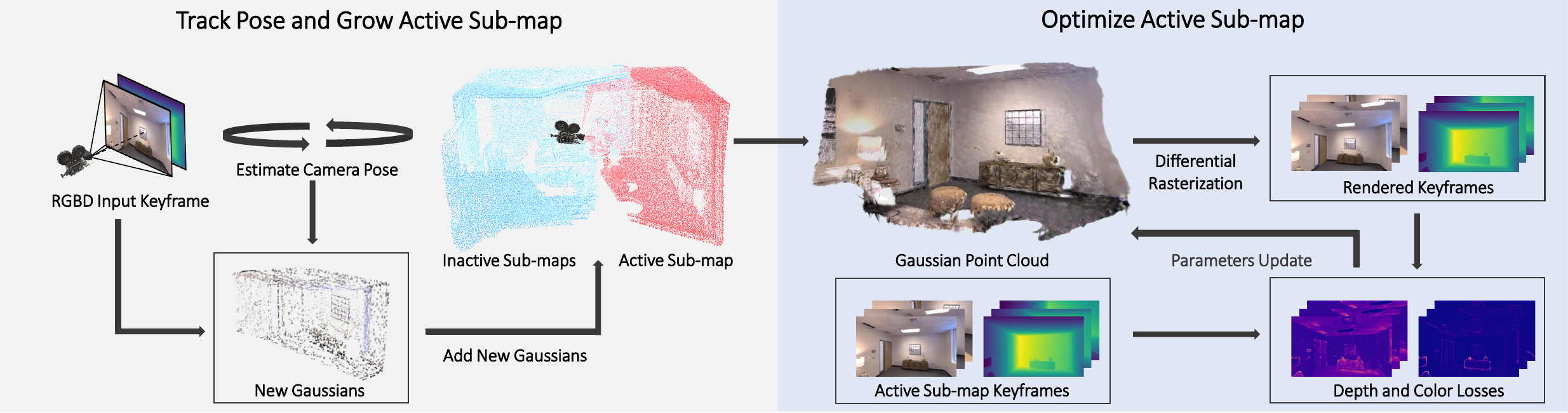}\\
  \caption{\textbf{\ours{} Architecture.} For every input keyframe the camera pose is estimated using depth and color losses against the \textit{active} sub-map. Given an estimated camera pose, the RGBD frame is transformed into 3D and subsampled based on color gradient and the rendered alpha mask. Points from the subsampled point clouds located in low-density areas of the \textit{active} sub-map are used to initialize new 3D Gaussians. These sparse 3D Gaussians are then added to the Gaussian point cloud of the \textit{active} sub-map and are jointly optimized with the depth maps and color images from all contributing keyframes of this sub-map.}
  \label{fig:architecture}
\end{figure*}

The key idea of our approach is to construct a map using 3D Gaussians\cite{kerbl3Dgaussians} as a main building block to make single-camera RGBD neural SLAM scalable, faster and achieve better rendering on real-world datasets. We introduce a novel efficient mapping process with bounded computational cost in a sequential single-camera setup, a challenging scenario for traditional 3D Gaussian Splatting. To enable traditional Gaussian splats to render accurate geometry we extend them by adding a differential depth rendering, explicitly computing gradients for the Gaussian parameters updates. Finally, we develop a novel frame-to-model tracking approach relying on our 3D map representation. \cref{fig:architecture} provides an overview of our method. We now explain our pipeline, starting with an overview of classical Gaussian splatting \cite{kerbl3Dgaussians}, and continuing with map construction and optimization, geometry encoding, and tracking.

\subsection{Gaussian Splatting}
Gaussian splatting~\cite{kerbl3Dgaussians} is an effective method for representing 3D scenes with novel-view synthesis capability.
This approach is notable for its speed, without compromising the rendering quality. In \cite{kerbl3Dgaussians}, 3D Gaussians are initialized from a sparse Structure-from-Motion point cloud of a scene. With images observing the scene from different angles, the Gaussian parameters are optimized using differentiable rendering. During training, 3D Gaussians are adaptively added or removed to better render the images based on a set of heuristics.

A single 3D Gaussian is parameterized by mean $\mu \in \mathbb{R}^3$, covariance $\Sigma \in \mathbb{R}^{3 \times 3}$, opacity $o \in \mathbb{R}$, and RGB color $C \in \mathbb{R}^3$. The mean of a projected (splatted) 3D Gaussian in the 2D image plane $\mu^{I}$ is computed as follows:
\begin{equation}
    \mu^{I} = \pi\big(P(T_{wc} \mu_\text{homogeneous} )\big) \enspace,
\end{equation}
where $T_{wc} \in SE(3)$ is the world-to-camera transformation, $P \in \mathbb{R}^{4 \times 4}$ is an OpenGL-style projection matrix, $\pi: \mathbb{R}^{4} \rightarrow \mathbb{R}^{2}$ is a projection to pixel coordinates. 
The 2D covariance $\Sigma^{I}$ of a splatted Gaussian is computed as:
\begin{equation}
    \Sigma^{I} = J R_{wc} \Sigma R_{wc}^T J^T \enspace,
\end{equation}
where $J \in \mathbb{R}^{2 \times 3}$ is an affine transformation from \cite{zwicker2001surface}, $R_{wc} \in SO(3)$ is the rotation component of world-to-camera transformation $T_{wc}$. 
We refer to \cite{zwicker2001surface} for further details about the projection matrices. Color $C$ along one channel $ch$ at a pixel $i$ influenced by $m$ ordered Gaussians is rendered as:
\begin{align}
    C^{ch}_{i} = \sum_{j \leq m}C^{ch}_{j} \cdot \alpha_{j} \cdot T_{j} \enspace, \;\text{with }\; T_{j} = \prod_{k < j}(1 - \alpha_{k}) \enspace,
\label{eq:color_blending}
\end{align}
with $\alpha_{j}$ is computed as:
\begin{equation}
 \alpha_{j} = o_{j} \cdot \exp(-\sigma_{j}) \quad\text{ and }\quad \sigma_{j} = \frac{1}{2} \Delta_{j}^T \Sigma_{j}^{I-1} \Delta_{j} \enspace,  
\end{equation}
where $\Delta_{j} \in \mathbb{R}^{2}$ is the offset between the pixel coordinates and the 2D mean of a splatted Gaussian. 
The parameters of the 3D Gaussians are iteratively optimized by minimizing the photometric loss between rendered and training images. During optimization, $C$ is encoded with spherical harmonics $SH \in \mathbb{R}^{15}$ to account for direction-based color variations. Covariance is decomposed as $\Sigma = RSS^{T}R^{T}$, where $R \in \mathbb{R}^{3 \times 3}$ and $S = \text{diag}(s) \in \mathbb{R}^{3 \times 3}$ are rotation and scale respectively to preserve covariance positive semi-definite property during gradient-based optimization.
\subsection{3D Gaussian-based Map} \label{sec:gs_map}
To avoid catastrophic forgetting and overfitting and make the mapping computationally feasible in a single-camera stream scenario we process the input in chunks (sub-maps). Every sub-map covers several keyframes observing it and is represented with a separate 3D Gaussian point cloud. 
Formally, we define a sub-map Gaussian point cloud $P^s$ as a collection of $N$ 3D Gaussians:
\begin{equation}
  P^s = \{G(\mu^{s}_{i}, \Sigma^{s}_{i}, o^{s}_{i}, C^{s}_{i}) \, | \, i=1,\ldots,N\} \enspace.
\end{equation}

\boldparagraph{Sub-map Initialization.} 
A sub-map starts with the first frame and grows incrementally with newly incoming keyframes. As the explored area grows, a new sub-map is needed to cover the unseen regions and avoid storing all the Gaussians in GPU memory.
Instead of using a fixed interval when creating a new sub-map \cite{choi2015robust, dai2017bundlefusion, maier2014submap}, an initialization strategy that relies on the camera motion~\cite{cao2018real, stuckler2014multi} is used. 
Specifically, a new sub-map is created when the current frame's estimated translation relative to the first frame of the active sub-map exceeds a predefined threshold, $d_\text{thre}$, or when the estimated Euler angle surpasses $\theta_\text{thre}$. At any time, only active sub-map is processed. This approach bounds the compute cost and ensures that optimization remains fast while exploring larger scenes.

\boldparagraph{Sub-map Building.} 
Every new keyframe may add 3D Gaussians to the active sub-map to account for the newly observed parts of the scene. Following the pose estimation for the current keyframe, a dense point cloud is computed from keyframe RGBD measurements. At the beginning of each sub-map, we sample $M_u$ uniformly and $M_c$ points from the keyframe point cloud in high color gradient regions to add new Gaussians. For the following keyframes of the sub-map, we sample $M_k$ points uniformly from the regions with the rendered alpha values lower than a threshold $\alpha_n$. This allows for growing the map in areas sparsely covered by the 3D Gaussians. New Gaussians are added to the sub-map using sampled points that have no neighbors within a search radius $\rho$ in the current sub-map. The new Gaussians are anisotropic and their scales are defined based on the nearest neighbor distance within the active sub-map. This densification strategy substantially differs from\cite{kerbl3Dgaussians} where new Gaussians were added and pruned based on the gradient values during optimization and gives fine-grained control over the number of Gaussians.

\boldparagraph{Sub-map Optimization.} All Gaussians in the active sub-map are jointly optimized every time new Gaussians are added to the sub-map for a fixed number of iterations minimizing the loss~\eqref{eq:joint_loss}. We do not clone or prune the Gaussians as done in \cite{kerbl3Dgaussians} during optimization to preserve geometry density obtained from the depth sensor, decrease computation time, and better control the number of Gaussians. We optimize the active sub-map to render the depth and color of all its keyframes. We directly optimize RGB color without using spherical harmonics to speed up optimization.
In Gaussian splatting \cite{kerbl3Dgaussians} the scene representation is optimized for many iterations over all the training views. However, this approach does not suit the SLAM setup where speed is crucial. Naively optimizing with an equal number of iterations for all keyframes results in underfitting or excessive time spent on optimization. We solve this by optimizing only the keyframes in the active sub-map and spending at least 40\% of iterations on the new keyframe.
\subsection{Geometry and Color Encoding} \label{sec:dc_rendering}
While Gaussian Splatting\cite{kerbl3Dgaussians} is good at rendering images, the rendered depth maps are of limited accuracy since there is no direct depth supervision. We tackle this problem with an additional depth loss.
To render the depth $D_i$ at pixel $i$ that is influenced by $m$ ordered Gaussians we compute:
\begin{equation}
     D_{i} = \sum_{j \leq m}\mu^{z}_{j} \cdot \alpha_{j} \cdot T_{j} \enspace,
\end{equation}
where $\mu^{z}_{j}$ is a $z$ component of the mean of a 3D Gaussian, $\alpha_j$ and $T_j$ are the same as in Eq.~\eqref{eq:color_blending}.
To update the 3D Gaussian parameters based on the observed depth, we derive the gradients of the depth loss w.r.t. the 3D Gaussians' means, covariances, and opacity. 
Denoting the depth loss as $L_\mathrm{depth}$, we follow the chain rule to compute the gradient for the mean update of the Gaussian $j$:
\begin{equation}
     \frac{\partial L_\mathrm{depth}}{\partial \mu_j} = \frac{\partial L_\mathrm{depth}}{\partial D_i} \frac{\partial D_i}{\partial \alpha_{j}} \frac{\partial \alpha_j}{\partial \mu_j} \enspace,
\end{equation}
where $\frac{\partial L_\mathrm{depth}}{\partial D_i}$ is computed with PyTorch autograd using Eq.~\eqref{eq:depth_loss} and $\frac{\partial \alpha_j}{\partial \mu_j}$ is derived as in \cite{kerbl3Dgaussians}. We derive $\frac{\partial D_i}{\partial \alpha_{j}}$ as:
\begin{equation}
    \frac{\partial D_i}{\partial \alpha_{j}} = \mu_j^z \cdot T_j - \frac{\sum_{u > j} \mu_u^z \alpha_u T_u}{1 - \alpha_j} \enspace.
\end{equation}
The gradients for covariance and opacity are computed similarly.
Apart from $\frac{\partial L_\mathrm{depth}}{\partial D_i}$, all gradients are explicitly computed in CUDA to preserve the optimization speed of the unified rendering pipeline.
For depth supervision, we use the loss:
\begin{equation}
    L_\text{depth} = |\hat{D} - D|_1 \enspace,
    \label{eq:depth_loss}
\end{equation}
with $D$ and $\hat{D}$ being the ground-truth and reconstructed depth maps, respectively.
For the color supervision we use a weighted combination of $L_{1}$ and $\mathrm{SSIM}$~\cite{wang2004image} losses:
\begin{equation}
    L_\text{color} = (1 - \lambda) \cdot |\hat{I} - I|_1 + 
       \lambda \big(1 - \mathrm{SSIM}(\hat{I}, I) \big) \enspace,
    \label{eq:color_loss}
\end{equation}
where $I$ is the original image, $\hat{I}$ is the rendered image, and $\lambda = 0.2$.

When seeded sparsely as in our case, a few 3D Gaussians sometimes elongate too much in scale. To overcome this, we add an isotropic regularization term $L_\text{reg}$ when optimizing a sub-map $K$:
\begin{equation}
    L_\text{reg} = \frac{\sum_{k \in K}|s_k - \overline{s}_k|_1}{|K|},
\end{equation}
where $s_k\in\mathbb{R}^3$ is the scale of a 3D Gaussian, $\overline{s}_k$ is the mean sub-map scale, and $|K|$ is the number of Gaussians in the sub-map.
Finally, we optimize color, depth, and regularization terms together:
\begin{equation}
    L = \lambda_\text{color} \cdot L_\text{color} + \lambda_\text{depth} \cdot L_\text{depth} + \lambda_\text{reg} \cdot L_\text{reg} \enspace,
\label{eq:joint_loss}
\end{equation}
where $\lambda_\text{color}, \lambda_\text{depth}, \lambda_\text{reg} \cdot \in\mathbb{R}_{\geq 0}$ are weights for the corresponding losses.

\subsection{Tracking} \label{method:tracking}
We perform frame-to-model tracking based on the mapped scene. 
We initialize the current camera pose $T_i$ with a constant speed assumption:
\begin{equation}
    T_{i} = T_{i - 1} + (T_{i - 1} - T_{i - 2}) \enspace,
\end{equation}
where pose $T_i = \{\mathbf{q_i}, \mathbf{t_i}\}$ encodes a quaternion and translation vector.
To estimate the camera pose we minimize the tracking loss $L_\text{tracking}$ with respect to relative camera pose $T_{i - 1, i}$ between frames $i - 1$ and $i$ as follows:
\begin{equation}
    \underset{T_{i - 1, i}}{\mathrm{arg\,min}} \, L_\text{tracking}\Big(\hat{I}(T_{i - 1, i}), \hat{D}(T_{i - 1, i}), I_{i}, D_{i}, \alpha\Big) \enspace,
\end{equation}
where $\hat{I}(T_{i - 1, i})$ and $\hat{D}(T_{i - 1, i})$ are the rendered color and depth from the sub-map transformed with the relative transformation $T_{i - 1, i}$, $C_{i}$ and $D_{i}$ are the input color and depth map at frame $i$.

We introduce soft alpha and error masking to not contaminate the tracking loss with the pixels from previously unobserved or poorly reconstructed areas. Soft alpha mask $M_\text{alpha}$ is a polynomial of the alpha map rendered directly from the active sub-map. Error boolean mask $M_\text{inlier}$ discards all the pixels where the color and depth errors are larger than a frame-relative error threshold:
\begin{equation}
    L_\text{tracking} = \sum M_\text{inlier} \cdot M_\text{alpha} \cdot(\lambda_{c}|\hat{I} - I|_1 + (1 - \lambda_{c})|\hat{D} - D|_1).
\end{equation}
The weighting ensures the optimization is guided by well-reconstructed regions where the accumulated alpha values are close to 1 and rendering quality is high. During optimization, all the 3D Gaussian parameters are frozen.

\section{Experiments}
\label{sec:exp}

We first describe our experimental setup and then evaluate our method against state-of-the-art dense neural RGBD SLAM methods on synthetic~\cite{straub2019replica} and real-world datasets~\cite{Sturm2012ASystems, Dai2017ScanNet, yeshwanth2023scannet++}. In addition, we compare our method with concurrent work with released source code. The reported results are the average of 3 runs using different seeds. The tables highlight best results as \colorbox{colorFst}{first}, \colorbox{colorSnd}{second}, \colorbox{colorTrd}{third}.

\boldparagraph{Implementation Details.}
We set $M_u=600000$ for Replica~\cite{straub2019replica}, $100000$ for TUM-RGBD~\cite{Sturm2012ASystems} and ScanNet~\cite{Dai2017ScanNet}, and $400000$ for ScanNet++~\cite{yeshwanth2023scannet++}. $M_c$ is set to $50000$ for all datasets. For the first keyframe in a sub-map, the number of mapping iterations is set to 1,000 for Replica, 100 for TUM-RGBD and ScanNet, and 500 for ScanNet++. For the subsequent keyframes in a sub-map, the iteration count is set to 100 across all datasets. Every 5th frame is considered as a keyframe for all the datasets. When selecting point candidates from subsequent keyframes, we use alpha threshold $\alpha_n=0.6$. We use FAISS~\cite{johnson2019billion} GPU implementation to find nearest neighbors when choosing point candidates to add as new Gaussians and set the search radius $\rho = 0.01\,m$ for all the datasets. For new sub-map initialization, we set $d_\text{thre}=0.5\,m$ and $\theta_\text{thre}=50\degree$. For sub-map optimization, the best results were obtained with $\lambda_\text{color}$, $\lambda_\text{reg}$ and $\lambda_\text{depth}$ to 1. We spend at least 40\% mapping iterations on the newly added keyframe during sub-map optimization. To mesh the scene, we render depth and color every fifth frame over the estimated trajectory and use TSDF Fusion~\cite{curless1996volumetric} with voxel size $1\,cm$ similar to \cite{sandstrom2023point}. Further details are provided in the supplement.

\boldparagraph{Datasets.} 
The Replica dataset~\cite{straub2019replica} comprises high-quality 3D reconstructions of a variety of indoor scenes. 
We utilize the publicly available dataset collected by Sucar~\etal~\cite{Sucar2021IMAP:Real-Time}, which provides trajectories from an RGBD sensor.
Further, we demonstrate that our framework achieves SOTA results on real-world data by using the TUM-RGBD~\cite{Sturm2012ASystems}, ScanNet~\cite{Dai2017ScanNet} and ScanNet++~\cite{yeshwanth2023scannet++} datasets. 
The poses for TUM-RGBD were captured using an external motion capture system while ScanNet uses poses estimated by BundleFusion~\cite{dai2017bundlefusion}, and ScanNet++ obtains poses by registering the images with a laser scan. Since ScanNet++ is not specifically designed for benchmarking neural SLAM, it has larger camera movements. Therefore, we choose 5 scenes where the first 250 frames are smooth in trajectory and use them for benchmarking.

\boldparagraph{Evaluation Metrics.}
To assess tracking accuracy, we use ATE RMSE~\cite{Sturm2012ASystems}, and for rendering we compute PSNR, SSIM~\cite{wang2004image} and LPIPS~\cite{zhang2018unreasonable}.
All rendering metrics are evaluated by rendering full-resolution images along the estimated trajectory with mapping intervals similar to \cite{sandstrom2023point}. We also follow \cite{sandstrom2023point} to measure reconstruction performance on meshes produced by marching cubes~\cite{lorensen1987marching}. 
The reconstructions are also evaluated using the F1-score - the harmonic mean of the Precision (P) and Recall (R). We use a distance threshold of $1$ cm for all evaluations. We further provide the depth L1 metric for unseen views as in~\cite{zhu2022nice}.

\boldparagraph{Baseline Methods.} 
We primarily compare our method to existing state-of-the-art dense neural RGBD SLAM methods such as NICE-SLAM~\cite{zhu2022nice}, Vox-Fusion~\cite{yang2022vox}, ESLAM~\cite{mahdi2022eslam}, and Point-SLAM~\cite{sandstrom2023point}. In addition, we compare against the concurrent work using the released code\cite{keetha2023splatam}.

\begin{figure*}[ht!] \centering
    \newcommand{\wratio}{0.19}
    \newcommand{\imgheight}{2.0cm}
    
    \makebox[0.01\textwidth][c]{}
    \makebox[\wratio\textwidth]{\scriptsize NICE-SLAM\cite{zhu2022nice}}
    \makebox[\wratio\textwidth]{\scriptsize ESLAM\cite{mahdi2022eslam}}
    \makebox[\wratio\textwidth]{\scriptsize Point-SLAM\cite{sandstrom2023point}}
    \makebox[\wratio\textwidth]{\scriptsize Ours}
    \makebox[\wratio\textwidth]{\scriptsize Ground-truth}
    \\
    
    \raisebox{0.5\height}{\makebox[0.01\textwidth]{\rotatebox{90}{\hspace{-0.5em}\scriptsize scene 0000}}}    
    \includegraphics[width=\wratio\textwidth,height=\imgheight]{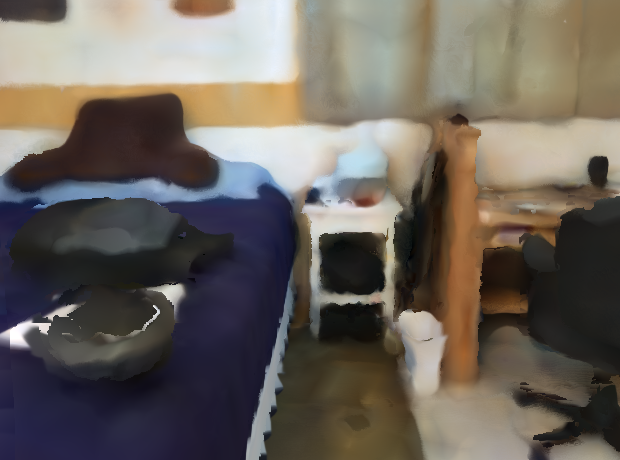}
    \includegraphics[width=\wratio\textwidth,height=\imgheight]{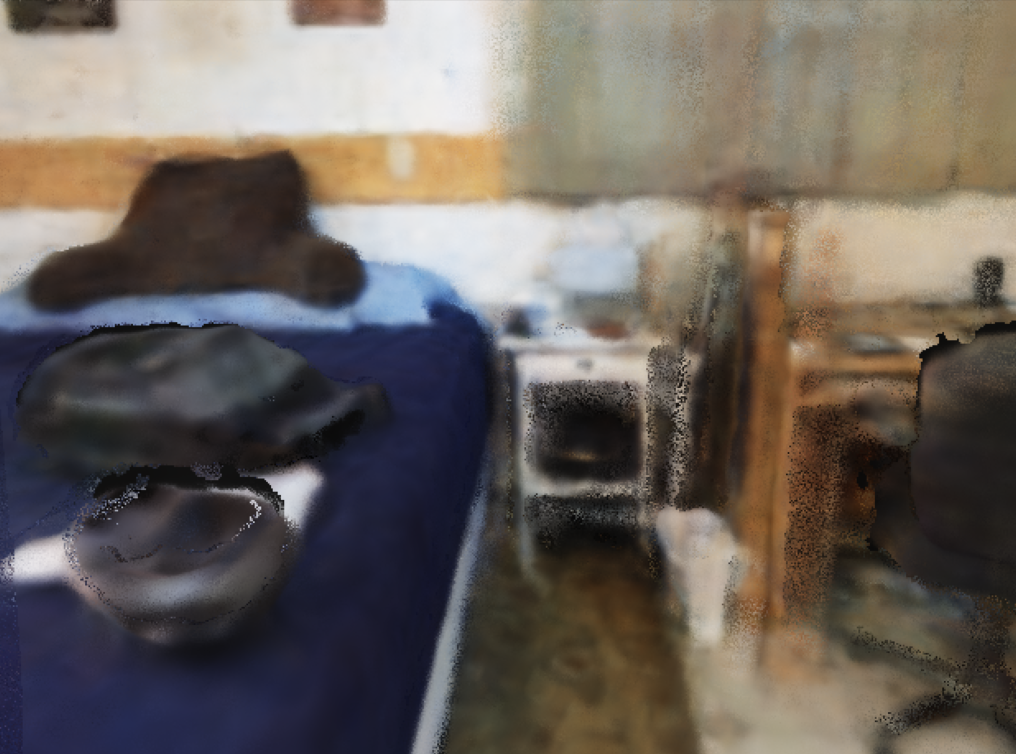}
    \includegraphics[width=\wratio\textwidth,height=\imgheight]{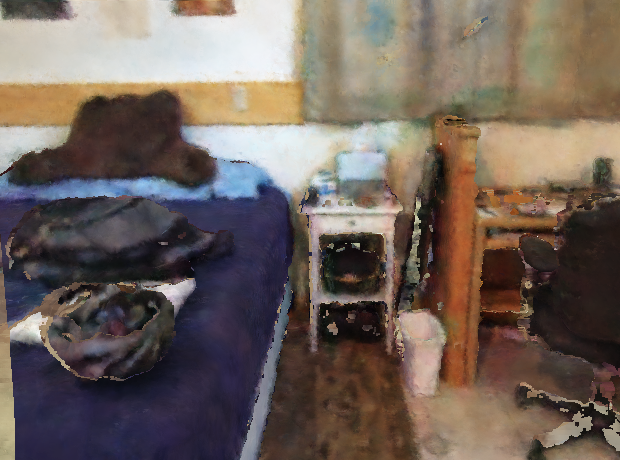}
    \includegraphics[width=\wratio\textwidth,height=\imgheight]{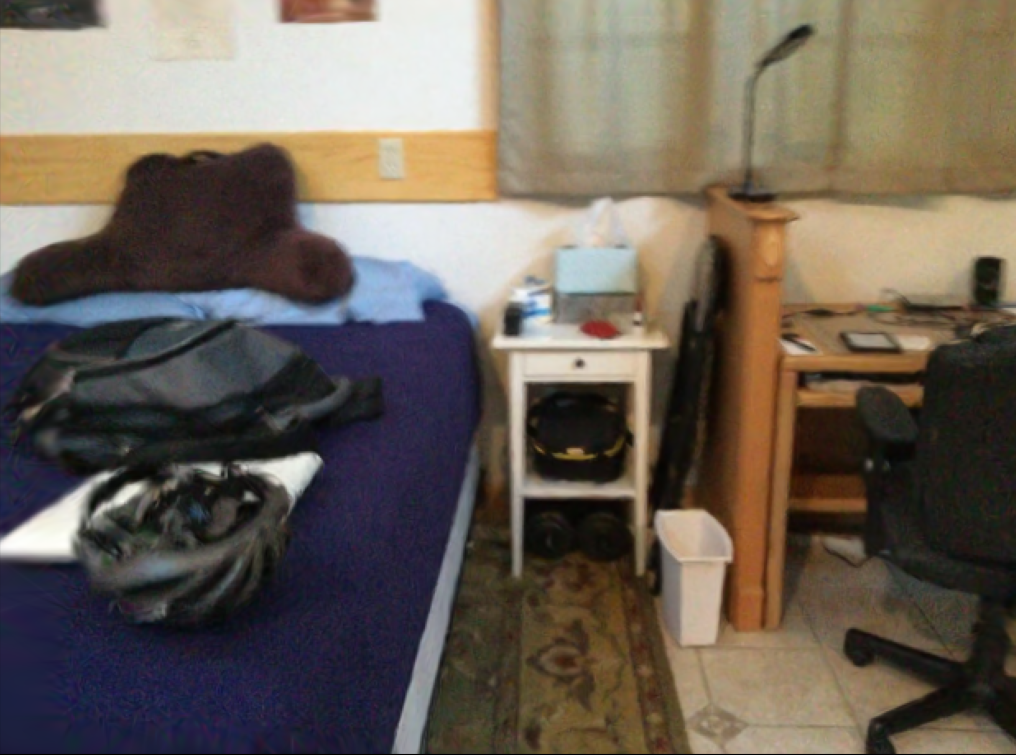}
    \includegraphics[width=\wratio\textwidth,height=\imgheight]{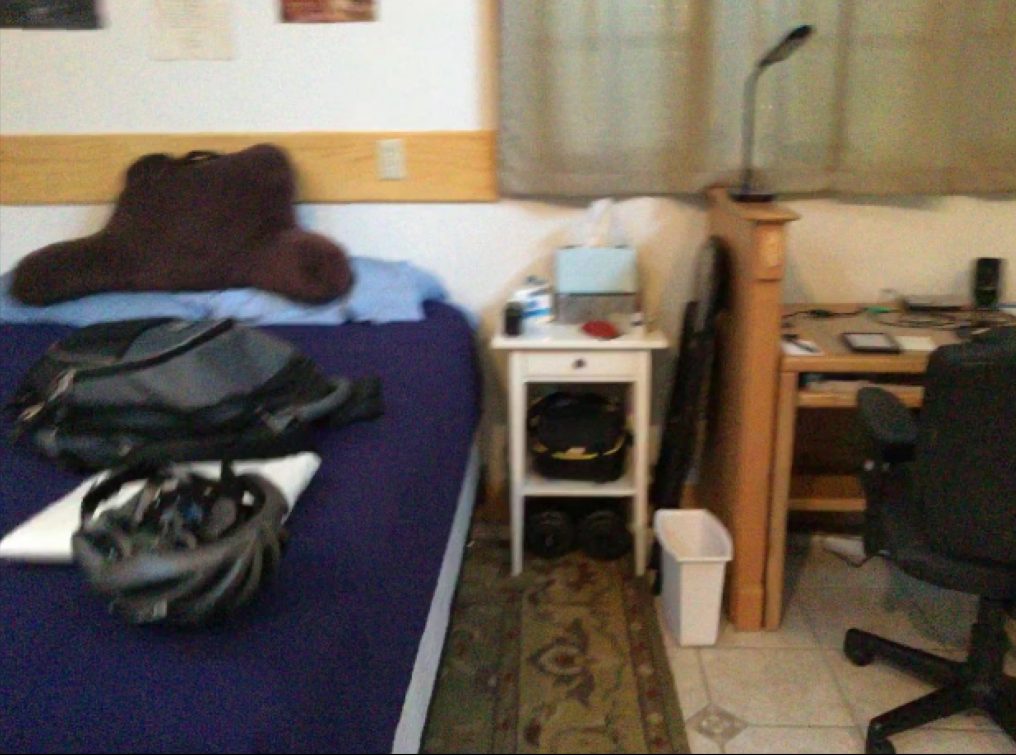}    
    \\
    
    \raisebox{0.5\height}{\makebox[0.01\textwidth]{\rotatebox{90}{\hspace{-0.5em}\scriptsize scene 0207}}}
    \includegraphics[width=\wratio\textwidth,height=\imgheight]{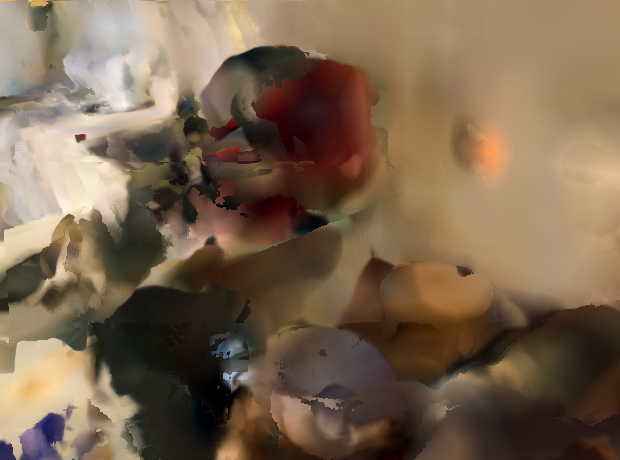}    
    \includegraphics[width=\wratio\textwidth,height=\imgheight]{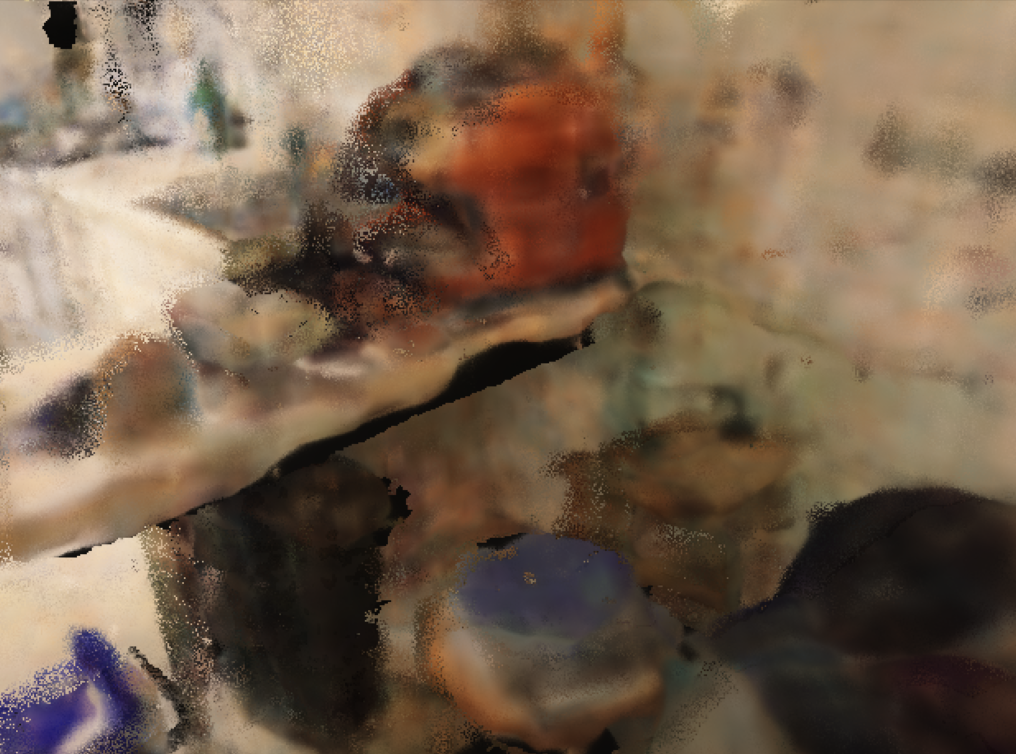}
    \includegraphics[width=\wratio\textwidth,height=\imgheight]{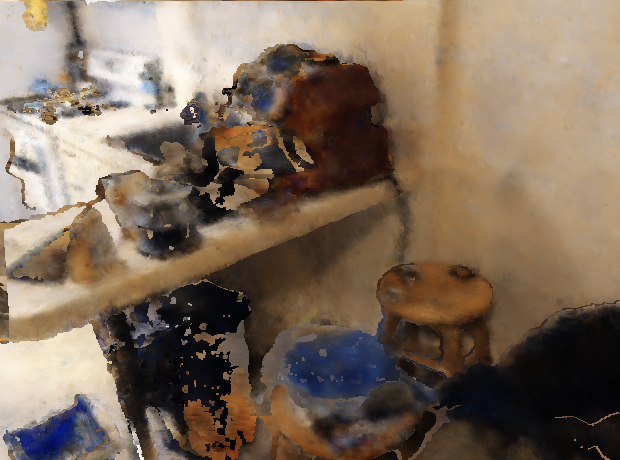}
    \includegraphics[width=\wratio\textwidth,height=\imgheight]{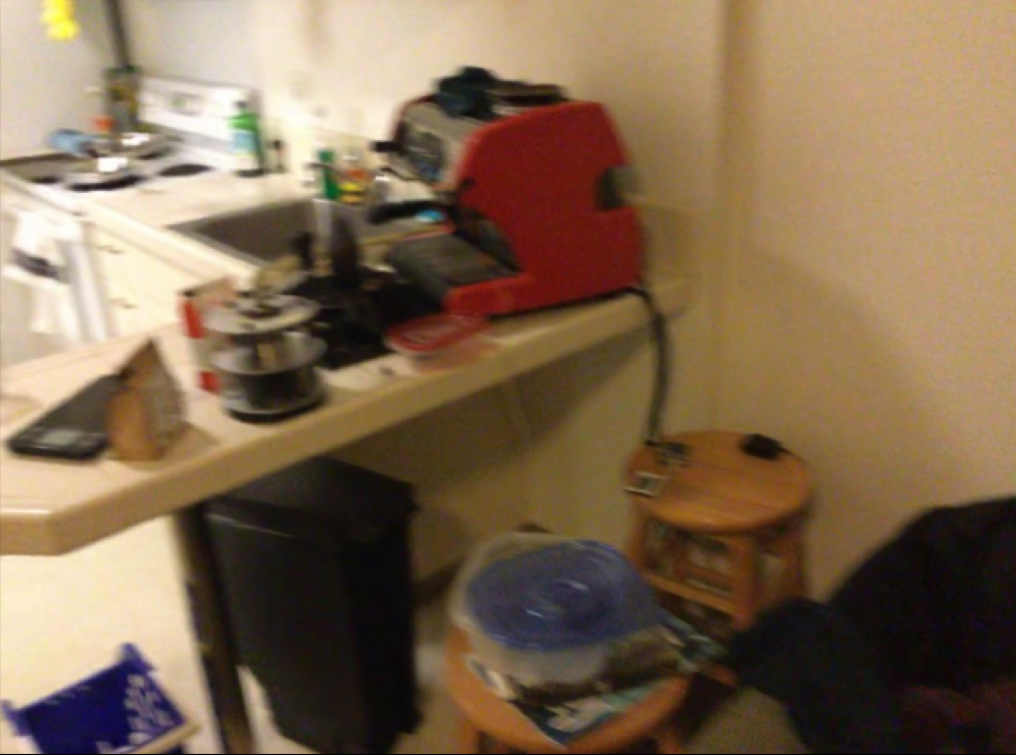}
    \includegraphics[width=\wratio\textwidth,height=\imgheight]{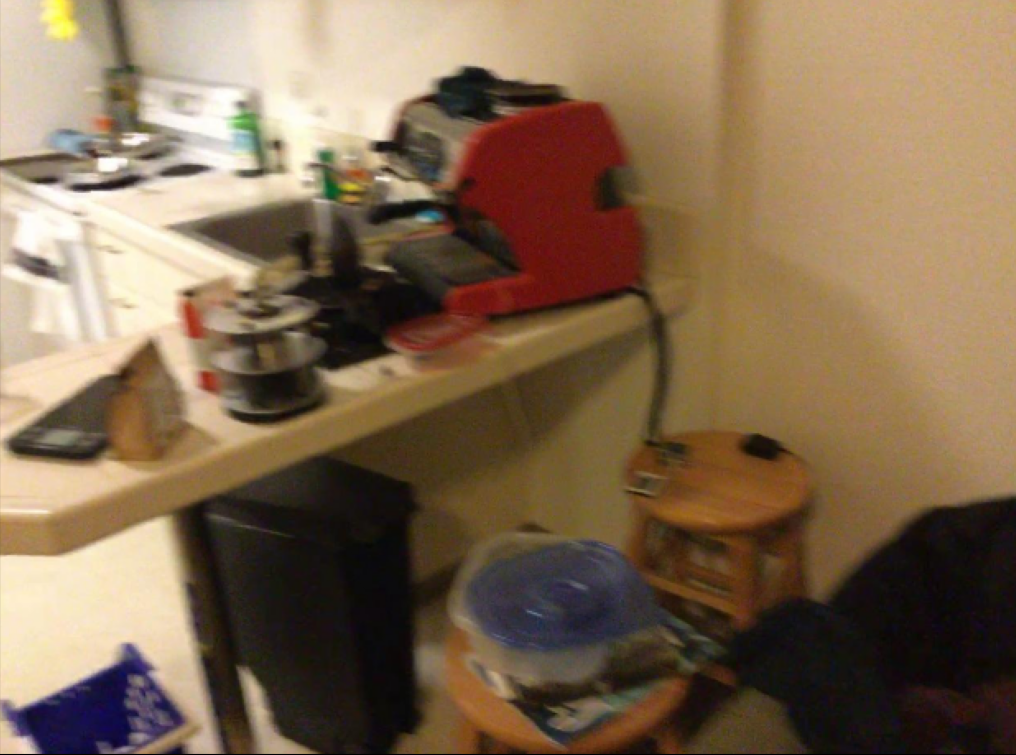}    
    \\
    
    \raisebox{0.5\height}{\makebox[0.01\textwidth]{\rotatebox{90}{\makecell{\scriptsize fr1-desk}}}}
    \includegraphics[width=\wratio\textwidth,height=\imgheight]{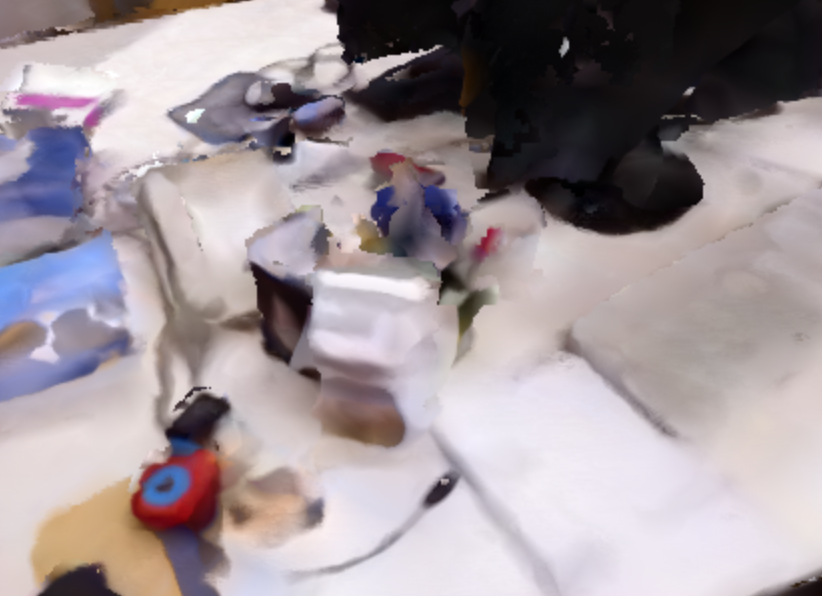}    
    \includegraphics[width=\wratio\textwidth,height=\imgheight]{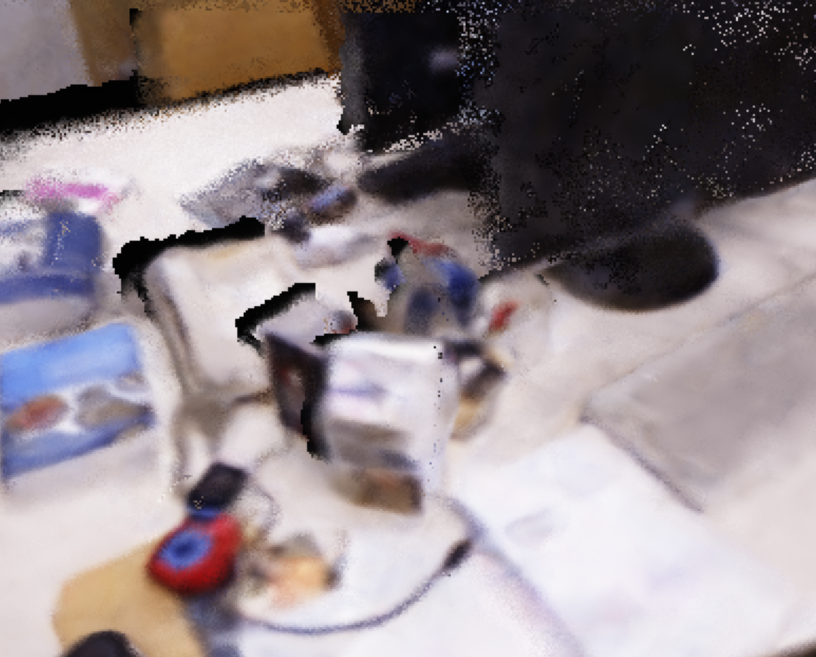}
    \includegraphics[width=\wratio\textwidth,height=\imgheight]{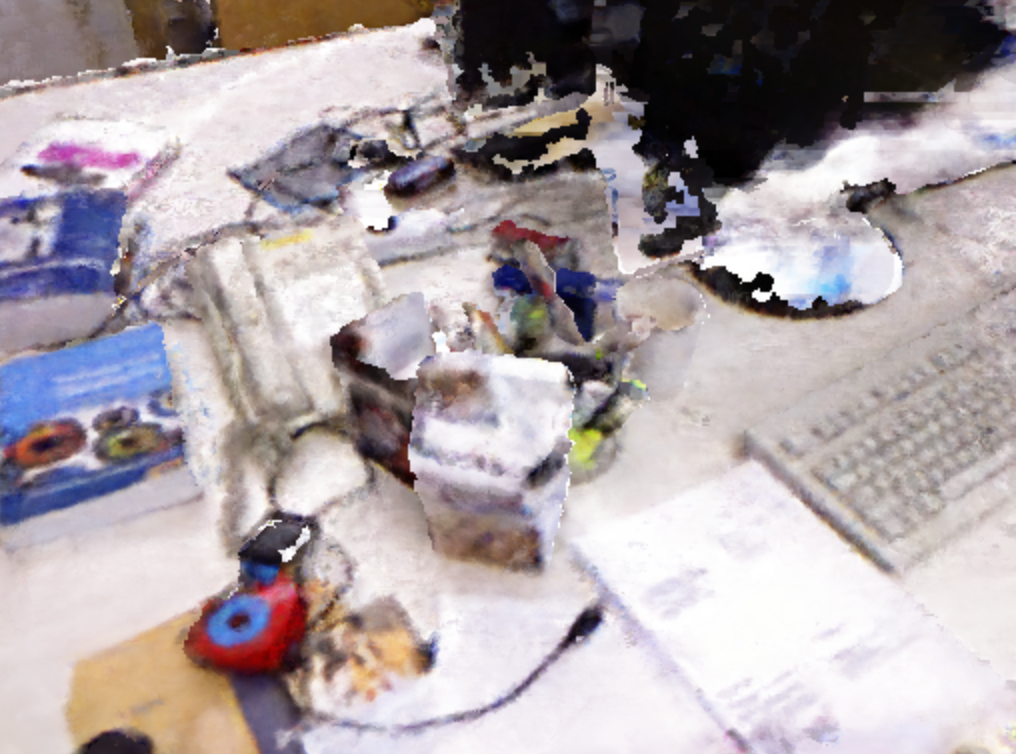}
    \includegraphics[width=\wratio\textwidth,height=\imgheight]{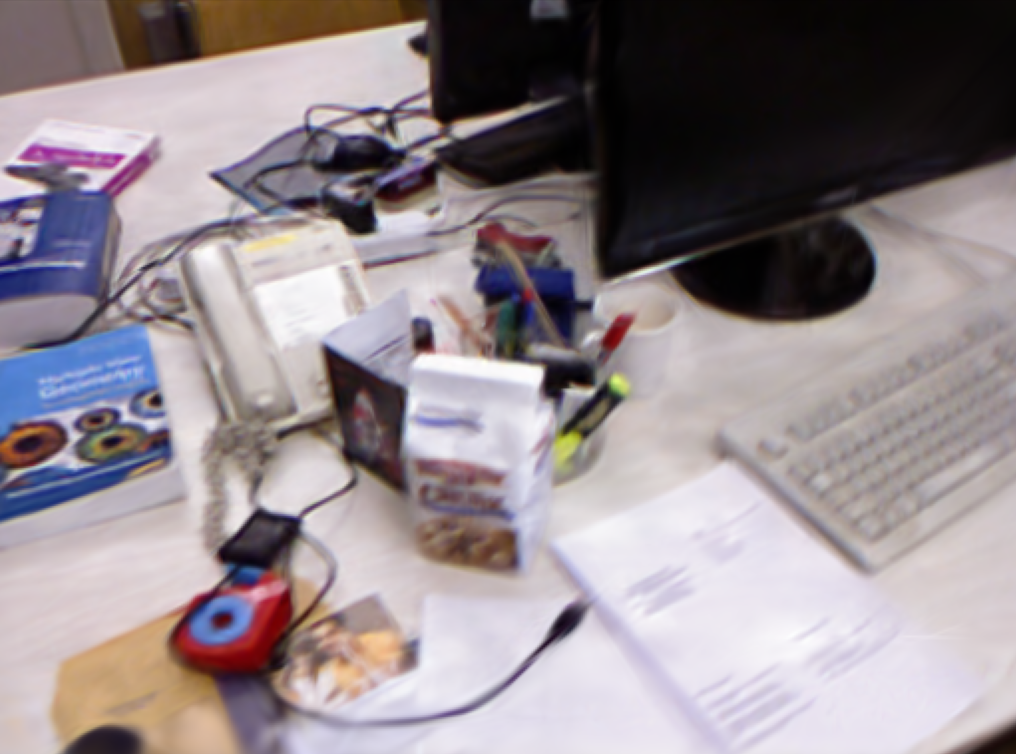}
    \includegraphics[width=\wratio\textwidth,height=\imgheight]{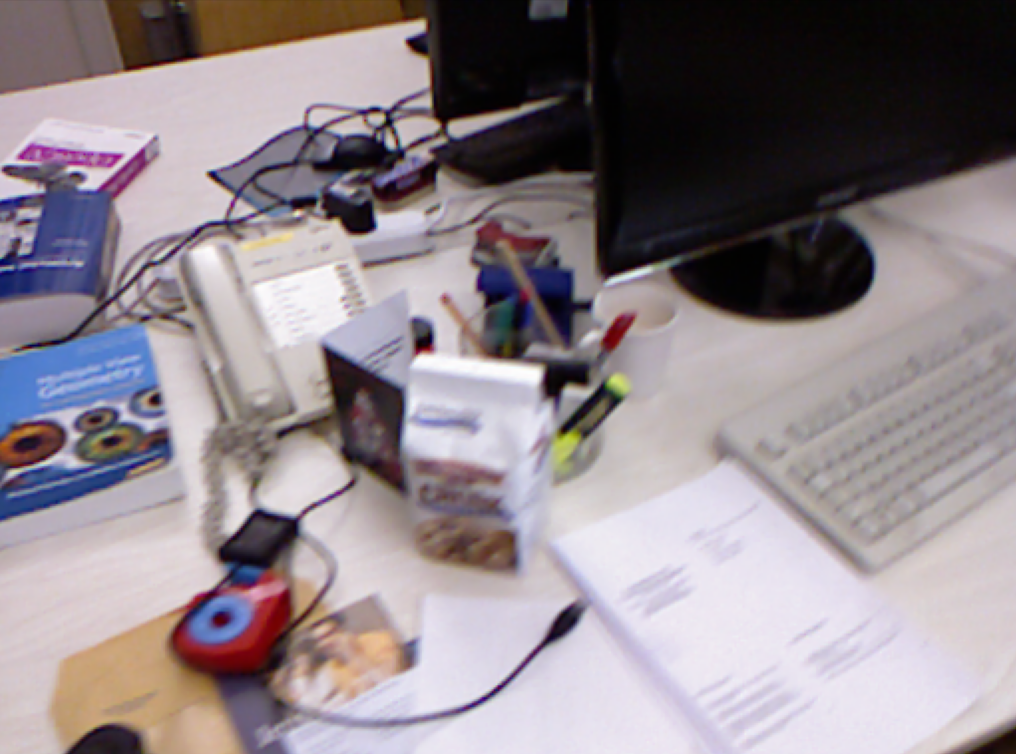}    
    \\
    
    \raisebox{0.5\height}{\makebox[0.01\textwidth]{\rotatebox{90}{\makecell{\scriptsize fr3-office}}}}
    \includegraphics[width=\wratio\textwidth,height=\imgheight]{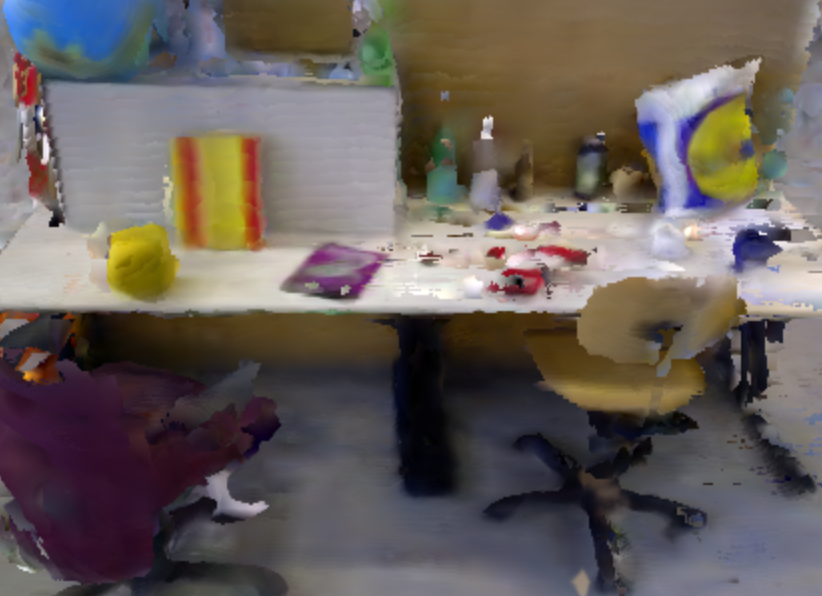}    
    \includegraphics[width=\wratio\textwidth,height=\imgheight]{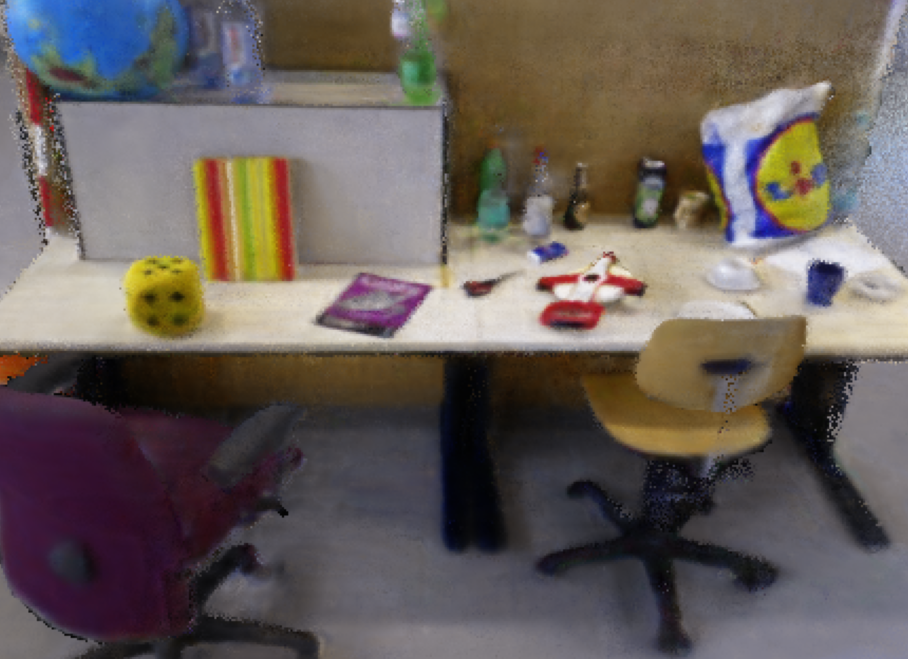}
    \includegraphics[width=\wratio\textwidth,height=\imgheight]{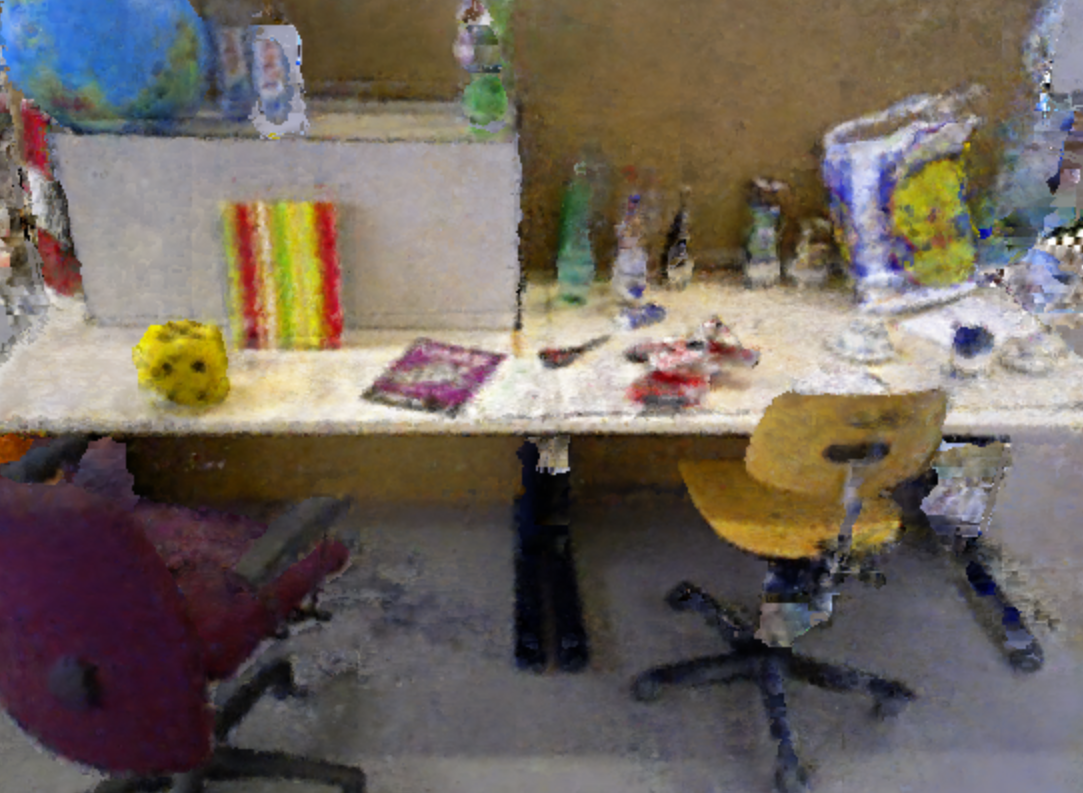}
    \includegraphics[width=\wratio\textwidth,height=\imgheight]{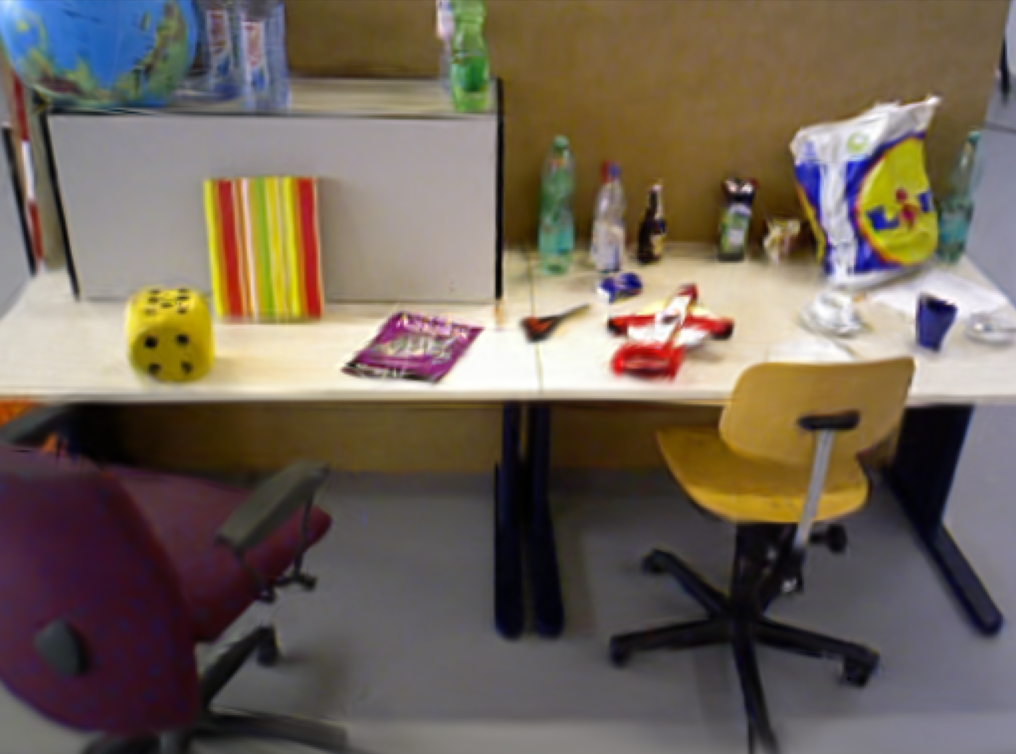}
    \includegraphics[width=\wratio\textwidth,height=\imgheight]{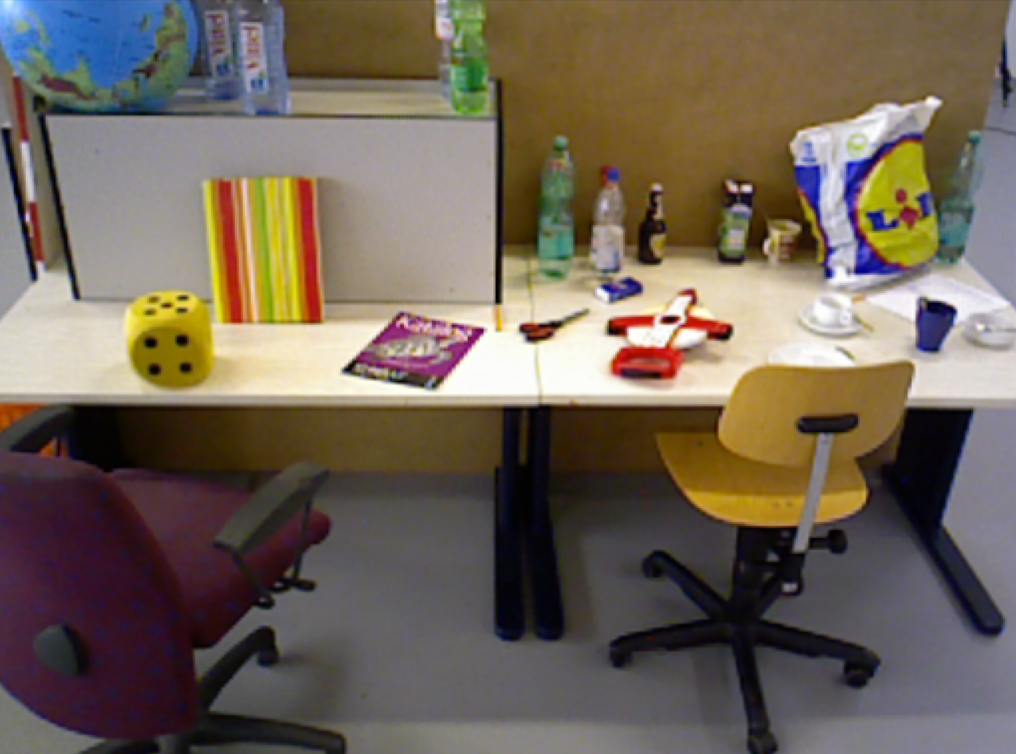}    
    \\    
        \raisebox{0.25\height}{\makebox[0.01\textwidth]{\rotatebox{90}{\makecell{\scriptsize 2e74812d00}}}}
    \includegraphics[width=\wratio\textwidth,height=\imgheight]{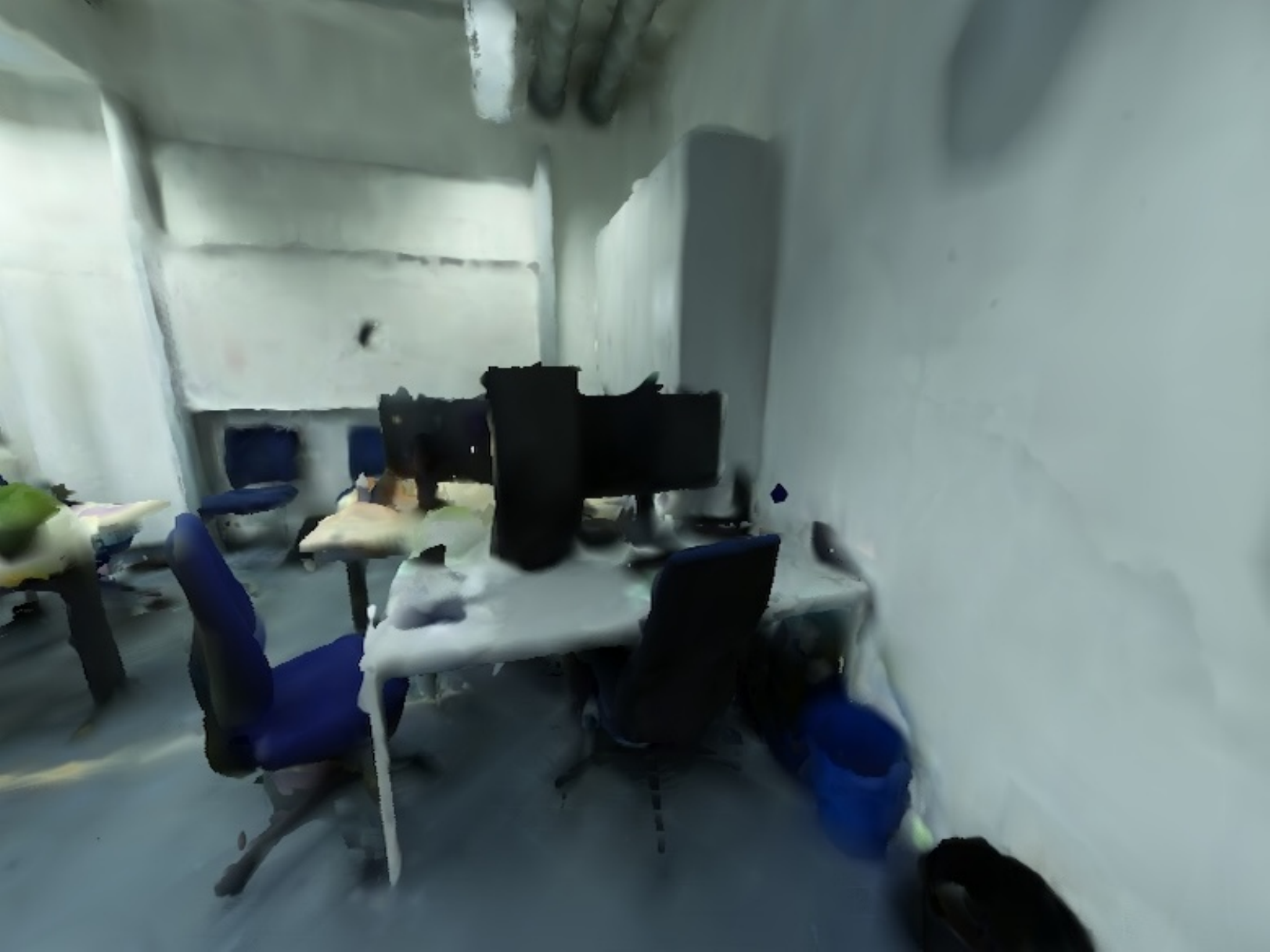}    
    \includegraphics[width=\wratio\textwidth,height=\imgheight]{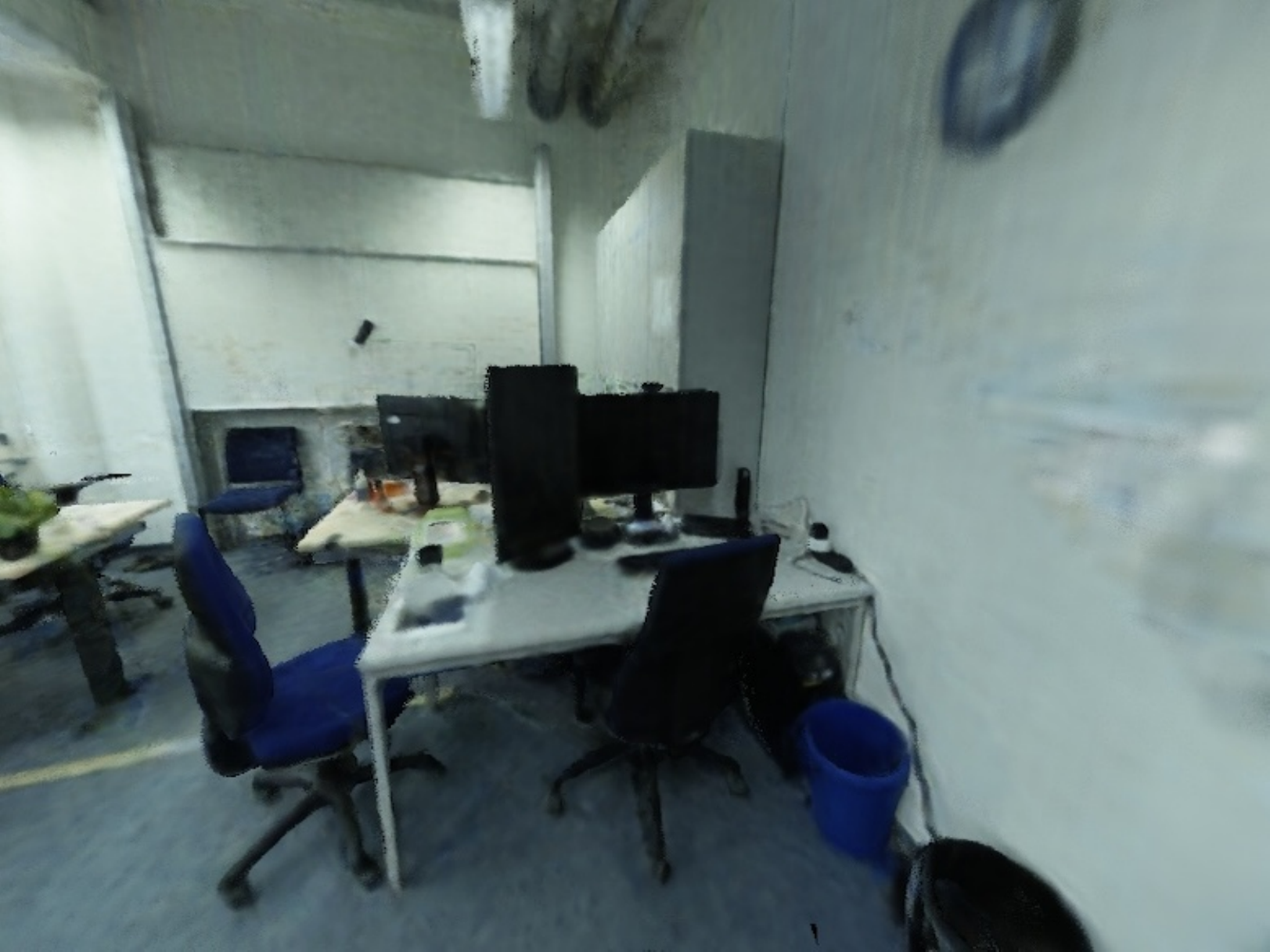}
    \includegraphics[width=\wratio\textwidth,height=\imgheight]{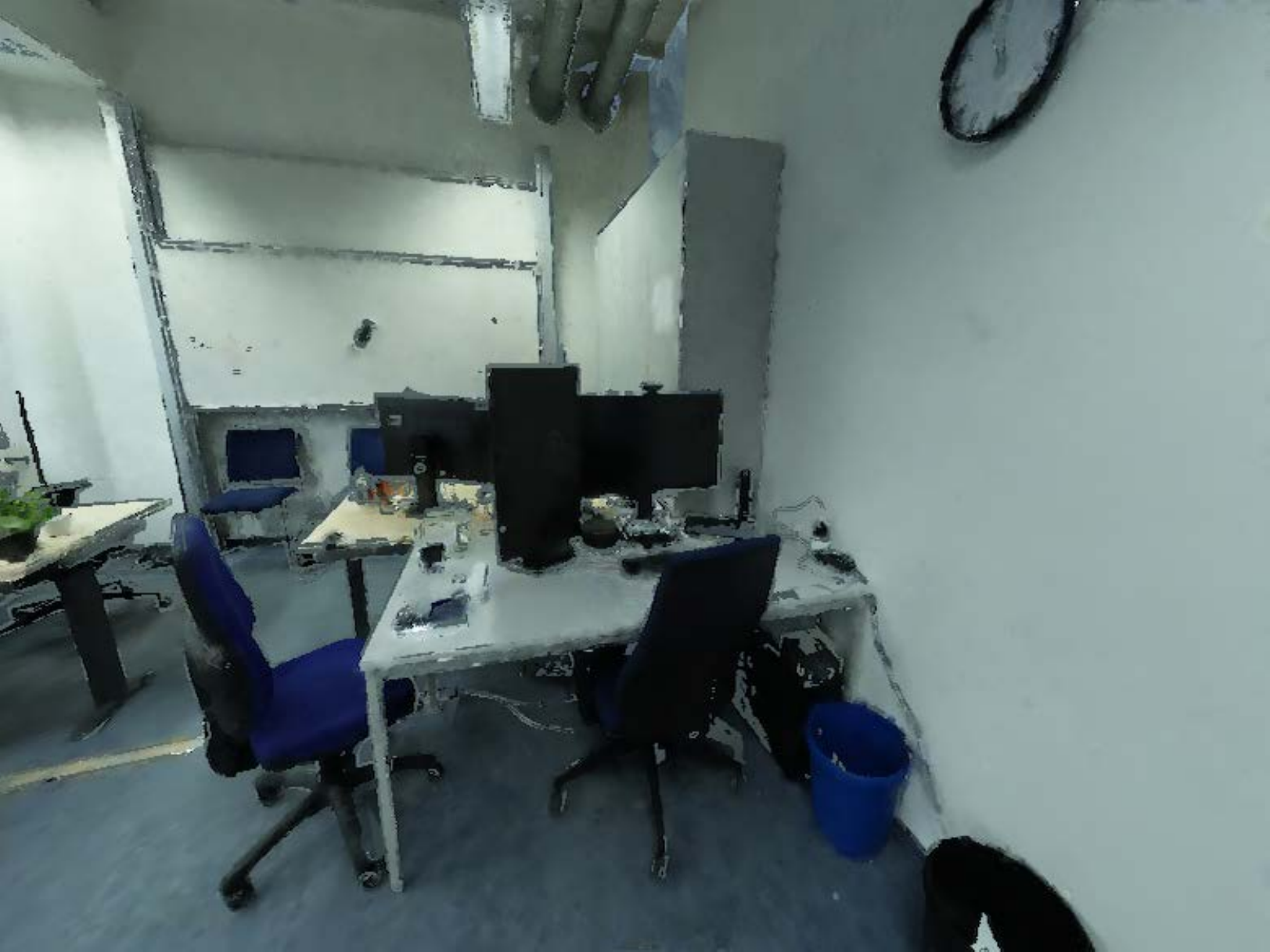}
    \includegraphics[width=\wratio\textwidth,height=\imgheight]{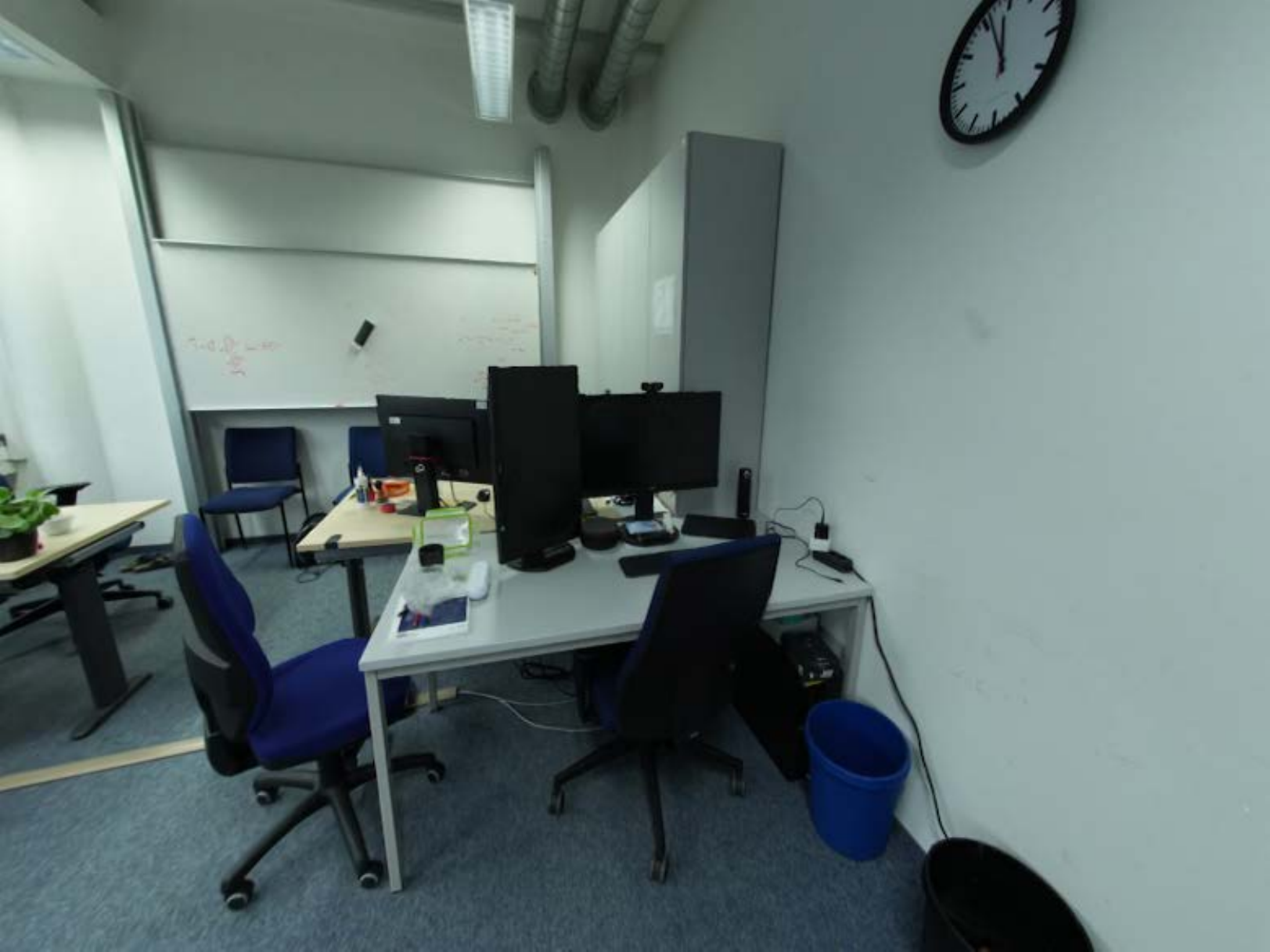}
    \includegraphics[width=\wratio\textwidth,height=\imgheight]{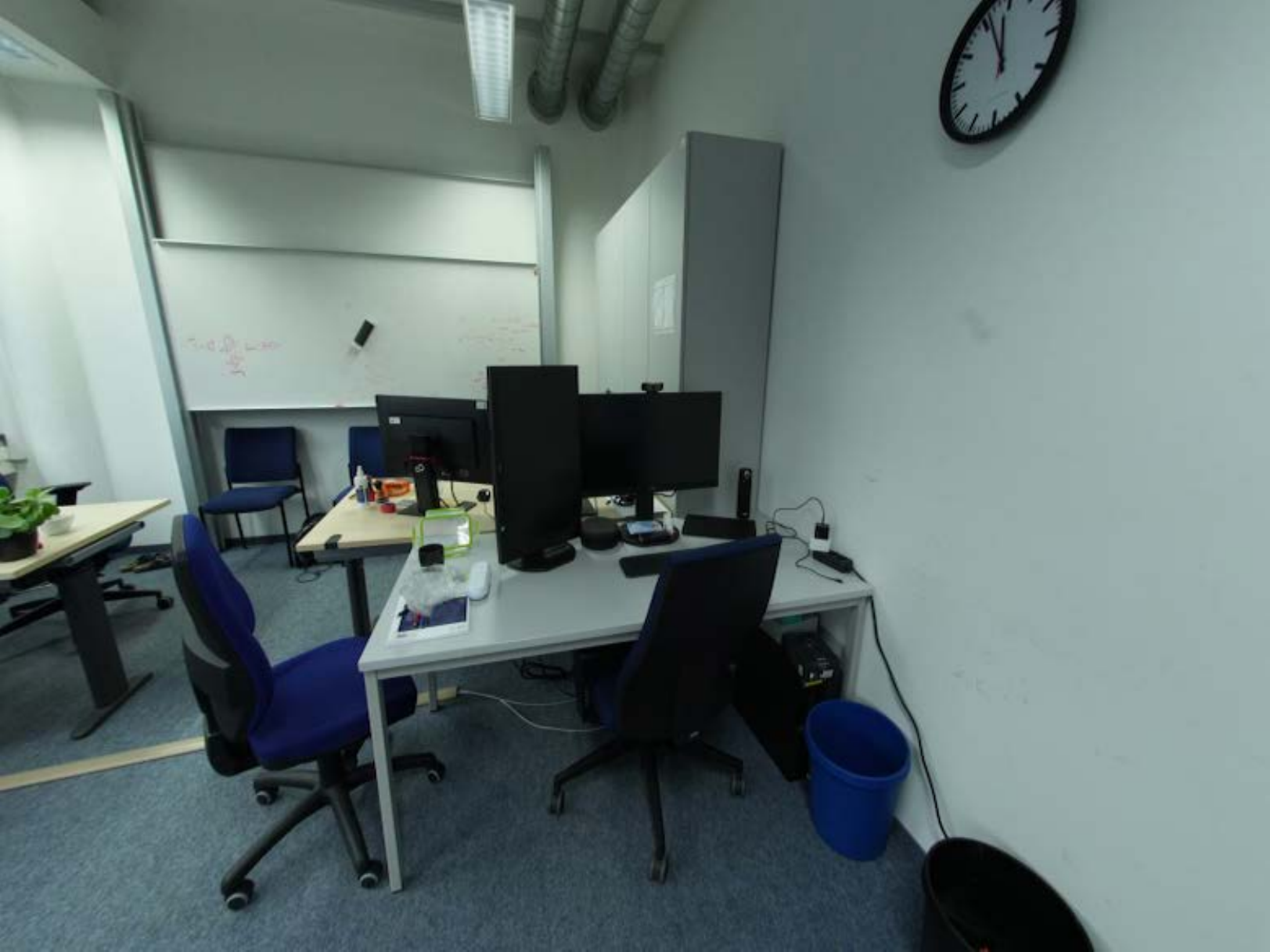}    
    \\ 
        \raisebox{0.25\height}{\makebox[0.01\textwidth]{\rotatebox{90}{\makecell{\scriptsize fb05e13ad1}}}}
    \includegraphics[width=\wratio\textwidth,height=\imgheight]{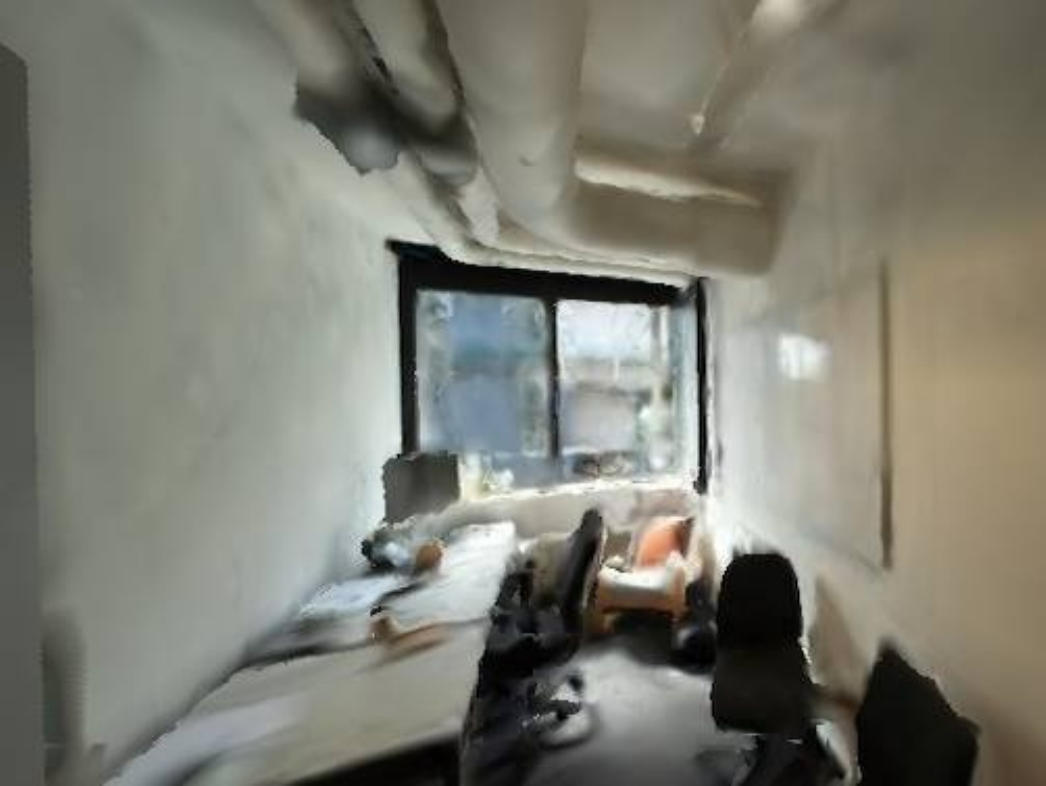}    
    \includegraphics[width=\wratio\textwidth,height=\imgheight]{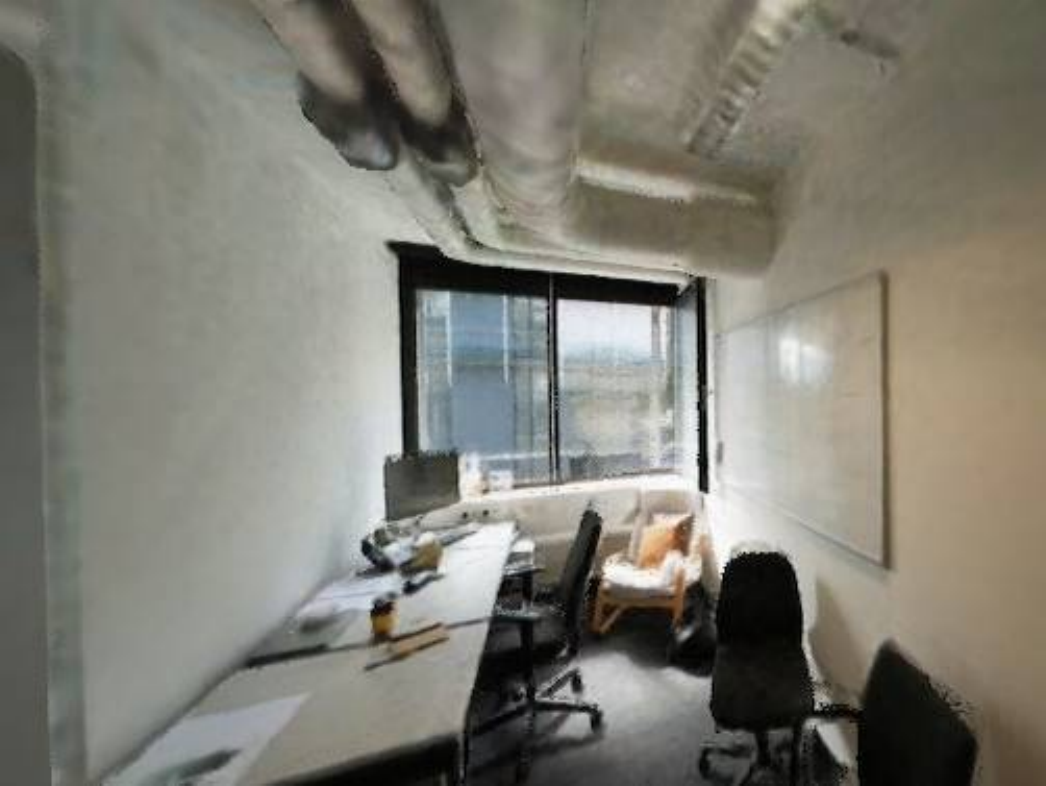}
    \includegraphics[width=\wratio\textwidth,height=\imgheight]{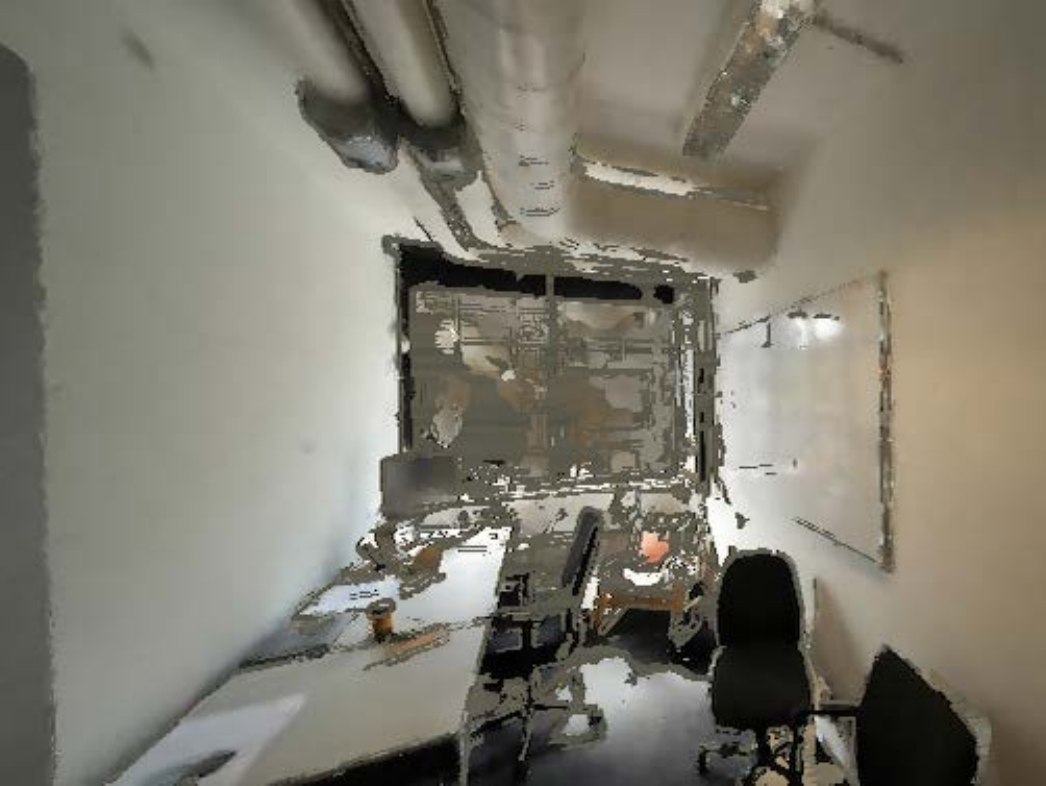}
    \includegraphics[width=\wratio\textwidth,height=\imgheight]{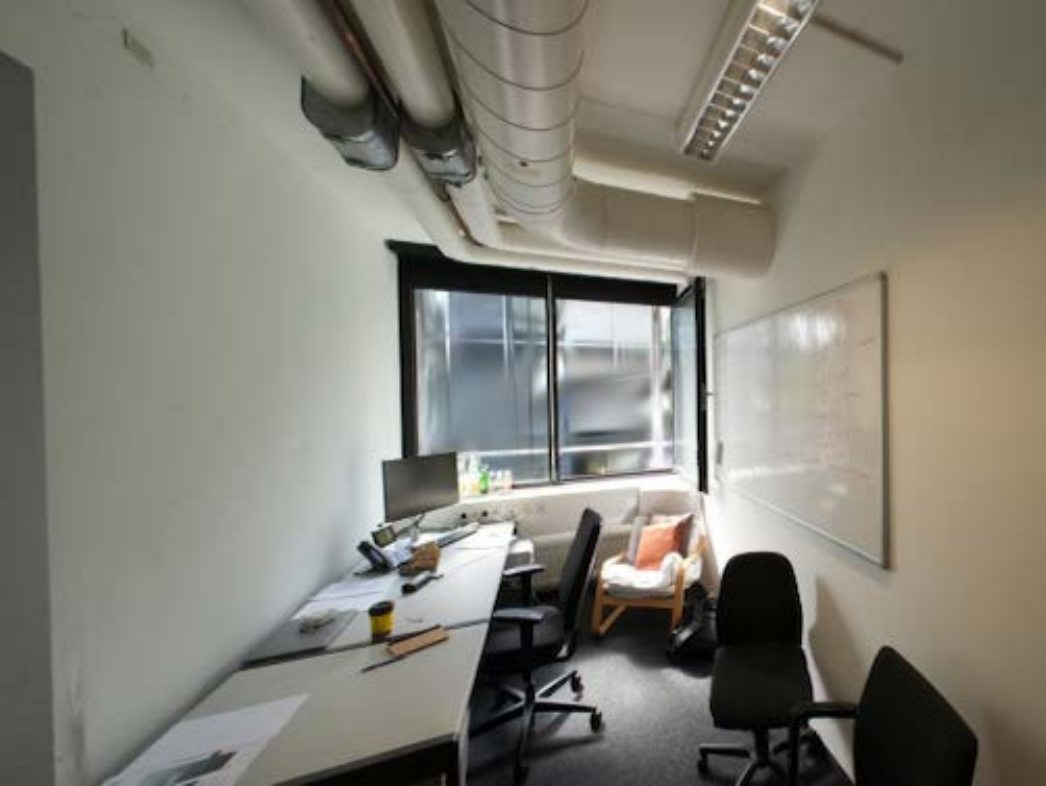}
    \includegraphics[width=\wratio\textwidth,height=\imgheight]{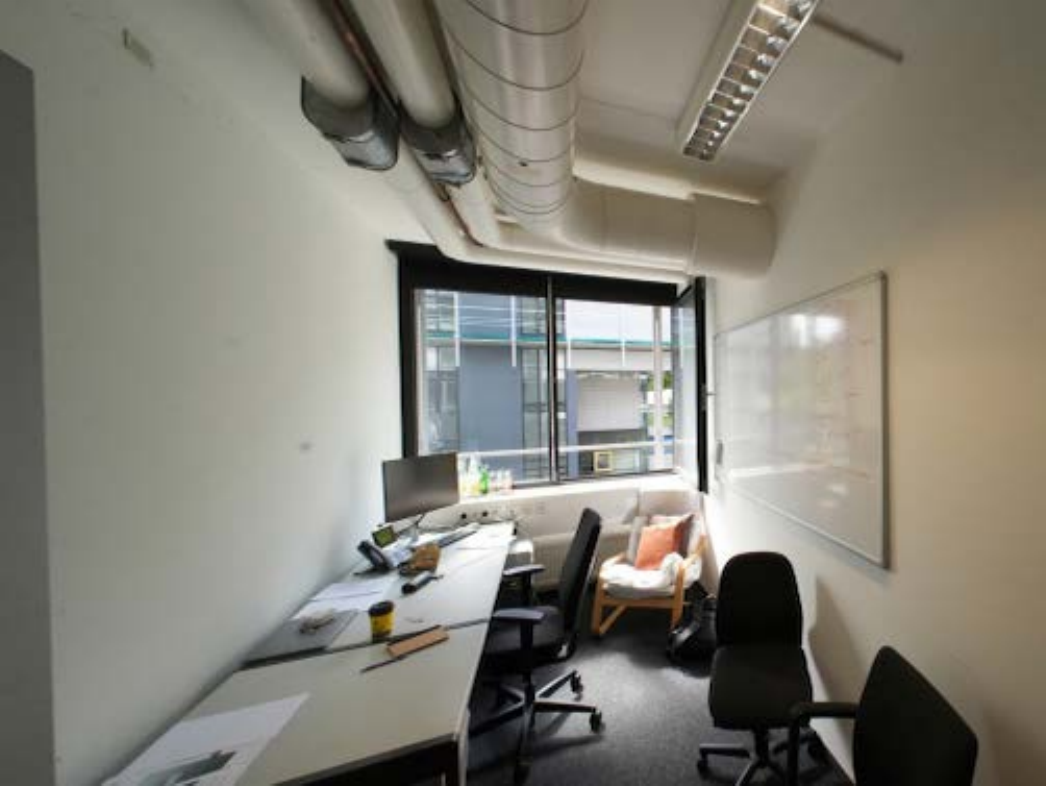}    
    \\ 
    \caption{\textbf{Rendering performance on ScanNet~\cite{Dai2017ScanNet}, TUM-RGBD~\cite{stuhmer2010real} and ScanNet++~\cite{yeshwanth2023scannet++}}. Thanks to 3D Gaussian splatting, \ours can encode more high-frequency details and substantially increase the quality of the renderings (please zoom in for a better view of the details). This is also supported by the quantitative results in \cref{tab:tum_rendering} and \cref{tab:scannet_rendering}.}
\label{fig:rendering_qualitative}
\end{figure*}

\boldparagraph{Rendering Performance.}
\cref{tab:replica_rendering} compares rendering performance and shows improvements over all the existing dense neural RGBD SLAM methods on synthetic data. \cref{tab:tum_rendering} and \cref{tab:scannet_rendering} show our state-of-the-art rendering performance on real-world datasets. \cref{fig:rendering_qualitative} shows exemplary full-resolution renderings where \ours yields more accurate details. Qualitative results on novel views are provided as a video in the supplementary.
\begin{table}[!htb]
\centering
\scriptsize
\caption{\textbf{Rendering Performance on Replica~\cite{straub2019replica}.} We outperform all existing dense neural RGBD methods on the commonly reported rendering metrics. Concurrent work is marked with an asterisk\textcolor{red}{$^*$}. }
\setlength{\tabcolsep}{1.5pt}
\renewcommand{\arraystretch}{1.1} %
\resizebox{\columnwidth}{!}{ 
\begin{tabular}{llccccccccc}
\toprule
Method & Metric & \texttt{Rm0} & \texttt{Rm1} & \texttt{Rm2} & \texttt{Off0} & \texttt{Off1} & \texttt{Off2} & \texttt{Off3} & \texttt{Off4} & Avg.\\
\midrule
\multirow{3}{*}{\makecell[l]{NICE-SLAM \cite{zhu2022nice}}}
& PSNR$\uparrow$ & 22.12 &  22.47 &  24.52 &  29.07 &  30.34 &  19.66 &  22.23 &  24.94 &  24.42 \\
& SSIM $\uparrow$ &  0.689 &   0.757 &  0.814 &  0.874 &   0.886 &  0.797 &  0.801 &  0.856 &  0.809 \\
& LPIPS $\downarrow$ &  0.330 &  0.271 &   0.208 &  0.229 &   0.181 &   0.235 &  0.209 &   0.198 &   0.233\\
[0.8pt] \hdashline \noalign{\vskip 1pt}
\multirow{3}{*}{\makecell[l]{Vox-Fusion \cite{yang2022vox}}} & PSNR$\uparrow$ &  22.39 &  22.36 &  23.92 &  27.79 &   29.83 &  20.33 &  23.47 &  25.21 &  24.41 \\
& SSIM$\uparrow$ &  0.683 &  0.751 &  0.798 &  0.857 &  0.876 &  0.794 &  0.803 &  0.847 &  0.801\\
& LPIPS$\downarrow$ &   0.303 &   0.269 &  0.234 &  0.241 &  0.184 &  0.243 &  0.213 &  0.199 &  0.236\\
[0.8pt] \hdashline \noalign{\vskip 1pt}
\multirow{3}{*}{\makecell[l]{ESLAM \cite{mahdi2022eslam}}} & PSNR$\uparrow$ &  25.25 &   27.39 &   28.09 &   30.33 &  27.04 &  27.99 &   29.27 &   29.15 &   28.06 \\
& SSIM$\uparrow$ &  0.874 &  0.89 &   0.935 &   0.934 &   0.910 &  0.942 &   0.953 &   0.948 &   0.923\\
& LPIPS$\downarrow$ & 0.315 &  0.296 &  0.245 &   0.213 &  0.254 & 0.238 &   0.186 &  0.210 & 0.245 \\
[0.8pt] \hdashline \noalign{\vskip 1pt}
\multirow{3}{*}{\makecell[l]{Point-SLAM \cite{sandstrom2023point}}}
& PSNR$\uparrow$  &  \rd32.40 & \nd34.08 & \nd35.50  &  \nd38.26 & \rd39.16 &  \nd33.99 &  \nd33.48 &  \nd33.49 &  \nd35.17\\
& SSIM$\uparrow$ &  \rd0.974 &  \nd0.977 &  \nd0.982 &  \nd0.983 &	 \nd0.986 &   \rd0.960 &   \nd0.960 &   \nd0.979 &   \nd0.975 \\
& LPIPS$\downarrow$ &  \rd0.113 &  \rd0.116 &  \rd0.111 &  \rd0.100 &	  \rd0.118 &  \rd0.156 &  \rd0.132 &  \nd0.142 &   \rd0.124 \\
[0.8pt] \hdashline \noalign{\vskip 1pt}
\multirow{3}{*}{\makecell[l]{SplaTAM\textcolor{red}{$^*$}\cite{keetha2023splatam}}} 
& PSNR$\uparrow$  & \nd32.86 & \rd33.89 & \rd35.25 & \rd38.26 & \nd39.17 & \rd31.97 & \rd29.70 & \rd31.81 & \rd34.11 \\
& SSIM$\uparrow$ & \nd0.98 & \rd0.97 & \rd0.98 & \rd0.98 & \rd0.98 & \nd0.97 & \rd0.95 & \rd0.95 & \rd0.97 \\
& LPIPS$\downarrow$ & \nd0.07 & \nd0.10 & \nd0.08 & \nd0.09 & \nd0.09 & \nd0.10 & \nd0.12 & \rd0.15 & \nd0.10 \\
\hdashline \noalign{\vskip 1pt}
\multirow{3}{*}{\makecell[l]{Gaussian-SLAM (ours)}} 
& PSNR$\uparrow$  & \fs38.88 & \fs41.80 & \fs42.44 & \fs46.40 & \fs45.29 & \fs40.10 & \fs39.06 & \fs42.65 & \fs42.08 \\
& SSIM$\uparrow$ & \fs0.993 & \fs0.996 & \fs0.996 & \fs0.998 & \fs0.997 & \fs0.997 & \fs0.997 & \fs0.997 & \fs0.996 \\
& LPIPS$\downarrow$ & \fs0.017 & \fs0.018 & \fs0.019 & \fs0.015 & \fs0.016 & \fs0.020 & \fs0.020 & \fs0.020 & \fs0.018 \\
\bottomrule
\end{tabular}}
\label{tab:replica_rendering}
\end{table}

\begin{table}[!htb]
\centering
\scriptsize
\caption{\textbf{Rendering Performance on TUM-RGBD~\cite{Sturm2012ASystems}.} We outperform existing dense neural RGBD methods on the commonly reported rendering metrics. For qualitative results, see \cref{fig:rendering_qualitative}.
Concurrent work is marked with an asterisk\textcolor{red}{$^*$}.}
\begin{tabularx}{\textwidth}{lXXXXXX}
\toprule
Method & Metric & \texttt{fr1/desk} & \texttt{fr2/xyz} & \texttt{fr3/office} & Avg.\\
\midrule
\multirow{3}{*}{\makecell[l]{NICE-SLAM~\cite{zhu2022nice}}}
& PSNR$\uparrow$ & 13.83 & \rd17.87 & 12.890 & 14.86 \\
& SSIM$\uparrow$ & 0.569 & 0.718 & 0.554 & 0.614 \\
& LPIPS$\downarrow$ & 0.482 & \rd0.344 & 0.498 & \rd0.441 \\[0.8pt] \hdashline \noalign{\vskip 1pt}
\multirow{3}{*}{\makecell[l]{Vox-Fusion~\cite{yang2022vox}}}
& PSNR$\uparrow$ & 15.79 & 16.32 & 17.27 & 16.46 \\
& SSIM$\uparrow$ & 0.647 & 0.706 & 0.677 & 0.677 \\
& LPIPS$\downarrow$ & 0.523 & 0.433 & 0.456 & 0.471 \\[0.8pt] \hdashline \noalign{\vskip 1pt}
\multirow{3}{*}{\makecell[l]{ESLAM~\cite{mahdi2022eslam}}}
& PSNR$\uparrow$ & 11.29 & 17.46 & 17.02 & 15.26 \\
& SSIM$\uparrow$ & \rd0.666 & 0.310 & 0.457 & 0.478 \\
& LPIPS$\downarrow$ & \rd0.358 & 0.698 & 0.652 & 0.569 \\[0.8pt] \hdashline \noalign{\vskip 1pt}
\multirow{3}{*}{\makecell[l]{Point-SLAM~\cite{sandstrom2023point}}}
& PSNR$\uparrow$ & \rd13.87 & 17.56 & \rd18.43 & \rd16.62 \\
& SSIM$\uparrow$ & 0.627 & \rd0.708 & \rd0.754 & \rd0.696 \\
& LPIPS$\downarrow$ & 0.544 & 0.585 & \rd0.448 & 0.526 \\[0.8pt] \hdashline \noalign{\vskip 1pt}
\multirow{3}{*}{\makecell[l]{SplaTAM\textcolor{red}{$^*$}~\cite{keetha2023splatam}}}
& PSNR$\uparrow$ & \nd22.00 & \nd24.50 & \nd21.90 & \nd22.80 \\
& SSIM$\uparrow$ & \nd0.857 & \fs0.947 & \nd0.876 & \nd0.893 \\
& LPIPS$\downarrow$ & \nd0.232 & \fs0.100 & \nd0.202 & \nd0.178 \\[0.8pt] \hdashline \noalign{\vskip 1pt}
\multirow{3}{*}{\makecell[l]{Gaussian-SLAM (ours)}}
& PSNR$\uparrow$ & \fs24.01 & \fs25.02 & \fs26.13 & \fs25.05 \\
& SSIM$\uparrow$ & \fs0.924 & \nd0.924 & \fs0.939 & \fs0.929 \\
& LPIPS$\downarrow$ & \fs0.178 & \nd0.186 & \fs0.141 & \fs0.168 \\
\bottomrule
\end{tabularx}
\label{tab:tum_rendering}
\end{table}

\begin{table}[!htb]
\centering
\scriptsize
\setlength{\tabcolsep}{3.2pt}
\renewcommand{\arraystretch}{1.1} %
\caption{\textbf{Rendering Performance on ScanNet~\cite{Dai2017ScanNet}.} We outperform existing dense neural RGBD methods on the commonly reported rendering metrics by a significant margin. For qualitative results, see \cref{fig:rendering_qualitative}. Concurrent work is marked with an asterisk\textcolor{red}{$^*$}.}
\begin{tabularx}{\textwidth}{llXXXXXXXXX}
\toprule
Method & Metric & \texttt{0000} & \texttt{0059} & \texttt{0106} & \texttt{0169} & \texttt{0181} & \texttt{0207} & Avg.\\
\midrule
\multirow{3}{*}{\makecell[l]{NICE-SLAM~\cite{zhu2022nice}}}
& PSNR$\uparrow$ & 18.71 &  16.55 &  17.29 &  \rd{18.75} & 15.56 & 18.38 & 17.54 \\
& SSIM$\uparrow$ & 0.641 & 0.605 & 0.646 & 0.629 & 0.562 & 0.646 & 0.621 \\
& LPIPS$\downarrow$ & 0.561 & 0.534 &  0.510 & 0.534 & 0.602 & 0.552 & 0.548 \\
\hdashline \noalign{\vskip 1pt}
\multirow{3}{*}{\makecell[l]{Vox-Fusion~\cite{yang2022vox}}} 
& PSNR$\uparrow$ & 19.06 & 16.38 &  \nd{18.46} &  18.69 &  16.75 &  19.66 &  18.17 \\
& SSIM$\uparrow$ & 0.662 & 0.615 &  \nd{0.753} & 0.650 & 0.666 &  \rd{0.696} &  \rd{0.673} \\
& LPIPS$\downarrow$ & 0.515 & 0.528 &  \rd{0.439} &  0.513 & 0.532 &  \rd{0.500} &  0.504 \\
\hdashline \noalign{\vskip 1pt}
\multirow{3}{*}{ESLAM~\cite{mahdi2022eslam}} 
& PSNR$\uparrow$ & 15.70 & 14.48 & 15.44 & 14.56 & 14.22 & 17.32 & 15.29 \\
& SSIM$\uparrow$ &  \rd{0.687} &  0.632 & 0.628 &  0.656 &  \rd{0.696} & 0.653 & 0.658 \\
& LPIPS$\downarrow$ &  \rd{0.449} &  \rd{0.450} & 0.529 &  \rd{0.486} &  0.482 & 0.534 &  \rd{0.488} \\
\hdashline \noalign{\vskip 1pt}
\multirow{3}{*}{\makecell[l]{Point-SLAM~\cite{sandstrom2023point}}} 
& PSNR$\uparrow$ &  \nd{21.30} &  \nd{19.48} & 16.80 & 18.53 &  \nd{22.27} &  \nd{20.56} &  \nd{19.82} \\
& SSIM$\uparrow$ &  \nd{0.806} &  \rd{0.765} &  \rd{0.676} &  \rd{0.686} &  \nd{0.823} &  \nd{0.750} &  \nd{0.751} \\
& LPIPS$\downarrow$ &  0.485 &  0.499 & 0.544 & 0.542 &  \nd{0.471} & 0.544 & 0.514 \\
\hdashline \noalign{\vskip 1pt}
\multirow{3}{*}{SplaTAM\textcolor{red}{$^*$}~\cite{keetha2023splatam}} 
& PSNR$\uparrow$ & \rd{19.33} & \rd{19.27} & \rd{17.73} & \nd{21.97} & \rd16.76 & \rd{19.8} & \rd{19.14} \\
& SSIM$\uparrow$ & 0.660 &  \nd{0.792} & 0.690 &  \nd{0.776} & 0.683 & 0.696 & 0.716 \\
& LPIPS$\downarrow$ & \nd0.438 &  \nd{0.289} & \nd{0.376} & \nd{0.281} & \rd{0.420} & \nd{0.341} & \nd{0.358} \\
\hdashline \noalign{\vskip 1pt}
\multirow{3}{*}{Gaussian-SLAM(ours)} 
& PSNR$\uparrow$ & \fs{28.539} & \fs{26.208} & \fs{26.258} & \fs{28.604} & \fs{27.789} & \fs{28.627} & \fs{27.67} \\
& SSIM$\uparrow$ & \fs{0.926} & \fs{0.9336} & \fs{0.9259} & \fs{0.917} & \fs{0.9223} & \fs{0.9135} & \fs{0.923} \\
& LPIPS$\downarrow$ & \fs{0.271} & \fs{0.211} & \fs{0.217} & \fs{0.226} & \fs{0.277} & \fs{0.288} & \fs{0.248} \\

\bottomrule
\end{tabularx}
\label{tab:scannet_rendering}
\end{table}

\boldparagraph{Tracking Performance.} \label{exp:tracking}
In \cref{tab:replica_tracking}, \cref{tab:tumrgb_tracking}, \cref{tab:scannet_tracking} and \cref{tab:scannetpp_tracking} we report the tracking accuracy. Our method outperforms the nearest competitor by 14\% on \cite{straub2019replica}. On TUM-RGBD dataset~\cite{Sturm2012ASystems}, \ours also performs better than all baseline methods. On ScanNet dataset, our method exhibits a drift due to low-quality depth maps and a large amount of motion blur. On ScanNet++, our Gaussian splatting-based method performs significantly better than NeRF-based methods. In addition, \ours demonstrates greater robustness compared to alternative approaches.
\begin{figure}[!htb]
\begin{floatrow}
\ffigbox[\FBwidth]{
    \begin{tabular}{@{}lp{0.7cm}p{0.6cm}p{1cm}p{0.7cm}@{}}
    \toprule
    \textbf{Method} & \texttt{desk} & \texttt{xyz} & \texttt{office} & \textbf{Avg.}\\
    \midrule
    NICE-SLAM \cite{zhu2022nice} &  4.3&  31.7 &  \nd3.9& 13.3\\
    Vox-Fusion \cite{yang2022vox} &  \rd3.5 & 1.5 & 26.0 & 10.3 \\
    Point-SLAM \cite{sandstrom2023point} &  4.3 & \nd1.3 & \fs3.5 & \nd3.0 \\
    SplaTAM\textcolor{red}{$^*$}\cite{keetha2023splatam} & \nd3.4 & \fs1.2 & 5.2 & \rd3.3 \\
    Gaussian SLAM & \fs 2.6 & \nd1.3 & \rd4.6 & \fs2.9 \\
    \bottomrule
    \end{tabular}    
}{
    \caption{\textbf{Tracking Performance on TUM-RGBD~\cite{Sturm2012ASystems}} (ATE RMSE$\downarrow$ [cm]). Our method outperforms all other methods on TUM\_RGBD. Concurrent work is marked with an asterisk\textcolor{red}{$^*$}.}
    \label{tab:tumrgb_tracking}
}
\ffigbox[\FBwidth]{
    \begin{tabular}{cccc}
    \toprule
    Alpha & Inlier & ATE RMSE  & PSNR \\
    mask  & mask  & [cm]$\downarrow$  & [dB]$\uparrow$ \\
    \midrule
    \redx & \redx              & 12.77 & 22.96  \\
    \greencheck & \redx        & \nd2.68 &  \nd24.01 \\
    \redx & \greencheck        & \rd8.96 &  \rd23.66 \\
    \greencheck & \greencheck  & \fs2.50 & \fs24.32 \\
    \bottomrule
    \end{tabular}
}{
    \caption{\textbf{Tracking Mask Ablation on \texttt{desk}.} Applying alpha and inlier masks to the tracking loss improves tracking leading to better rendering.}
    \label{tab:tracking_mask_ablation}
}
\end{floatrow}
\end{figure}

\begin{table}[!htb]
\centering
\scriptsize 
\caption{\textbf{Tracking Performance on Replica~\cite{straub2019replica}} (ATE RMSE $\downarrow$ [cm]). We outperform all other methods in on Replica. Concurrent work is marked with an asterisk\textcolor{red}{$^*$}.}
\begin{tabularx}{\textwidth}{lXXXXXXXXXXX}
\toprule
\textbf{Method} & \texttt{Rm0} & \texttt{Rm1} & \texttt{Rm2} & \texttt{Off0} & \texttt{Off1} & \texttt{Off2} & \texttt{Off3} & \texttt{Off4} & \textbf{Avg.}\\
\midrule
NICE-SLAM \cite{zhu2022nice} & 1.69 & 2.04 & 1.55 & 0.99 & 0.90 & 1.39 & 3.97 & 3.08 & 1.95 \\
Vox-Fusion \cite{yang2022vox} &  \fs0.27 & 1.33 & 0.47 & 0.70 & 1.11 & \rd0.46 & \fs0.26 & \rd0.58 & 0.65 \\
ESLAM \cite{mahdi2022eslam} &  0.71 & 0.70 & 0.52 & 0.57 & 0.55 & 0.58 & 0.72 & 0.63 & 0.63 \\
Point-SLAM \cite{sandstrom2023point} &  0.61 & \rd0.41 & \rd0.37 & \nd0.38 & \rd0.48 & 0.54 & 0.72 & 0.63 & \rd0.52\\
SplaTAM\textcolor{red}{$^*$}\cite{keetha2023splatam} &  \rd0.31 & \nd0.40 & \nd0.29 & \rd0.47 & \nd0.27 & \fs0.29 & \rd0.32 & \nd0.55 & \nd0.36\\
Gaussian SLAM (ours) &  \nd0.29 &  \fs0.29 &  \fs0.22 &  \fs0.37 & \fs0.23 & \nd0.41 & \nd0.30 & \fs0.35 & \fs0.31 \\
\bottomrule
\end{tabularx}
\label{tab:replica_tracking}
\end{table}

\begin{table}[!htb]
\centering
\scriptsize
\caption{\textbf{Tracking Performance on ScanNet~\cite{Dai2017ScanNet}} (ATE RMSE$\downarrow$ [cm]). Tracking on ScanNet is especially challenging due to low-quality depth maps and motion blur.}
\label{tab:scannet_tracking}
\begin{tabularx}{\textwidth}{lXXXXXXX}
\toprule
\textbf{Method} & \texttt{0000} & \texttt{0059} & \texttt{0106} & \texttt{0169} & \texttt{0181} & \texttt{0207} & \textbf{Avg.} \\
\midrule
NICE-SLAM \cite{zhu2022nice} & \nd12.00 & 14.00 & \fs7.90 & \fs10.90 & \nd13.40 & \fs6.20 & \fs10.70 \\
Vox-Fusion \cite{yang2022vox} & 68.84 & 24.18 & \nd8.41 & 27.28 & 23.30 & \rd9.41 & 26.90 \\
Point-SLAM \cite{sandstrom2023point} &  \fs10.24 & \fs7.81 & \rd8.65 & 22.16 & \rd14.77 & 9.54 & \rd12.19 \\
SplaTAM\textcolor{red}{$^*$}\cite{keetha2023splatam} & \rd12.83 & \rd10.10 & 17.72 & \nd12.08 & \fs11.10 & \nd7.46 & \nd11.88 \\
Gaussian SLAM  & 24.75 & \nd8.63 & 11.27  & \rd14.59 & 18.70 & 14.36 & 15.38 \\
\bottomrule
\end{tabularx}
\end{table}

\begin{table}[!htb]
\centering
\scriptsize 
\caption{\textbf{Tracking Performance on ScanNet++~\cite{yeshwanth2023scannet++}} (ATE RMSE $\downarrow$ [cm]). Our tracking proves to be robust and competitive in various real-world scenes. %
}
\begin{tabularx}{\textwidth}{lXXXXXX}
\toprule
\textbf{Method} & \texttt{b20a261fdf} & \texttt{8b5caf3398} & \texttt{fb05e13ad1} & \texttt{2e74812d00} & \texttt{281bc17764}  & \textbf{Avg.} \\
\midrule
Point-SLAM \cite{sandstrom2023point} & 246.16 & 632.99 & 830.79	& \rd271.42 & 574.86 & 511.24\\
ESLAM \cite{mahdi2022eslam} & \rd25.15	& \nd2.15 & \rd27.02 & \nd20.89	& \rd35.47 & \nd22.14 \\
SplaTAM\textcolor{red}{$^*$}\cite{keetha2023splatam} & \nd1.50 & \fs0.57 & \fs0.31	& 443.10 & \nd1.58	& \rd89.41 \\
Gaussian SLAM (ours) & \fs1.37 & \rd5.97 & \nd2.70	& \fs2.35	& \fs1.02	& \fs2.68 \\
\bottomrule
\end{tabularx}
\label{tab:scannetpp_tracking}
\end{table}

\boldparagraph{Reconstruction Performance.}
In \cref{tab:reconstruction} we compare our method to NICE-SLAM~\cite{zhu2022nice}, Vox-Fusion~\cite{yang2022vox}, ESLAM~\cite{mahdi2022eslam}, Point-SLAM~\cite{sandstrom2023point}, and concurrent SplaTAM\cite{keetha2023splatam} in terms of the geometric reconstruction accuracy on the Replica dataset~\cite{straub2019replica}. Our method performs on par with other existing dense SLAM methods.
\begin{table}[!htb]
\centering
\caption{\textbf{Reconstruction Performance on Replica~\cite{straub2019replica}.} Our method is comparable to the SOTA baseline Point-SLAM\cite{sandstrom2023point} which requires ground truth depth maps for inference while superior to other dense SLAM methods. Concurrent work is marked with an asterisk\textcolor{red}{$^*$}.}
\scriptsize
\setlength{\tabcolsep}{1.5pt}
\renewcommand{\arraystretch}{1.1} %
\resizebox{\columnwidth}{!}{ 
\begin{tabular}{llccccccccc}
\toprule
Method & Metric & \texttt{Rm0} & \texttt{Rm1} & \texttt{Rm2} & \texttt{Off0} & \texttt{Off1} & \texttt{Off2} & \texttt{Off3} & \texttt{Off4} & Avg.\\
\midrule
\multirow{2}{*}{\makecell[l]{NICE-SLAM \cite{zhu2022nice}}}
& Depth L1 [cm]$\downarrow$ & 1.81  &  1.44	&  2.04	&  1.39	&  1.76	&8.33	&4.99	&2.01	&2.97 \\
& F1 [$\%$]$\uparrow$ & 45.0	&  44.8 & 43.6	& 50.0	& 51.9	& 39.2	& 39.9	& 36.5	& 43.9\\
[0.8pt] \hdashline \noalign{\vskip 1pt}
\multirow{2}{*}{\makecell[l]{Vox-Fusion \cite{yang2022vox}}}
& Depth L1 [cm]$\downarrow$ & 1.09 & 1.90 & 2.21  & 2.32 & 3.40  &  4.19  & 2.96 & 1.61 & 2.46\\
& F1 [$\%$]$\uparrow$  & 69.9 &  34.4 &  59.7 &  46.5 &   40.8 &  51.0 & 64.6 & 50.7 &  52.2\\
[0.8pt] \hdashline \noalign{\vskip 1pt}
\multirow{2}{*}{ESLAM~\cite{mahdi2022eslam}} &
Depth L1 [cm] $\downarrow$ &  0.97  & 1.07  &  1.28  &  0.86 &  1.26  &  1.71  & \nd1.43  &  1.06 &  1.18 \\
& F1 [$\%$] $\uparrow$  &  81.0 &  82.2  &  83.9  &  78.4 &  75.5  &  77.1  &  75.5  &  79.1 &  79.1 \\ [0.8pt] \hdashline \noalign{\vskip 1pt}
\multirow{2}{*}{\makecell[l]{Point-SLAM \cite{sandstrom2023point}}}
& Depth L1 [cm]$\downarrow$ & \nd0.53 & \fs0.22 & \fs0.46 & \fs0.30 &  \nd0.57 & \fs0.49 &  \fs0.51 &  \nd0.46 &  \fs0.44 \\
& F1 [$\%$]$\uparrow$  & \rd86.9 & \fs92.3 & \fs90.8 & \fs93.8 & \fs91.6 & \fs89.0 &   \fs88.2  &   \nd85.6 &   \fs89.8 \\
[0.8pt] \hdashline \noalign{\vskip 1pt}
\multirow{2}{*}{\makecell[l]{SplaTAM\textcolor{red}{$^*$}\cite{keetha2023splatam}}} 
& Depth L1 [cm]$\downarrow$ & \fs0.43 & \rd0.38 &  \nd0.54  & \nd0.44 & \rd0.66 & \rd1.05 & \rd1.60 & \rd0.68  & \rd0.72 \\
& F1 [$\%$]$\uparrow$ & \fs89.3 & \rd88.2 & \rd88.0 & \nd91.7 & \rd90.0 & \rd85.1 & \rd77.1 & \rd80.1  & \rd86.1 \\
\hdashline \noalign{\vskip 1pt}
\multirow{2}{*}{\makecell[l]{Gaussian SLAM (ours)}} 
& Depth L1 [cm]$\downarrow$ & \rd0.61 & \nd0.25 & \nd0.54 & \rd0.50 & \fs0.52 & \nd0.98 & 1.63 & \fs0.42 & \nd0.68 \\
& F1 [$\%$]$\uparrow$  & \nd88.8 & \nd91.4 & \nd90.5 & \nd91.7 & \nd90.1 & \nd87.3 & \nd84.2 & \fs87.4 & \nd88.9 \\
\bottomrule
\end{tabular}}
\label{tab:reconstruction}
\end{table}

\boldparagraph{Runtime Comparison.}
In \cref{tab:runtime} we compare runtime usage on the Replica \texttt{office0} scene. We report both per-iteration and per-frame runtime. The per-frame runtime is calculated as the optimization time spent on all mapped frames divided by sequence length, while the per-iteration runtime is the average mapping iteration time.
\begin{table}[!htb]
  \centering
  \scriptsize
  \setlength{\tabcolsep}{5.0pt}
  \renewcommand{\arraystretch}{1.0}
  \caption{\textbf{Average Mapping, Tracking, and Rendering Speed on Replica} \texttt{office0}. Per-frame runtime is calculated as the total optimization time divided by sequence length. All metrics are profiled using an NVIDIA RTX A6000 GPU.}  
  \begin{tabular}{lccccc}
    \toprule
    Method & Mapping    & Mapping & Tracking  & Tracking & Rendering \\
           & /Iteration(ms) &/Frame(s)   &/Iteration(ms) & /Frame(s)   & (FPS) \\
    \midrule
    NICE-SLAM~\cite{zhu2022nice} & 89  & \rd1.15  & \rd27 & \rd1.06 & 2.64 \\
    Vox-Fusion~\cite{yang2022vox} & 98 &  1.47  & 64 & 1.92  & 1.63 \\
    ESLAM~\cite{mahdi2022eslam} & \nd30 & \fs0.62 & \nd18 & \fs0.15 & 0.65 \\
    Point-SLAM~\cite{sandstrom2023point} &  \rd57 & 3.52 & \rd27 & 1.11 & \nd2.96 \\
    SplaTAM\textcolor{red}{$^*$}~\cite{keetha2023splatam} & 81 & 4.89 & 67 & 2.70  & \fs2175 \\
    \ours (ours) &  \fs24  &  \nd0.93 & \fs14 & \nd0.83 &  \fs2175  \\
    \bottomrule
  \end{tabular}
  \label{tab:runtime}
\end{table}

\boldparagraph{Ablation Study.}\label{par:ablation}
In \cref{tab:tracking_mask_ablation} we ablate the effectiveness of soft alpha mask $M_\text{alpha}$ and inlier mask $M_\text{inlier}$ for the tracking performance on TUM-RGBD \texttt{fr1/desk} scene. 
It demonstrates the effectiveness of both masks with the soft alpha mask having more performance impact.

\boldparagraph{Limitations and Future Work.} Although we have effectively used 3D Gaussians for online dense SLAM, tracking a camera trajectory on data with lots of motion blur and low-quality depth maps remains challenging. We also believe that some of our empirical hyperparameters like keyframe selection strategy can be made test time adaptive or learned. Finally, trajectory drift is inevitable in frame-to-model tracking without additional techniques like loop-closure or bundle adjustment which might be an interesting future work.

\section{Conclusion}
\label{sec:conclusion}
We introduced \ours, a dense SLAM system based on 3D Gaussian Splatting as the scene representation that enables unprecedented re-rendering capabilities.
We proposed effective strategies for efficient seeding and online optimization of 3D Gaussians, their organization in sub-maps for better scalability, and a frame-to-model tracking algorithm. Compared to previous SOTA neural SLAM systems like Point-SLAM~\cite{sandstrom2023point} we achieve faster tracking and mapping while obtaining better rendering results on synthetic and real-world datasets. We demonstrated that \ours yields top results in rendering, camera pose estimation, and scene reconstruction on a variety of datasets. 
\\

\appendix
\section{Abstract}
This supplementary material accompanies the main paper by providing further information for better reproducibility as well as additional evaluations and qualitative results.

\section{Further Implementation Details} 
\label{sec:supp_implementation}
The inlier mask $M_\text{inlier}$ in tracking loss filters out pixels that have depth errors $50$ times larger than the median depth error of the current re-rendered depth map. Pixels without valid depth input are also excluded as the inconsistent re-rendering in those areas can misguide the pose optimization. For soft alpha mask, we adopt $M_\text{alpha} = \alpha^3$ for per-pixel loss weighting. The opacity values for added Gaussians are initialized as 0.5 and their initial scales are set to the nearest neighbor distances in the active sub-map. At the middle and the end of mapping iterations on new keyframes, we prune Gaussians having opacity values lower than a threshold $o_\text{thre}$. We set $o_\text{thre}=0.1$ for Replica~\cite{straub2019replica} and $0.5$ for all other datasets. Additionally, we use multi-scale RGBD odometry~\cite{park2017colored} to help initialize the pose for tracking optimization on Replica dataset, for all other datasets, we use pose initialization based on constant speed assumption. On Scannet++ dataset~\cite{yeshwanth2023scannet++}, if at the initialized pose, the re-rendering loss is 50 times larger than the running average of the re-rendering loss after tracking optimization, we use the odometry to re-initialize the pose for the current frame.

Upon completing the pipeline for the input sequence, we merge the saved sub-maps into a global map. Specifically, we select Gaussians from each sub-map in sequence as candidates and add them using the same nearest neighbor checking rule as in the sub-map building. Finally, as a post-processing step, we perform color refinement on the global map for 10000 iterations.

\section{Additional Experiments} 
\label{sec:supp_exps}

\begin{figure}[htb]
  \centering
  \begin{subfigure}[c]{0.45\textwidth}
    \includegraphics[width=\textwidth]{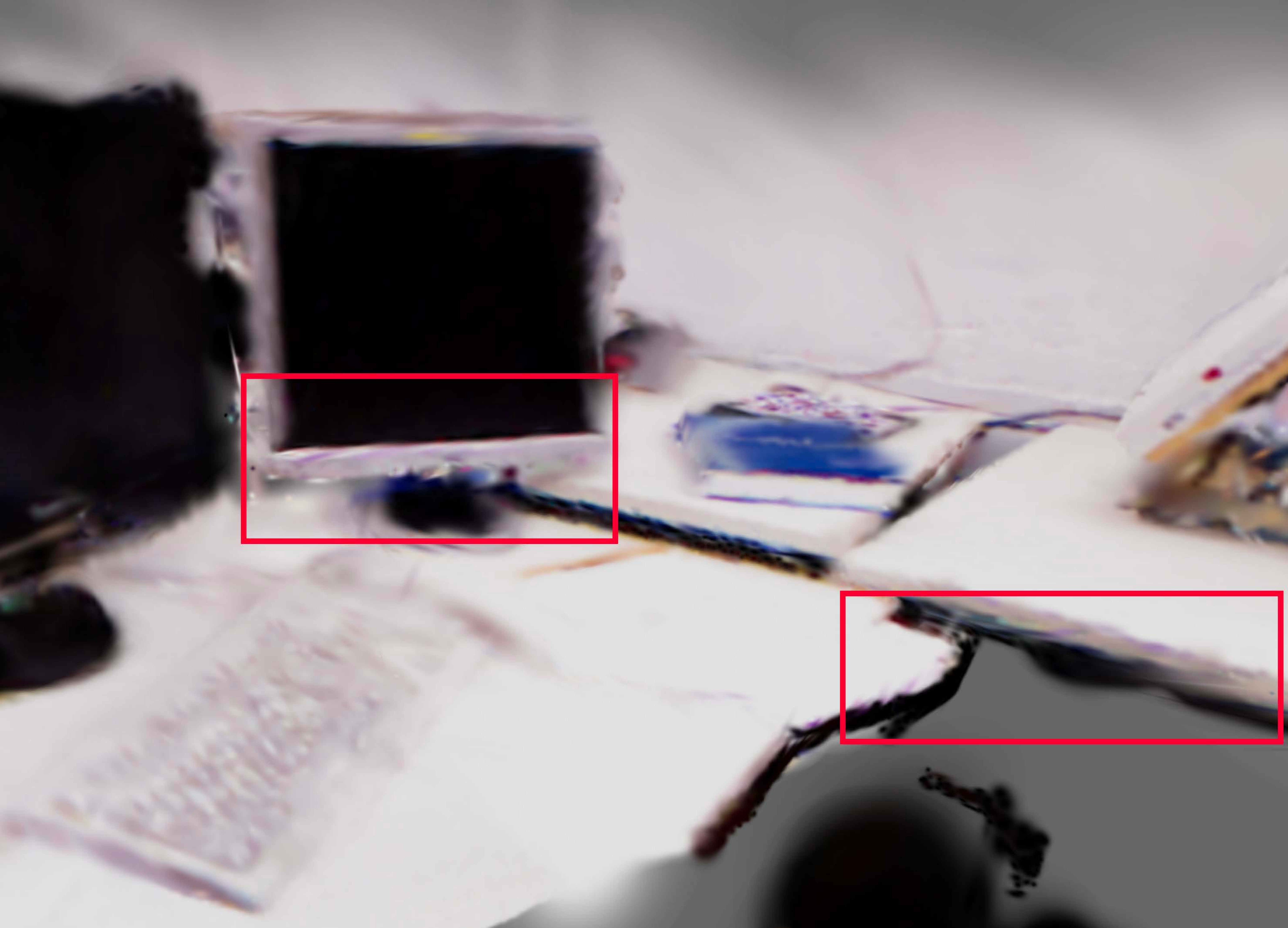}
    \caption{with isotropic regularization}
    \label{fig:sub1}
  \end{subfigure}
  \hfill
  \begin{subfigure}[c]{0.45\textwidth}
    \includegraphics[width=\textwidth]{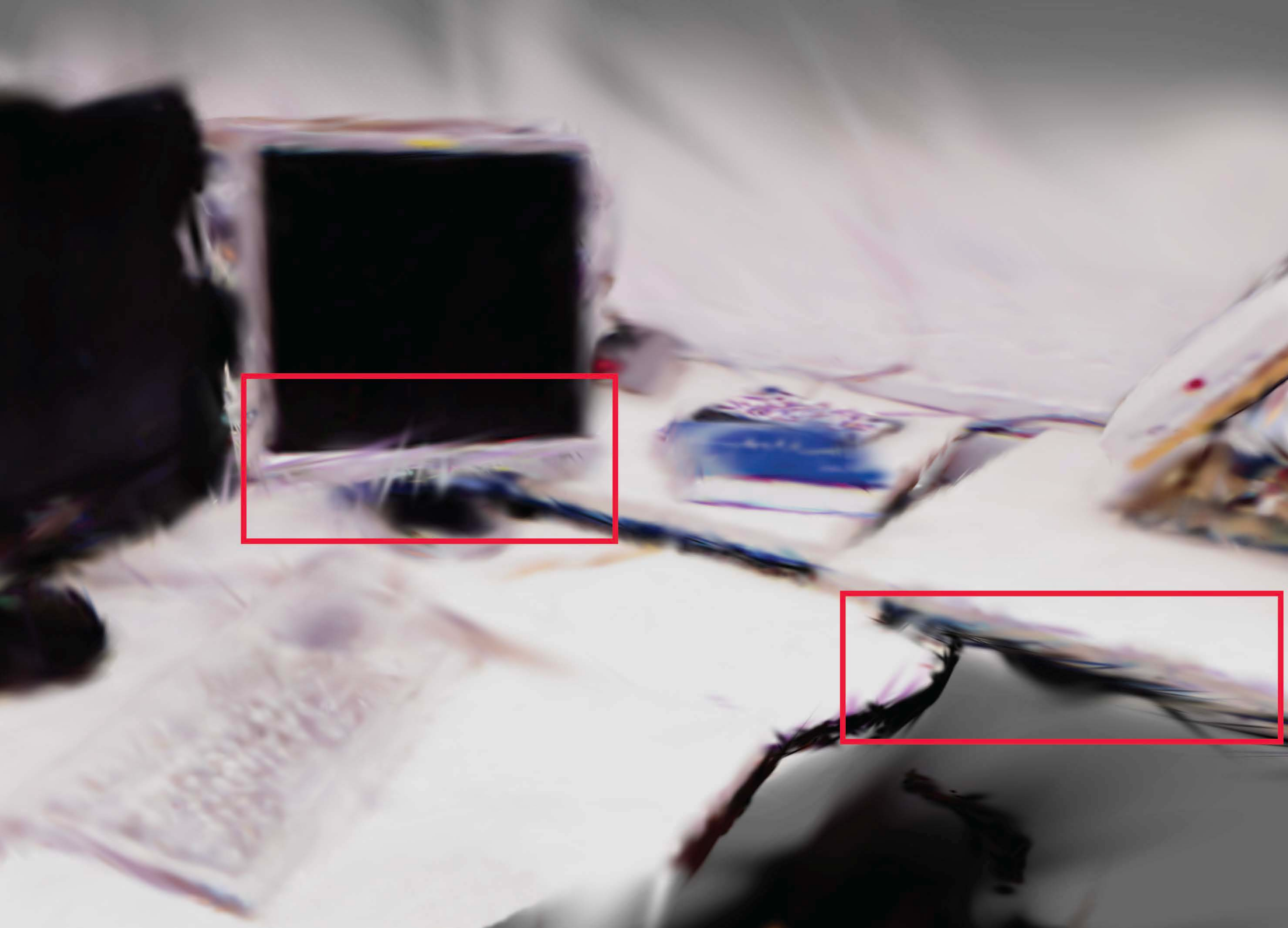}
    \caption{without isotropic regularization}
    \label{fig:supp_splats_a}
  \end{subfigure}

  \caption{\textbf{Gaussians splatted at a held-out view in the mapped \texttt{fr1/desk} scene.} Notice the highlighted areas in red rectangles, where elongated Gaussians are clearly noticeable if isotropic regularization is not applied.}
  \label{fig:iso_reg_splats}
\end{figure}

\boldparagraph{Isotropic Regularization Ablation.}
In \cref{tab:iso_reg_ablation_replica} and \cref{tab:iso_reg_ablation_tum} we ablate the isotropic regularization term $L_\text{reg}$ in the mapping loss. On Replica dataset~\cite{straub2019replica}, where the RGBD inputs are synthetic and noise-free, the $L_\text{reg}$ terms improves both tracking and rendering performance marginally. While on real-world TUM-RGBD dataset~\cite{stuhmer2010real}, $L_\text{reg}$ proves critical for accurate camera tracking. \cref{fig:iso_reg_splats} further examines the underlying Gaussians on \texttt{fr1/desk} scene at a held-out view. Without isotropic regularization, the elongated Gaussians are evident. While they slightly enhance rendering by overfitting to training views, they are detrimental to pose optimization, which relies on re-rendering at novel views.
\label{par:supp_iso_reg}
\begin{table}[!htb]
\centering
\scriptsize
\renewcommand{\arraystretch}{1.5}

\begin{tabularx}{0.8\textwidth}{lcXXXXXXXXX}
\toprule
Metric & $L_\text{reg}$ & \texttt{Rm0} & \texttt{Rm1} & \texttt{Rm2} & \texttt{Off0} & \texttt{Off1} & \texttt{Off2} & \texttt{Off3} & \texttt{Off4} & Avg. \\
\midrule
\multirow{2}{*}[-0em]{\makecell[l]{ATE $\downarrow$}}
& \redx       &\fs0.25 &0.36 &0.27  &0.52  &\fs0.23  &\fs0.37  &\fs0.30  &0.41  & 0.34 \\
& \greencheck &0.29 &\fs0.29 &\fs0.22  & \fs0.37 &\fs0.23  &0.41  &\fs0.30  &\fs0.35 & \fs0.31 \\
\midrule
\multirow{2}{*}[-0em]{\makecell[l]{PSNR $\uparrow$}}
& \redx       &38.83& 41.71& 42.18& 46.12& 44.72& 39.72&38.94&42.58&41.85\\
& \greencheck &\fs38.88&\fs41.80&\fs42.44&\fs46.40&\fs45.29&\fs40.10&\fs39.06&\fs42.65&\fs42.08\\

\bottomrule
\end{tabularx}

\caption{\textbf{Isotropic Regularization Ablation on Replica dataset~\cite{straub2019replica}.} The regularization term $L_\text{reg}$ improves both tracking and rendering performance.}
\label{tab:iso_reg_ablation_replica}
\end{table}

\begin{table}[!htb]
\centering
\scriptsize
\renewcommand{\arraystretch}{1.5}

\begin{tabularx}{0.8\textwidth}{m{1.5cm} cYYYY}
\toprule
Metric & $L_\text{reg}$ & \texttt{fr1/desk} & \texttt{fr2/xyz} & \texttt{fr3/office} & Average \\
\midrule
\multirow{2}{*}[-0em]{\makecell[l]{ATE $\downarrow$}}
& \redx       &2.7  &22.4  &4.7  &9.9 \\
& \greencheck &\fs2.6  &\fs1.3  &\fs4.6  &\fs2.9 \\
\midrule
\multirow{2}{*}[-0em]{\makecell[l]{PSNR $\uparrow$}}
& \redx       &\fs24.50  &23.03  &\fs26.60  &24.71 \\
& \greencheck &24.01  &\fs25.02  &26.13  &\fs25.05 \\

\bottomrule
\end{tabularx}

\caption{\textbf{Isotropic Regularization Ablation on TUM-RGBD dataset~\cite{stuhmer2010real}.} The regularization term $L_\text{reg}$ is critical for tracking accuracy and improves the rendering performance.}
\label{tab:iso_reg_ablation_tum}
\end{table}

\begin{figure*}[!htb] \centering
    \newcommand{\wratio}{0.24}
    \setlength{\tabcolsep}{0.5pt}
    \renewcommand{\arraystretch}{1}
    \begin{tabular}{lllll}
    & \multicolumn{1}{c}{ESLAM~\cite{mahdi2022eslam}}
    & \multicolumn{1}{c}{Point-SLAM~\cite{sandstrom2023point}}
    & \multicolumn{1}{c}{\ours}
    & \multicolumn{1}{c}{Ground Truth}
    \\
        \rotatebox[origin=c]{90}{\texttt{room 0}} & 
        \includegraphics[valign=c,width=\wratio\textwidth]{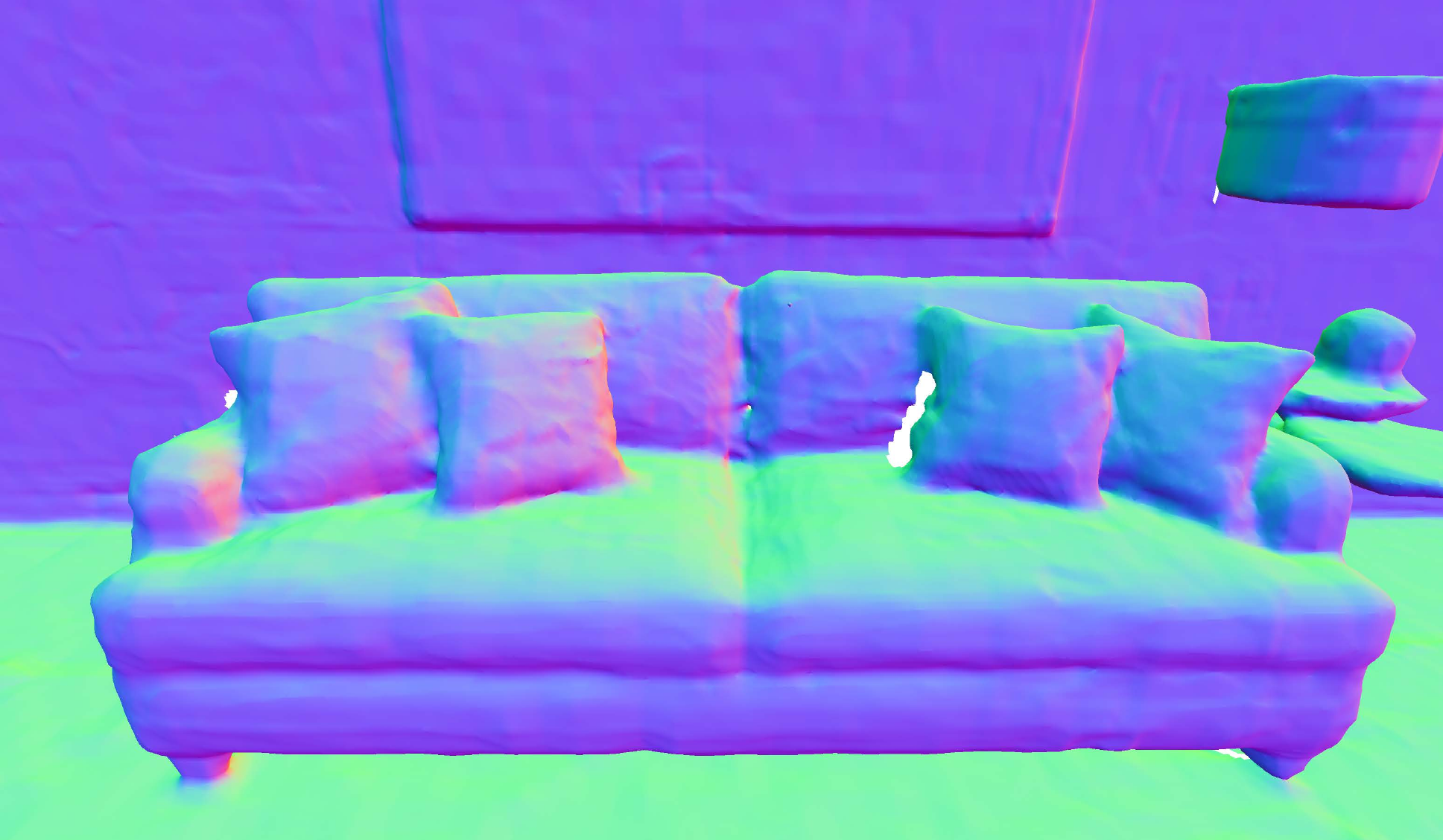} &
        \includegraphics[valign=c,width=\wratio\textwidth]{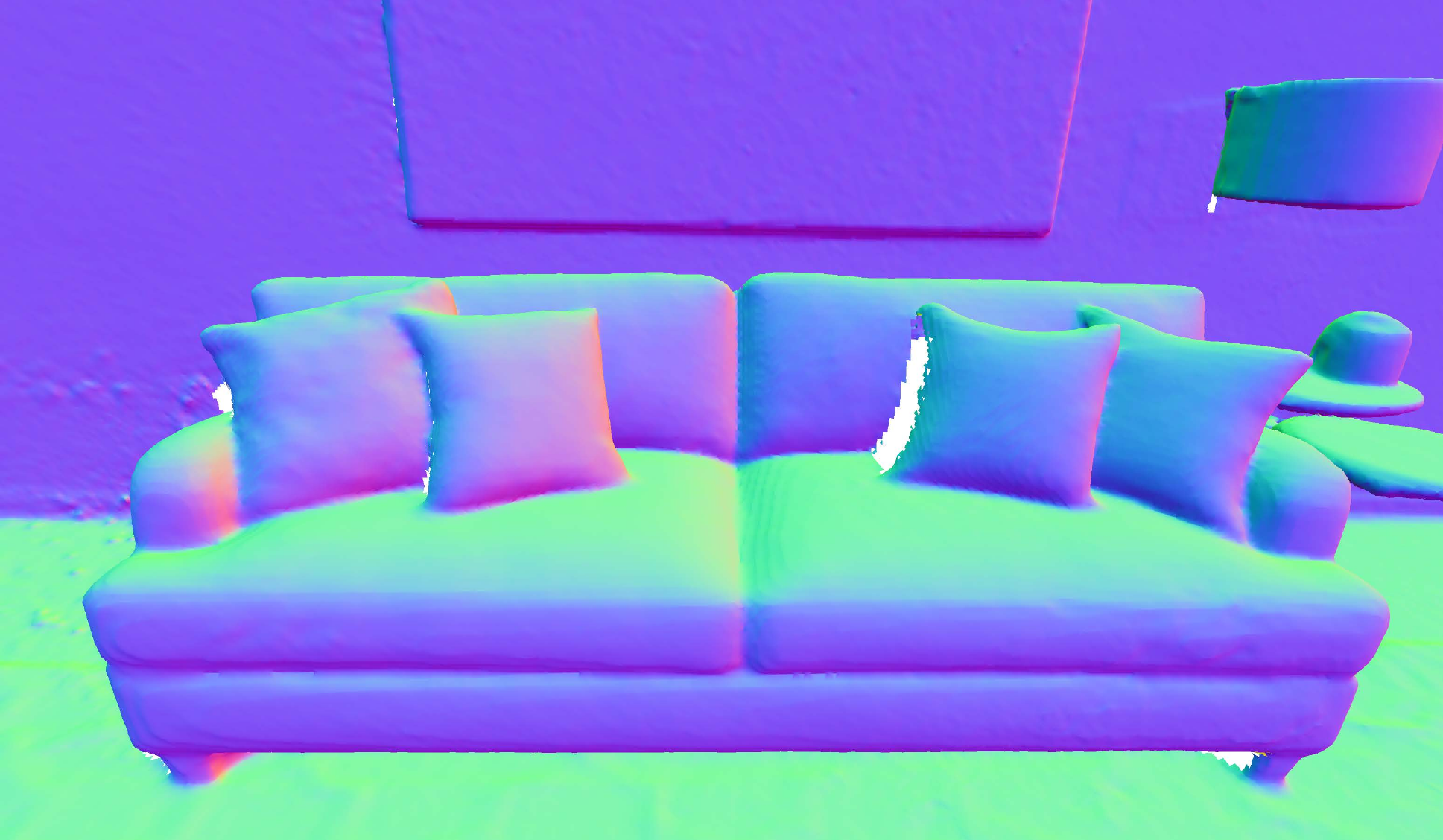} &
        \includegraphics[valign=c,width=\wratio\textwidth]{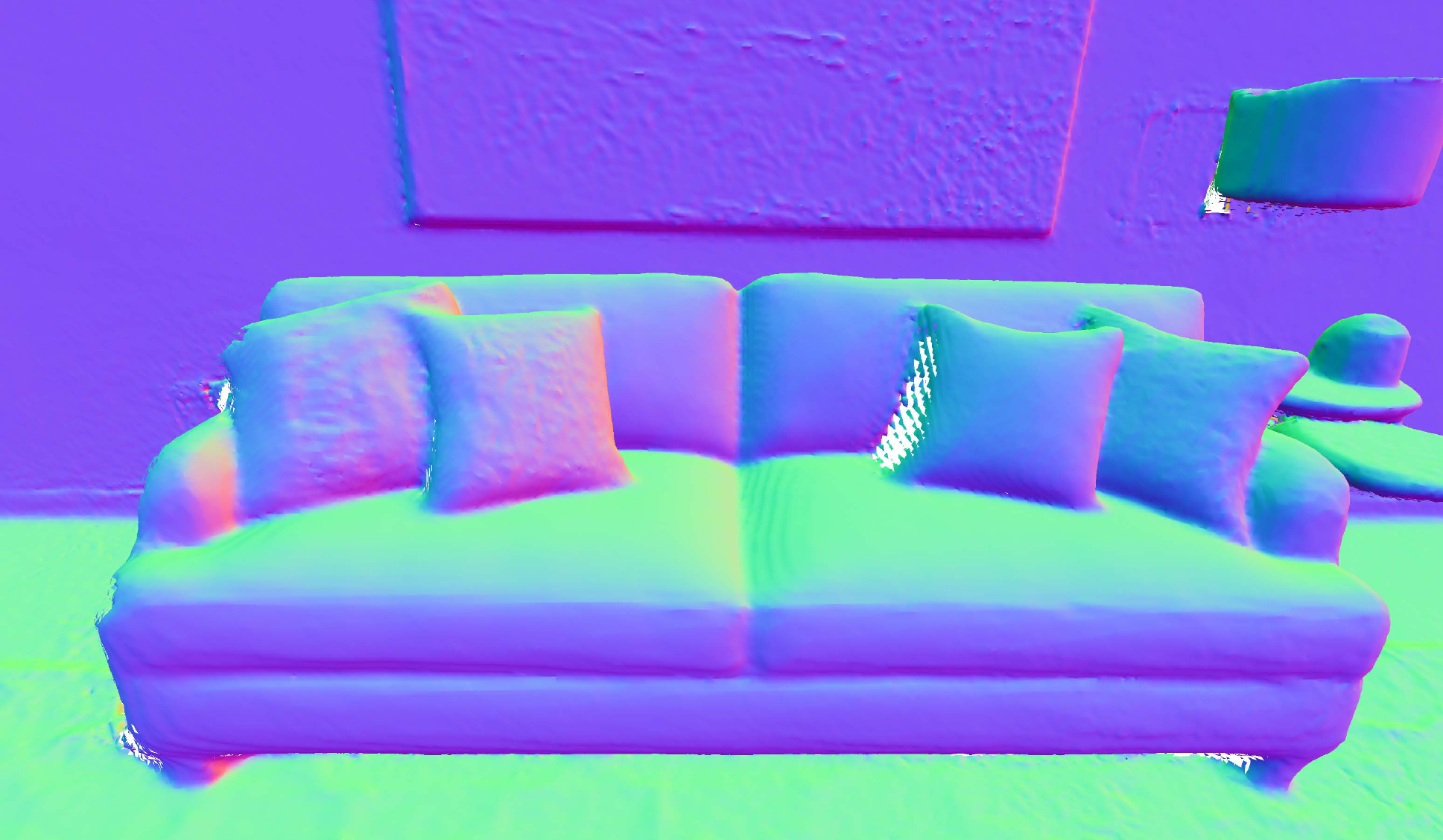} &
        \includegraphics[valign=c,width=\wratio\textwidth]{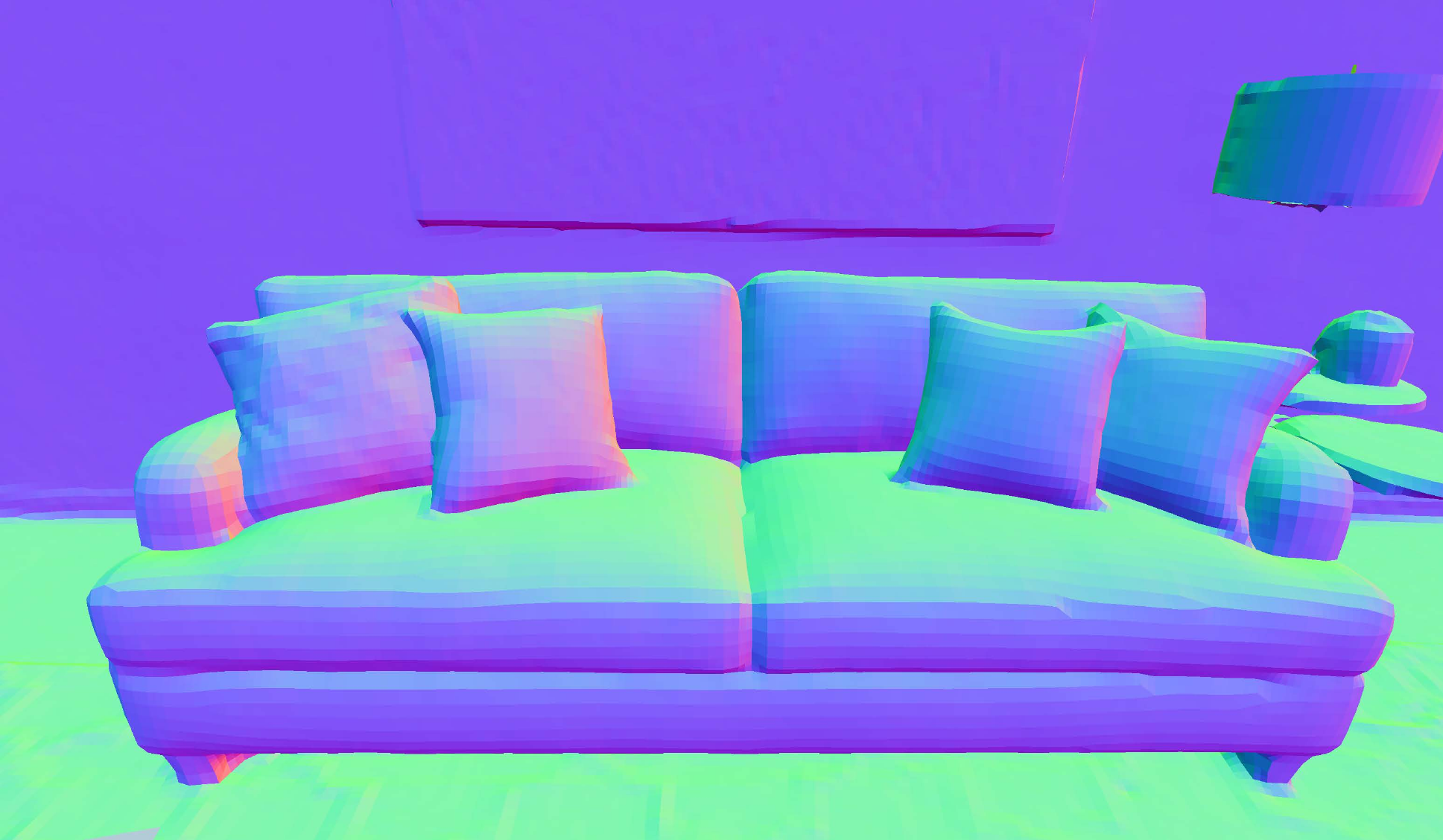} 
        \\[22pt]
        \rotatebox[origin=c]{90}{\texttt{room 1}} & 
        \includegraphics[valign=c,width=\wratio\textwidth]{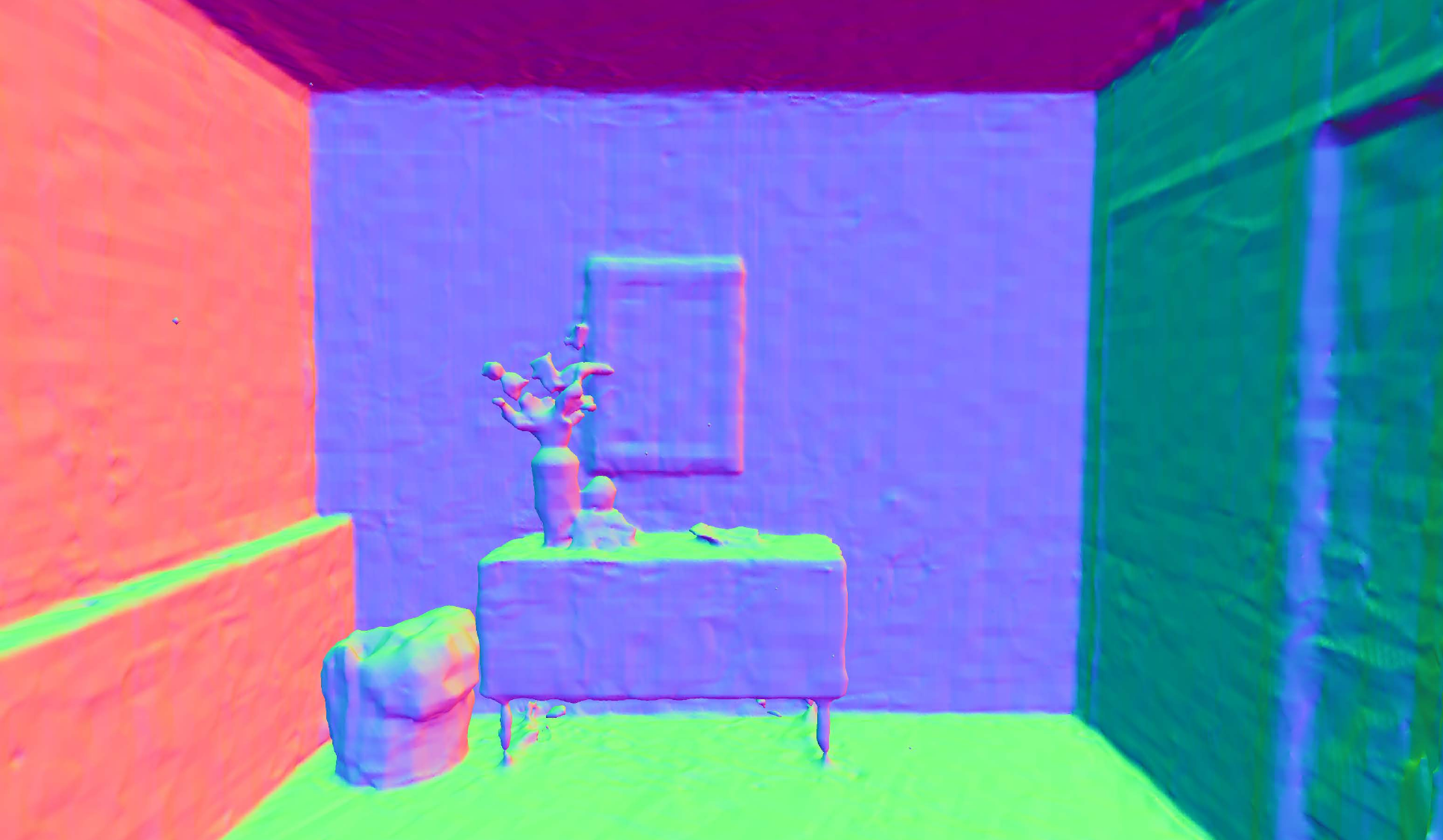} &
        \includegraphics[valign=c,width=\wratio\textwidth]{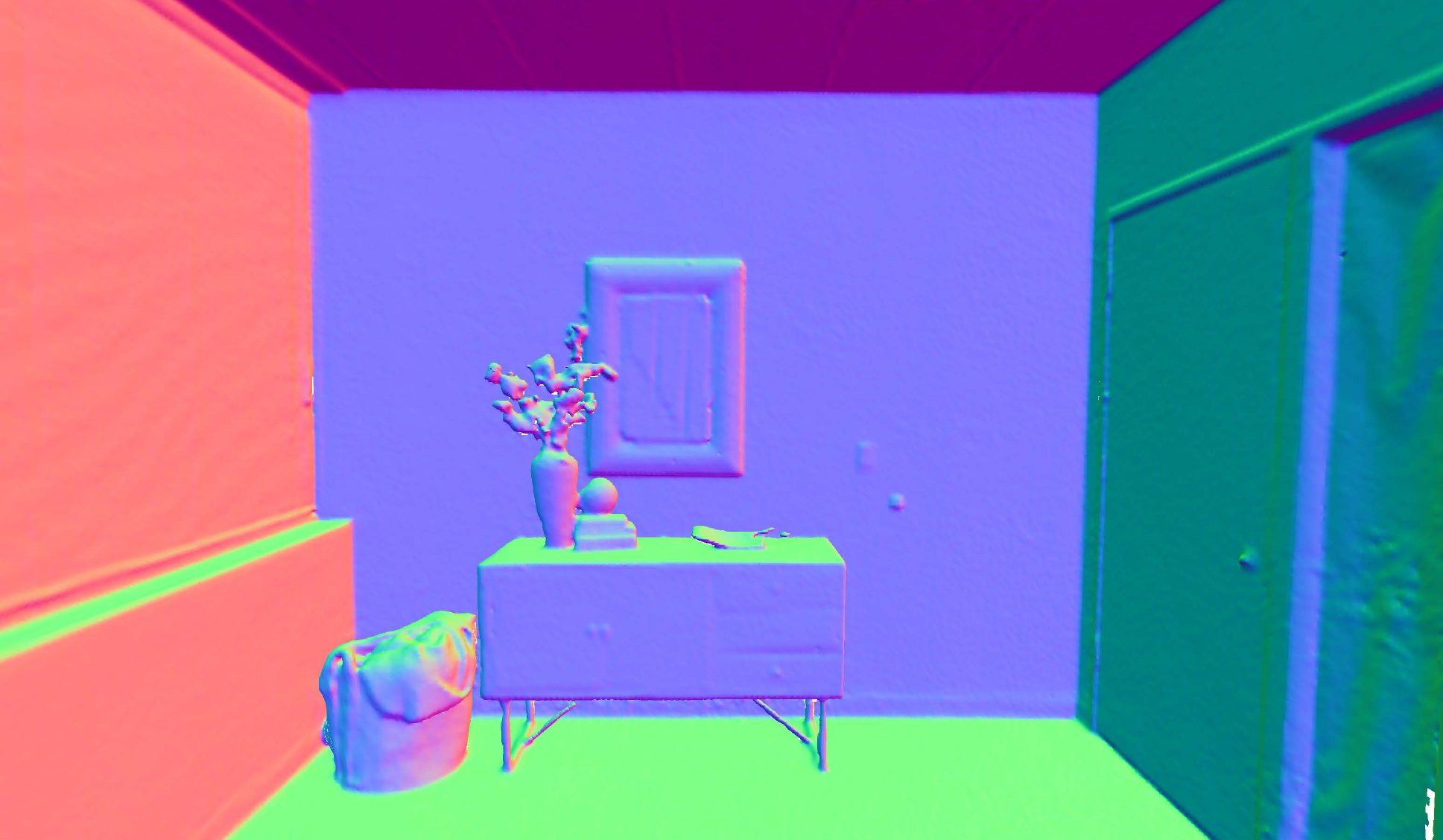} &
        \includegraphics[valign=c,width=\wratio\textwidth]{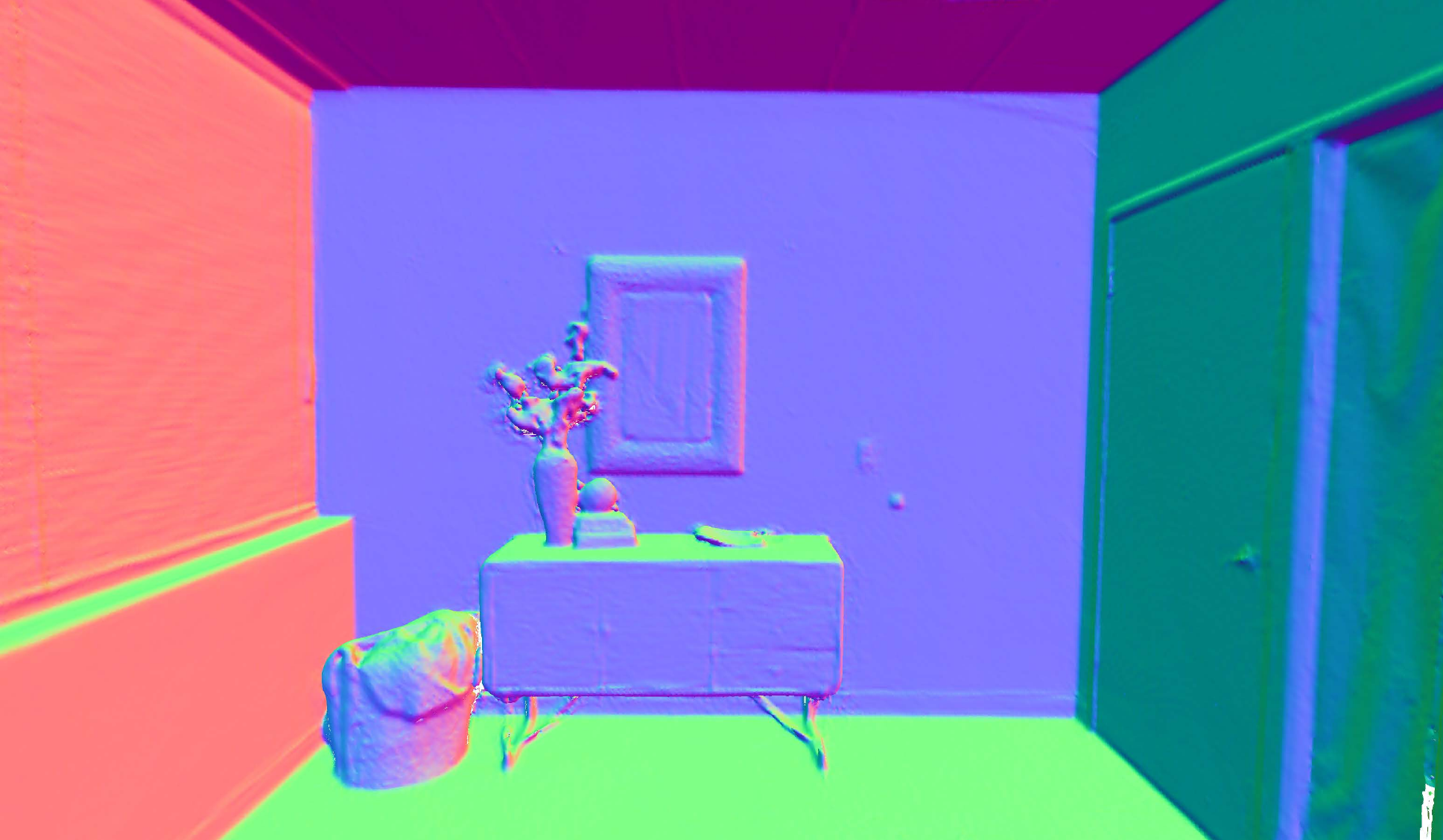} &
        \includegraphics[valign=c,width=\wratio\textwidth]{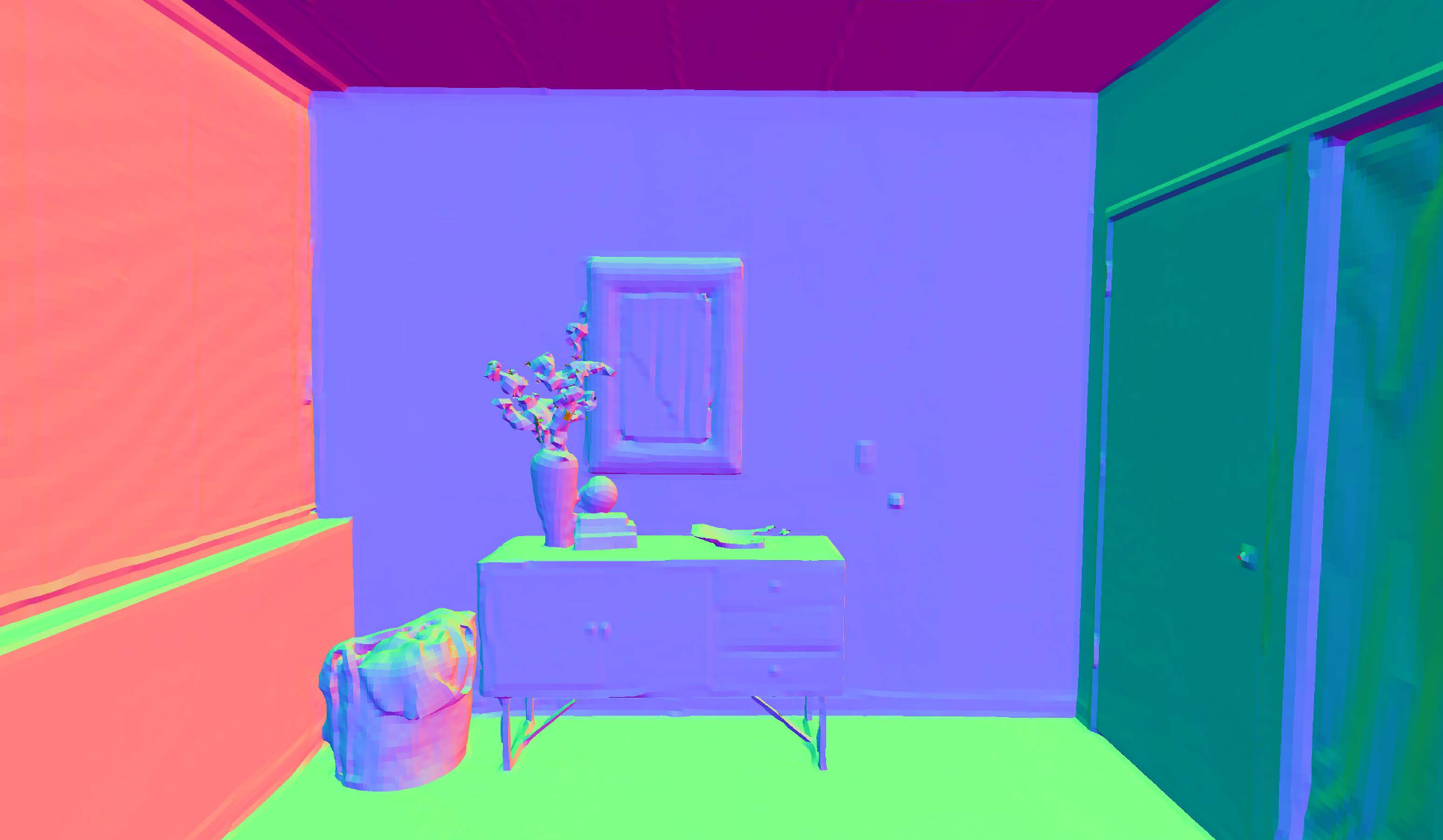}
        \\[22pt]
        \rotatebox[origin=c]{90}{\texttt{office 0}} & 
        \includegraphics[valign=c,width=\wratio\textwidth]{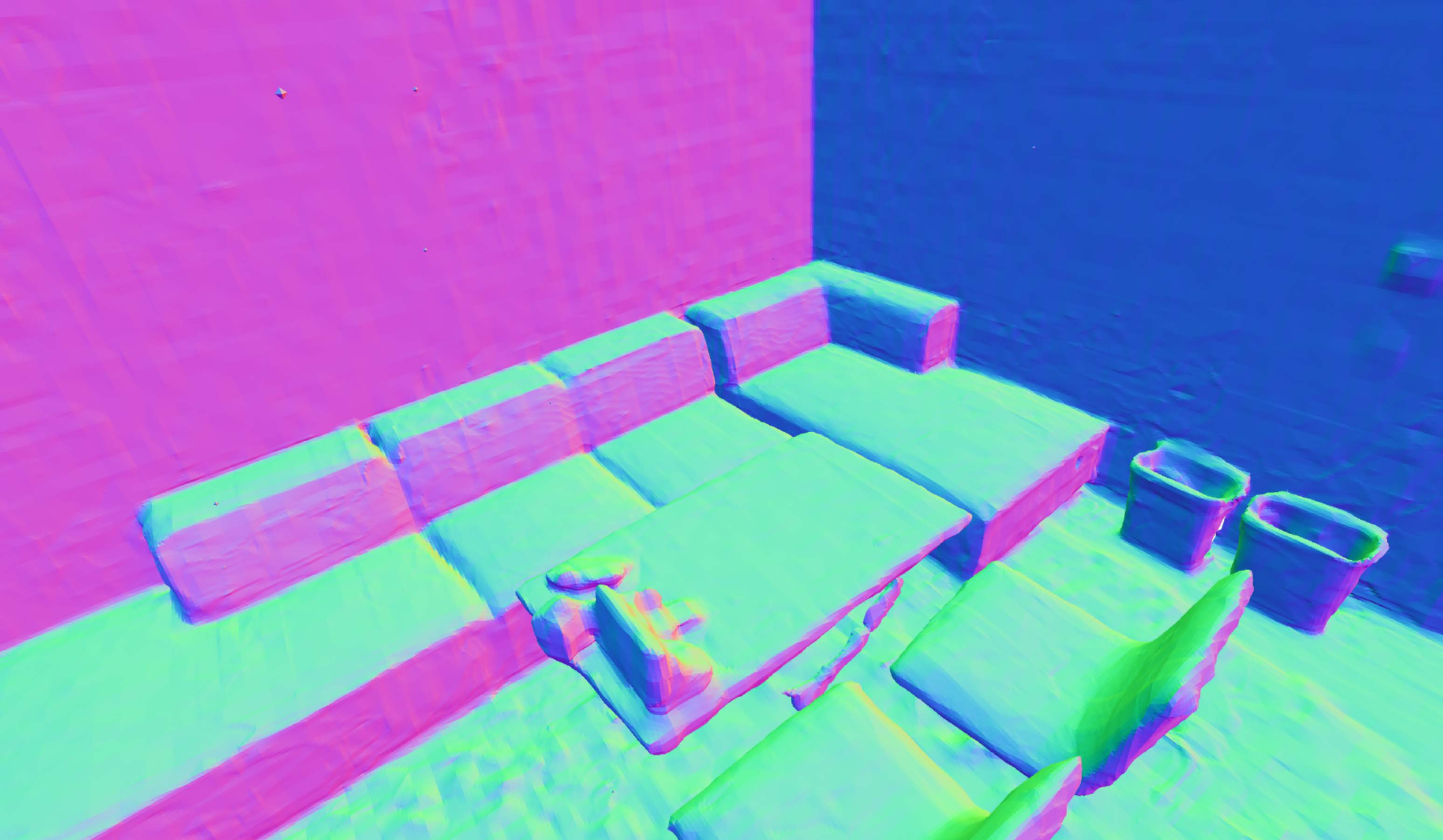} &
        \includegraphics[valign=c,width=\wratio\textwidth]{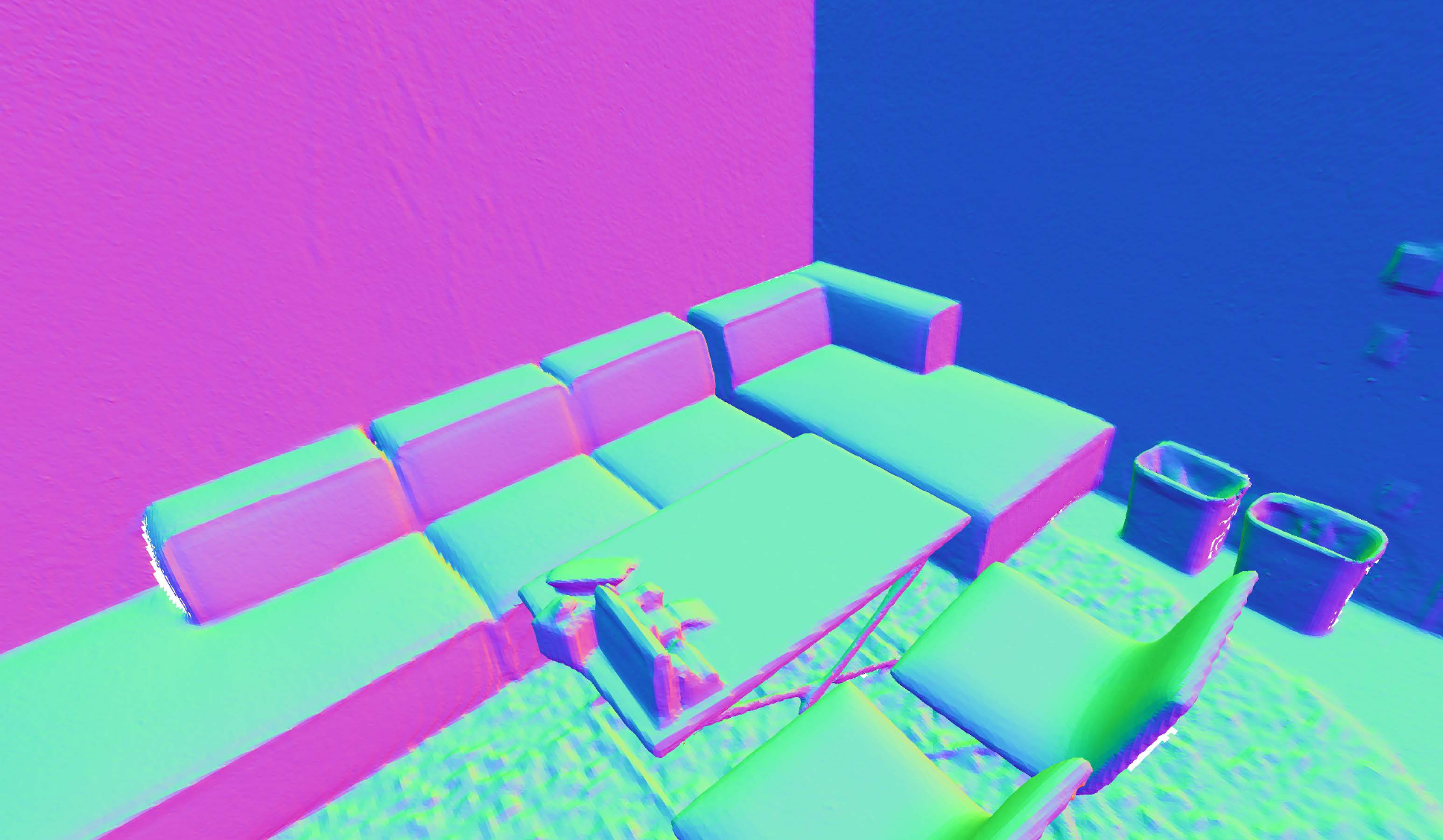} &
        \includegraphics[valign=c,width=\wratio\textwidth]{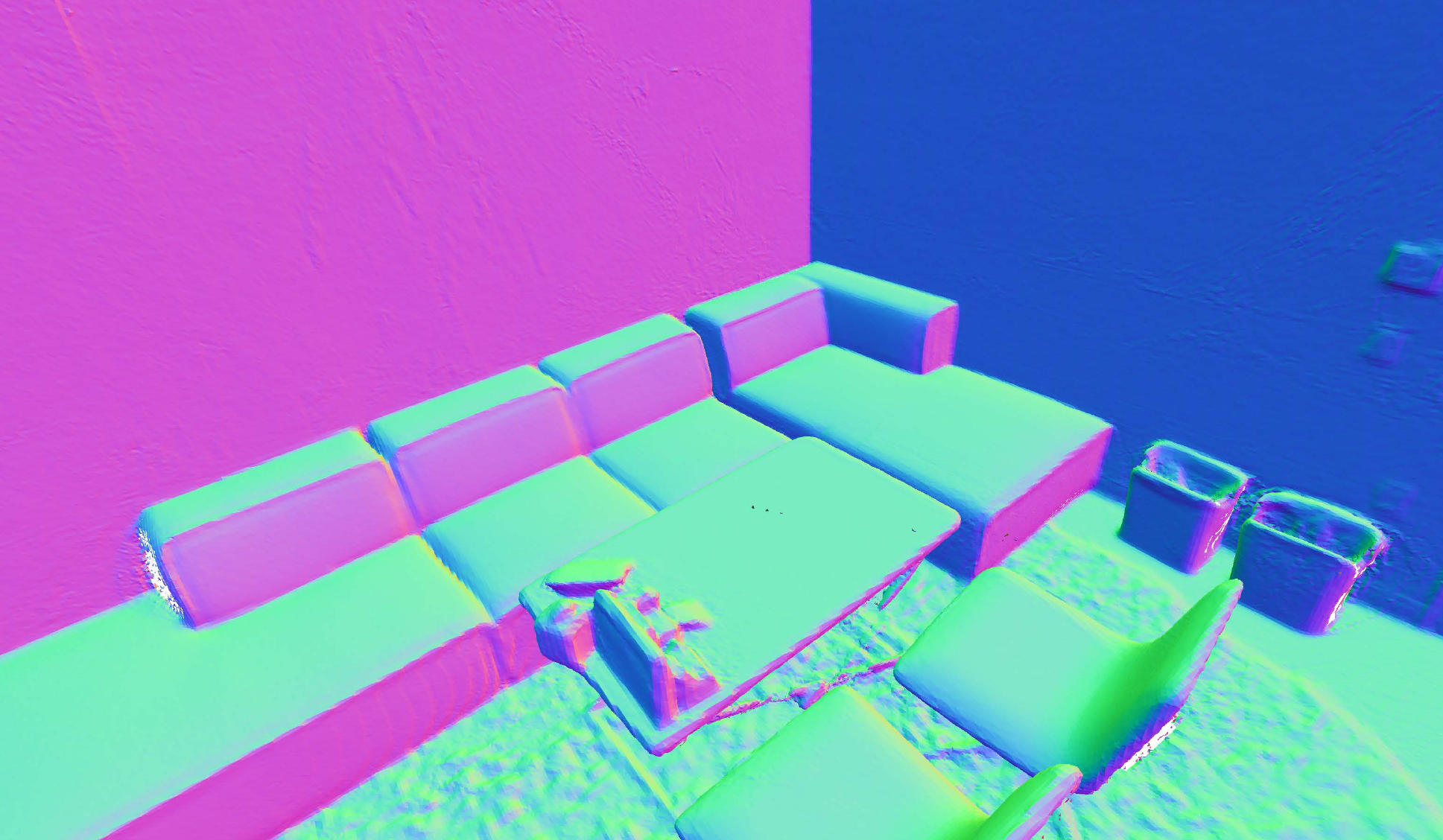} &
        \includegraphics[valign=c,width=\wratio\textwidth]{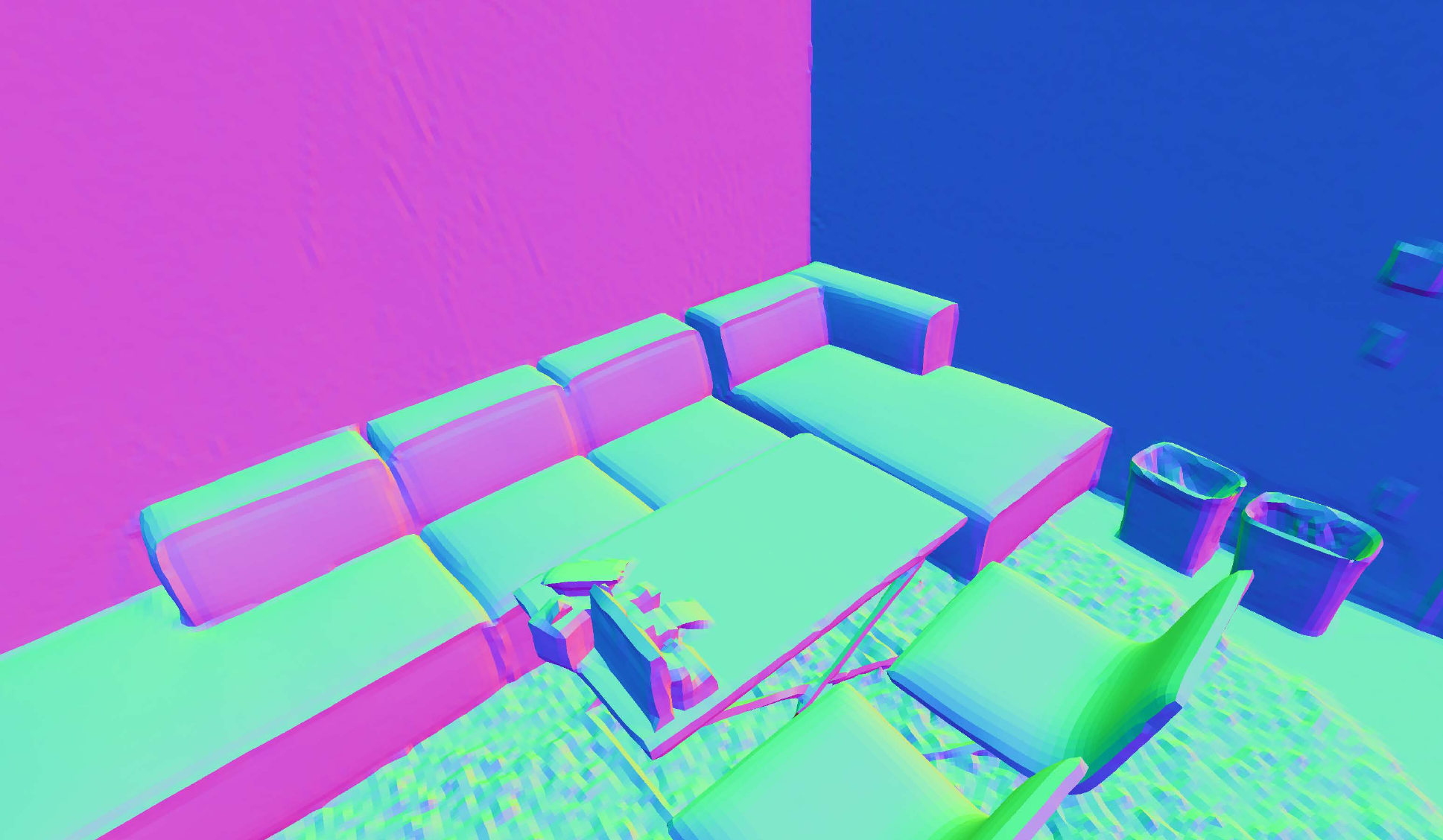} 
        \\[22pt]
        \rotatebox[origin=c]{90}{\texttt{office 4}} & 
        \includegraphics[valign=c,width=\wratio\textwidth]{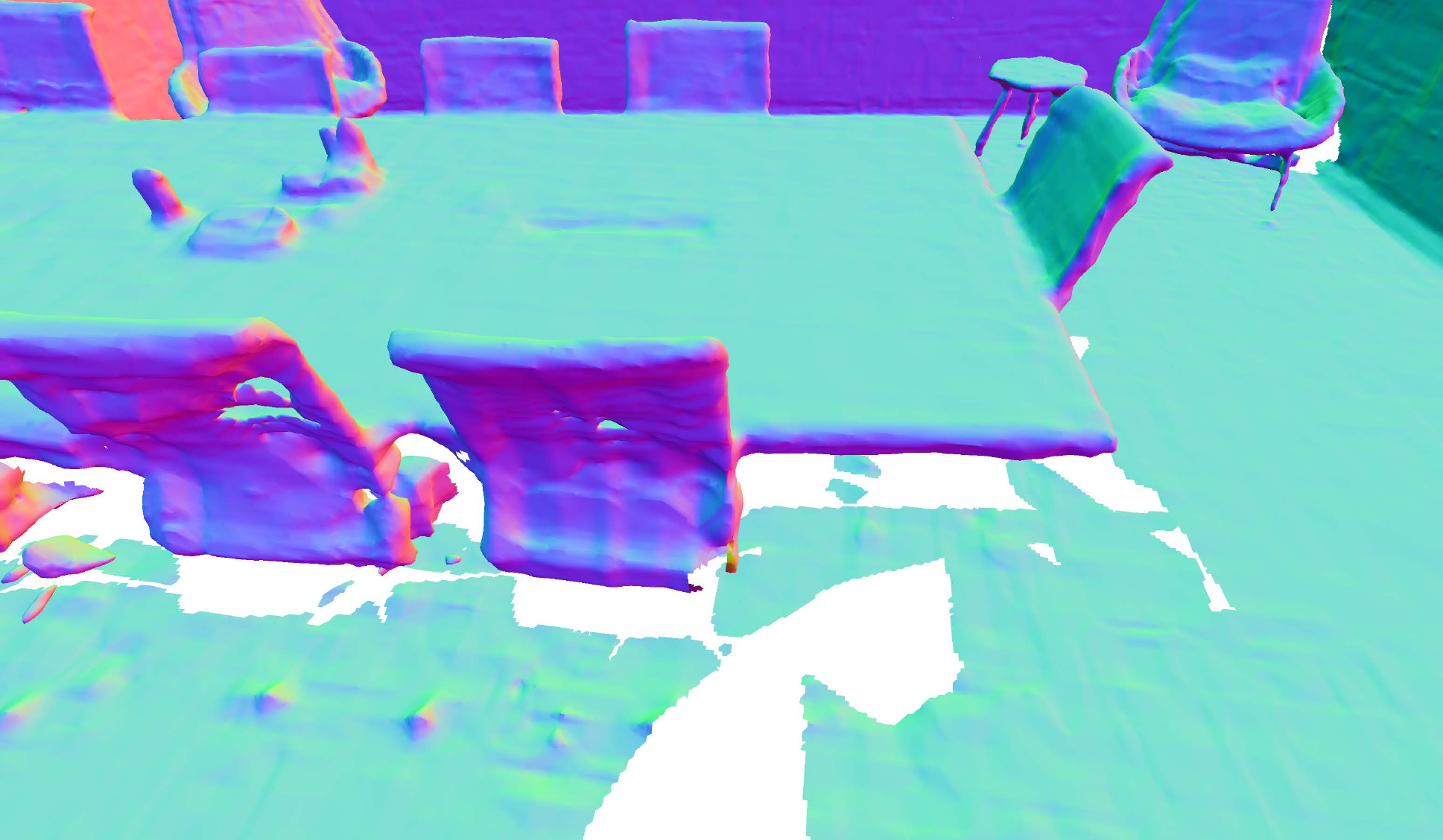} &
        \includegraphics[valign=c,width=\wratio\textwidth]{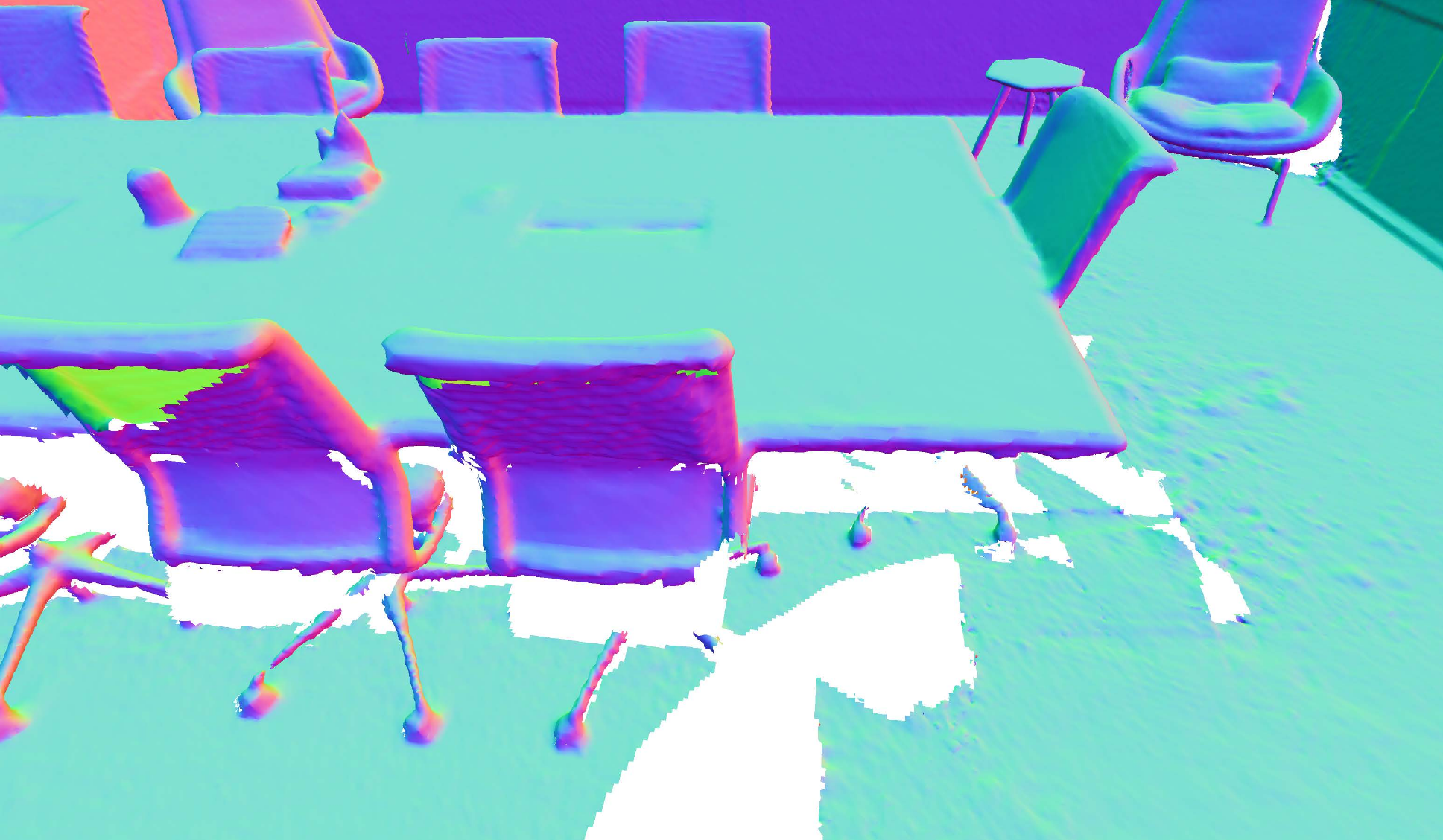} &
        \includegraphics[valign=c,width=\wratio\textwidth]{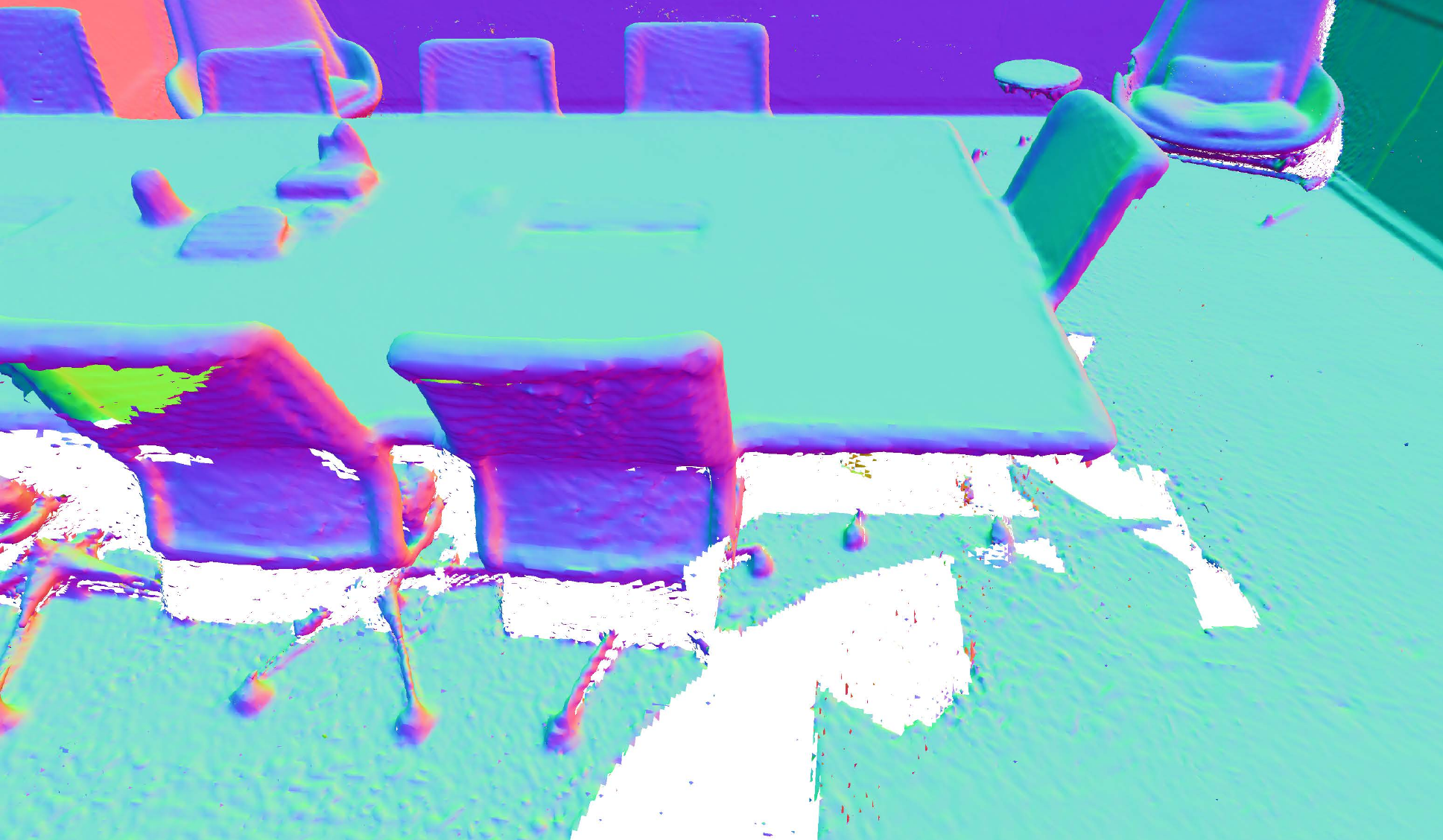} &
        \includegraphics[valign=c,width=\wratio\textwidth]{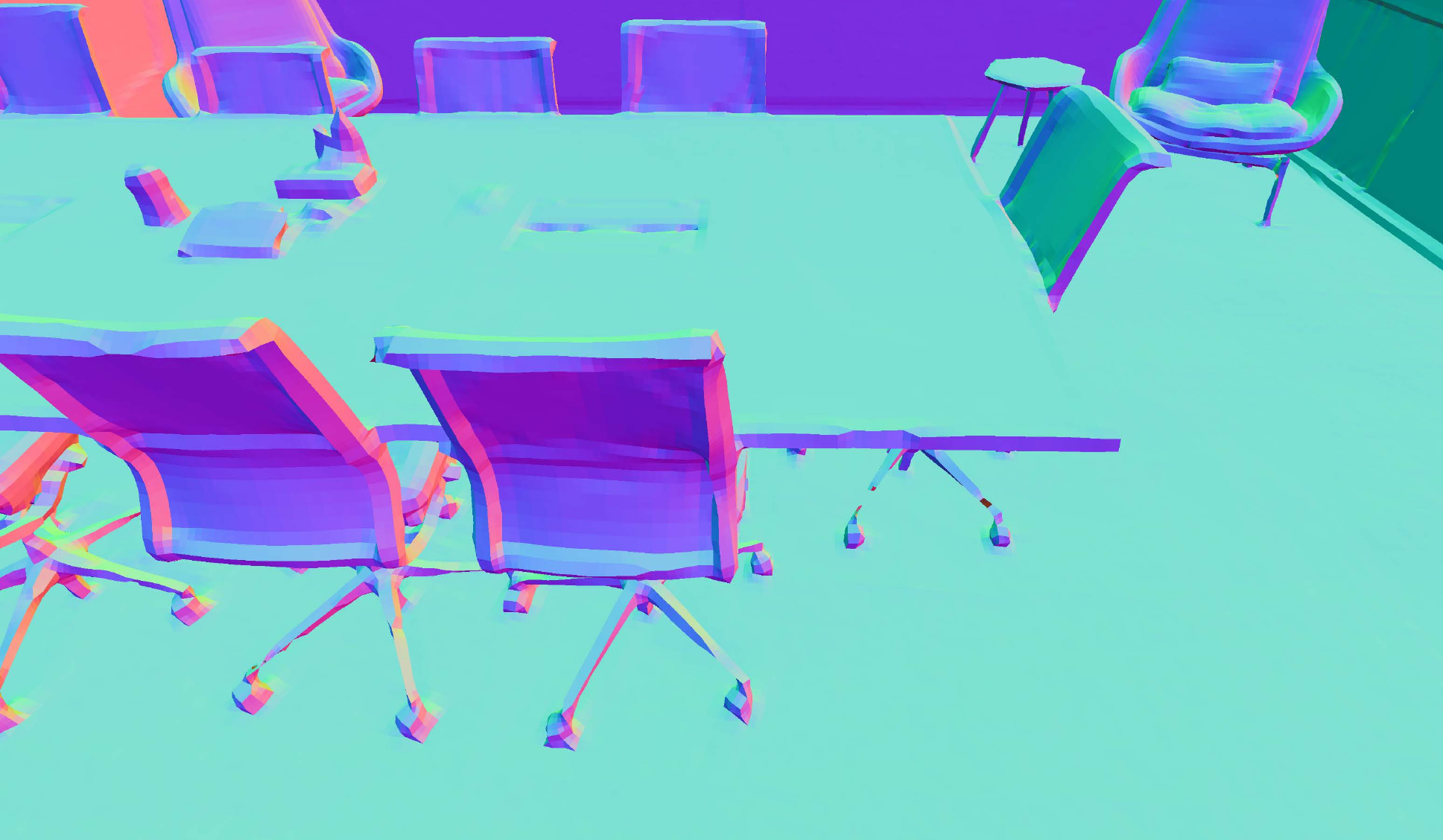} 
        \\
    \end{tabular}
    \caption{\textbf{Qualitative Reconstruction Comparison on the Replica dataset~\cite{straub2019replica}}. Results are rendered with a normal map shader. \ours achieves comparable reconstruction performance with the state-of-the-art dense neural SLAM methods.}
\label{fig:replica_mesh}
\end{figure*}

\begin{figure*}[!htb] \centering
    \newcommand{\wratio}{0.24}
    \setlength{\tabcolsep}{0.5pt}
    \renewcommand{\arraystretch}{1}
    \begin{tabular}{lllll}
    & \multicolumn{1}{c}{ESLAM~\cite{mahdi2022eslam}}
    & \multicolumn{1}{c}{Point-SLAM~\cite{sandstrom2023point}}
    & \multicolumn{1}{c}{\ours}
    & \multicolumn{1}{c}{Ground Truth}
    \\
        \rotatebox[origin=c]{90}{\texttt{0059}} & 
        \includegraphics[valign=c,width=\wratio\textwidth]{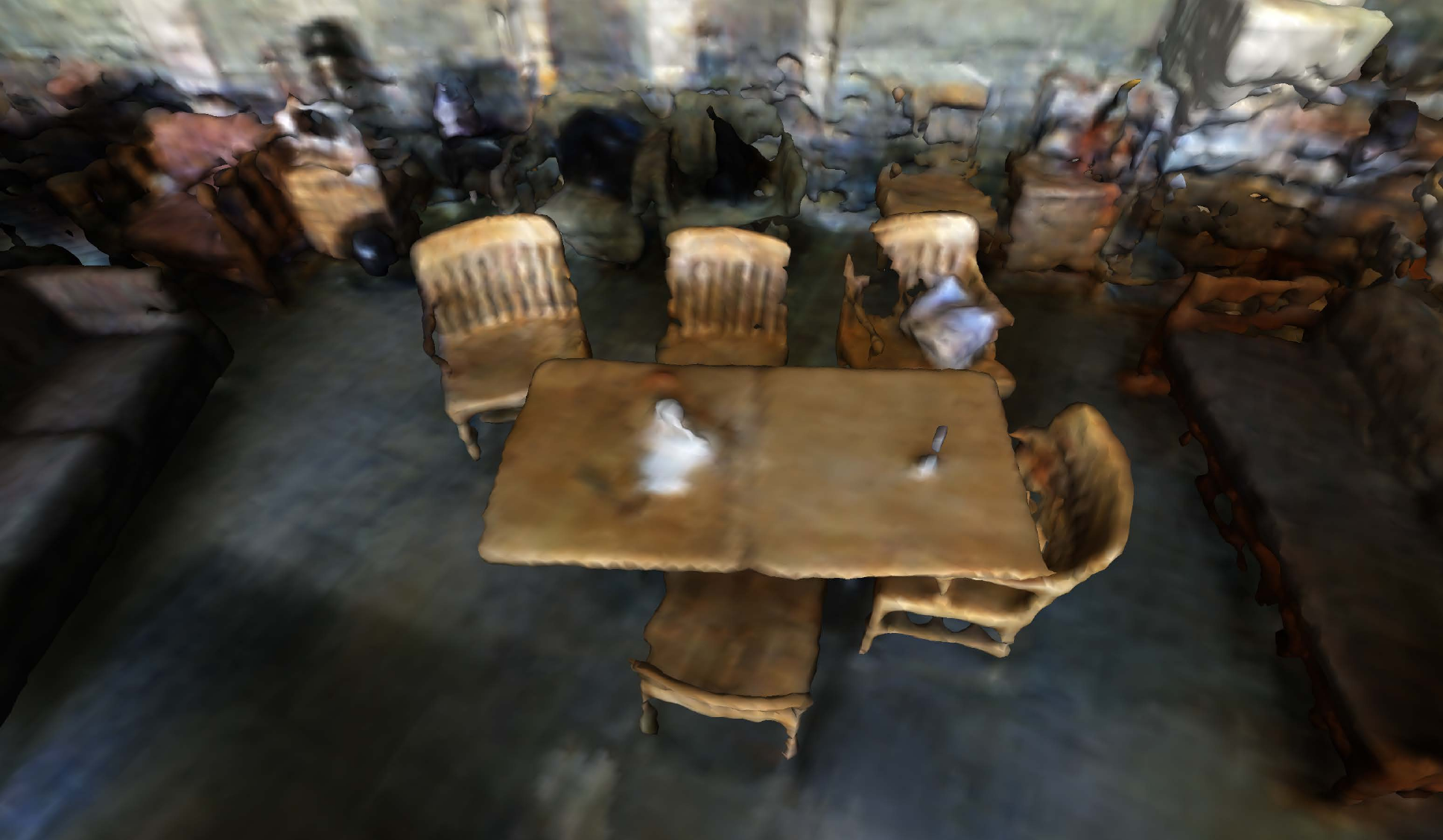} &
        \includegraphics[valign=c,width=\wratio\textwidth]{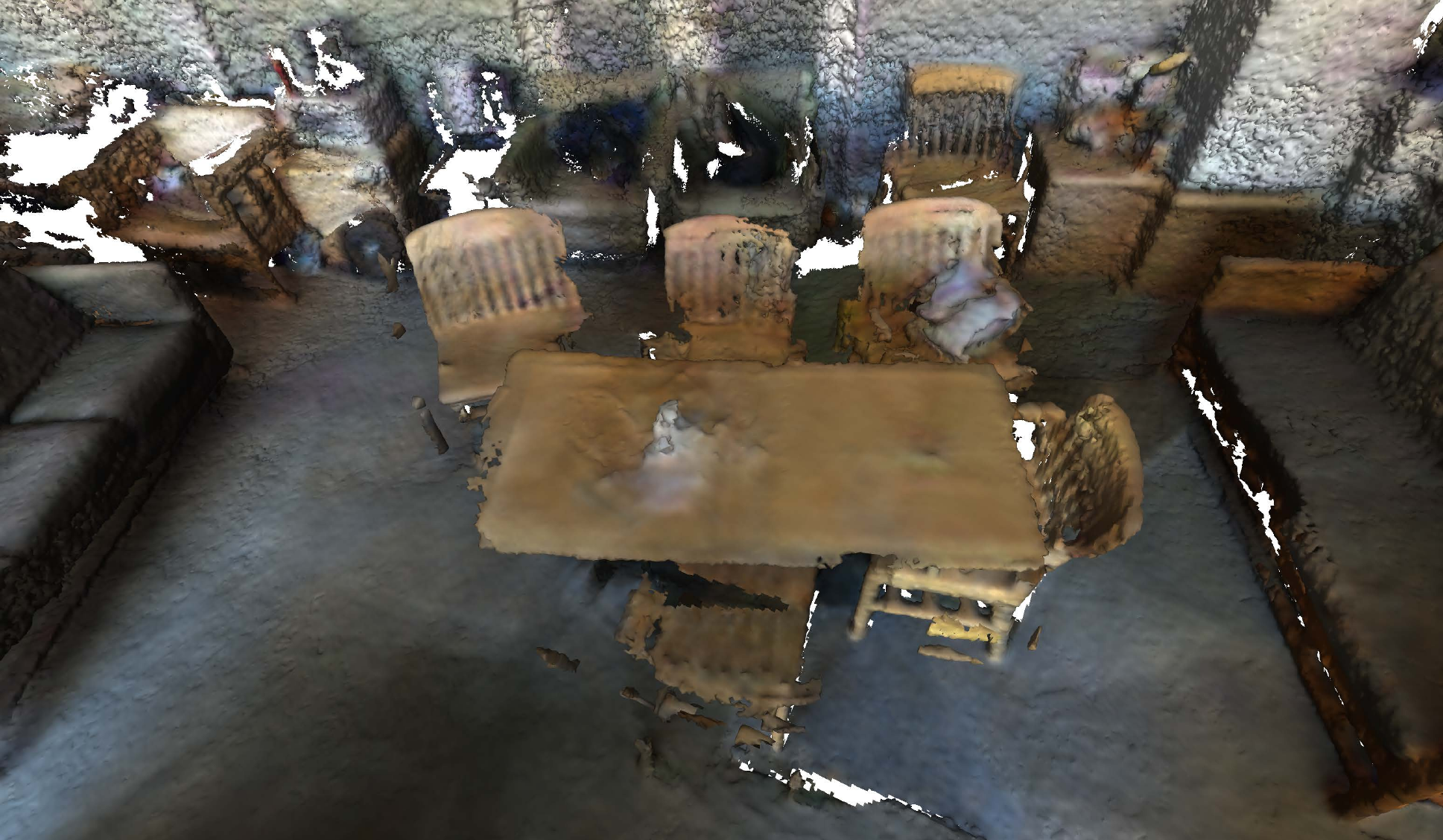} &
        \includegraphics[valign=c,width=\wratio\textwidth]{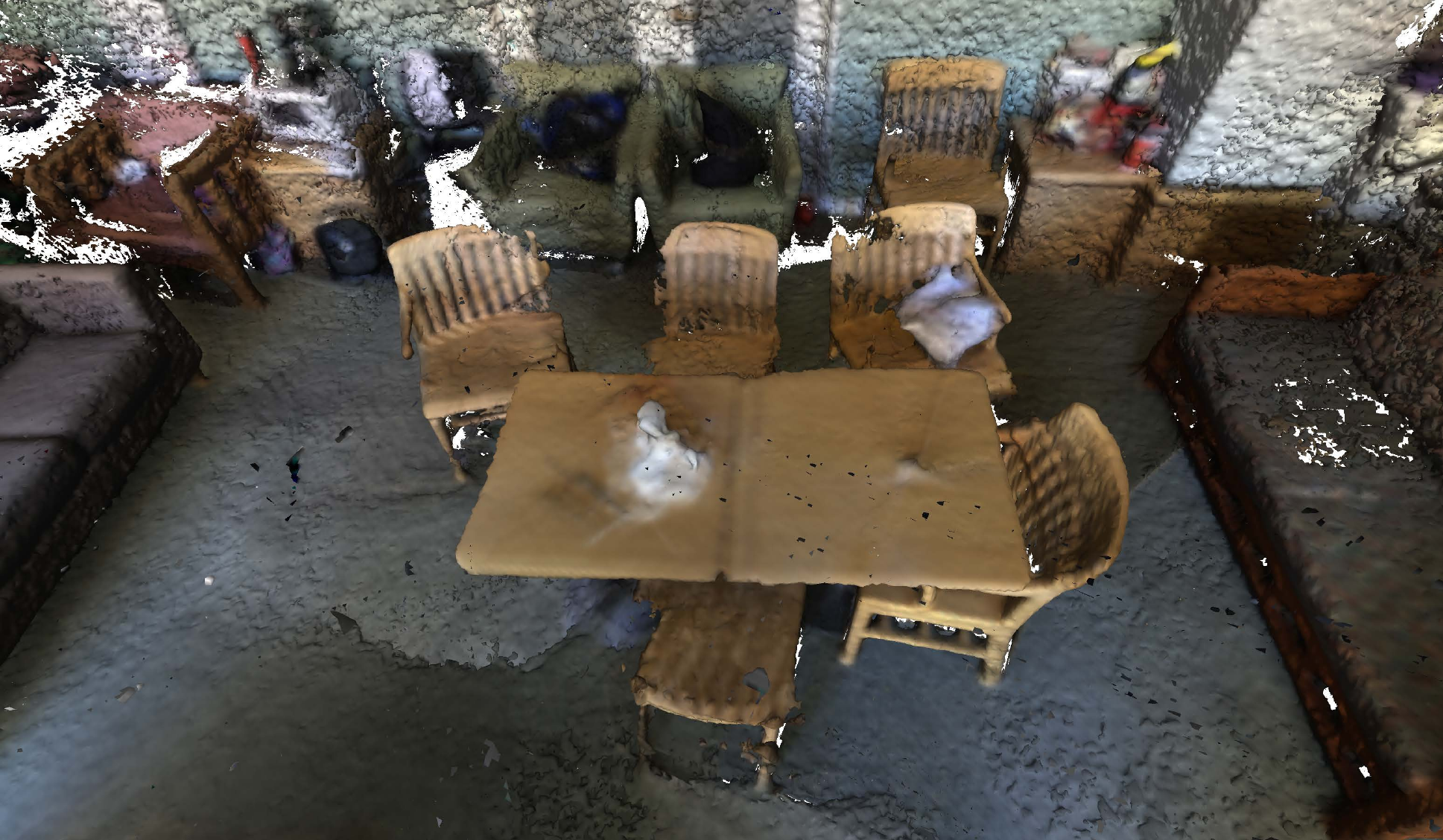} &
        \includegraphics[valign=c,width=\wratio\textwidth]{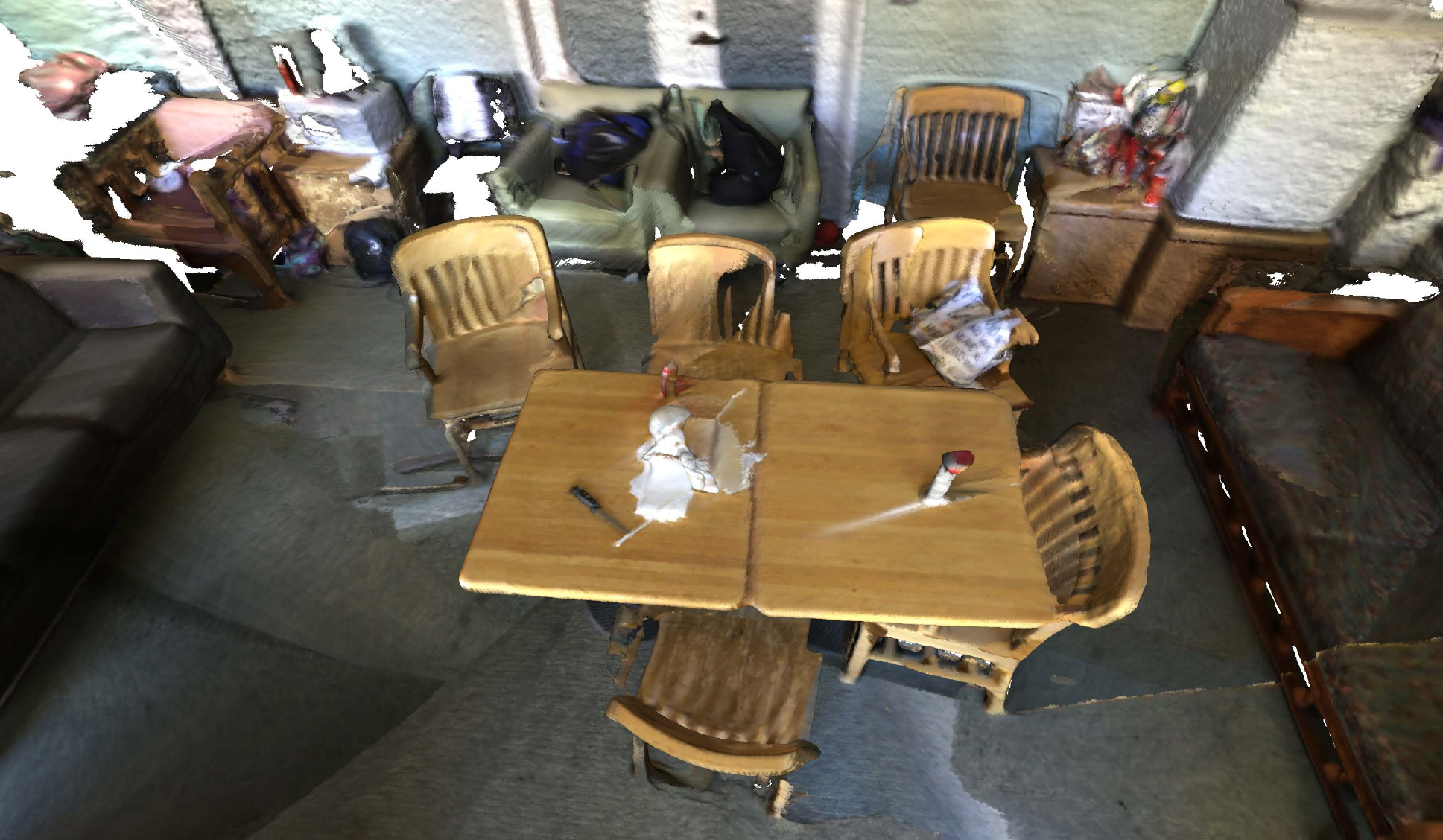} 
        \\[22pt] 
        \rotatebox[origin=c]{90}{\texttt{0169}} & 
        \includegraphics[valign=c,width=\wratio\textwidth]{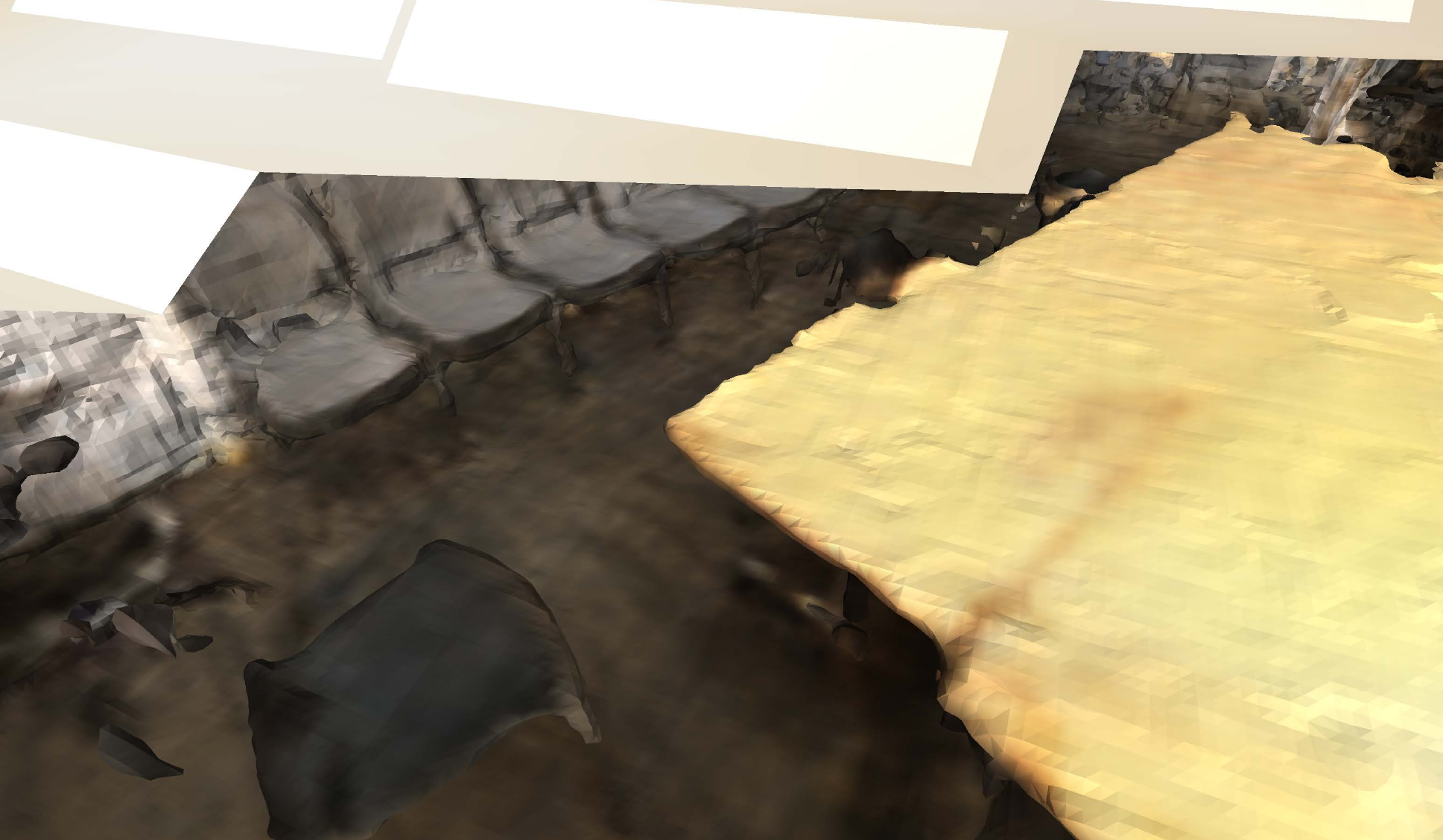} &
        \includegraphics[valign=c,width=\wratio\textwidth]{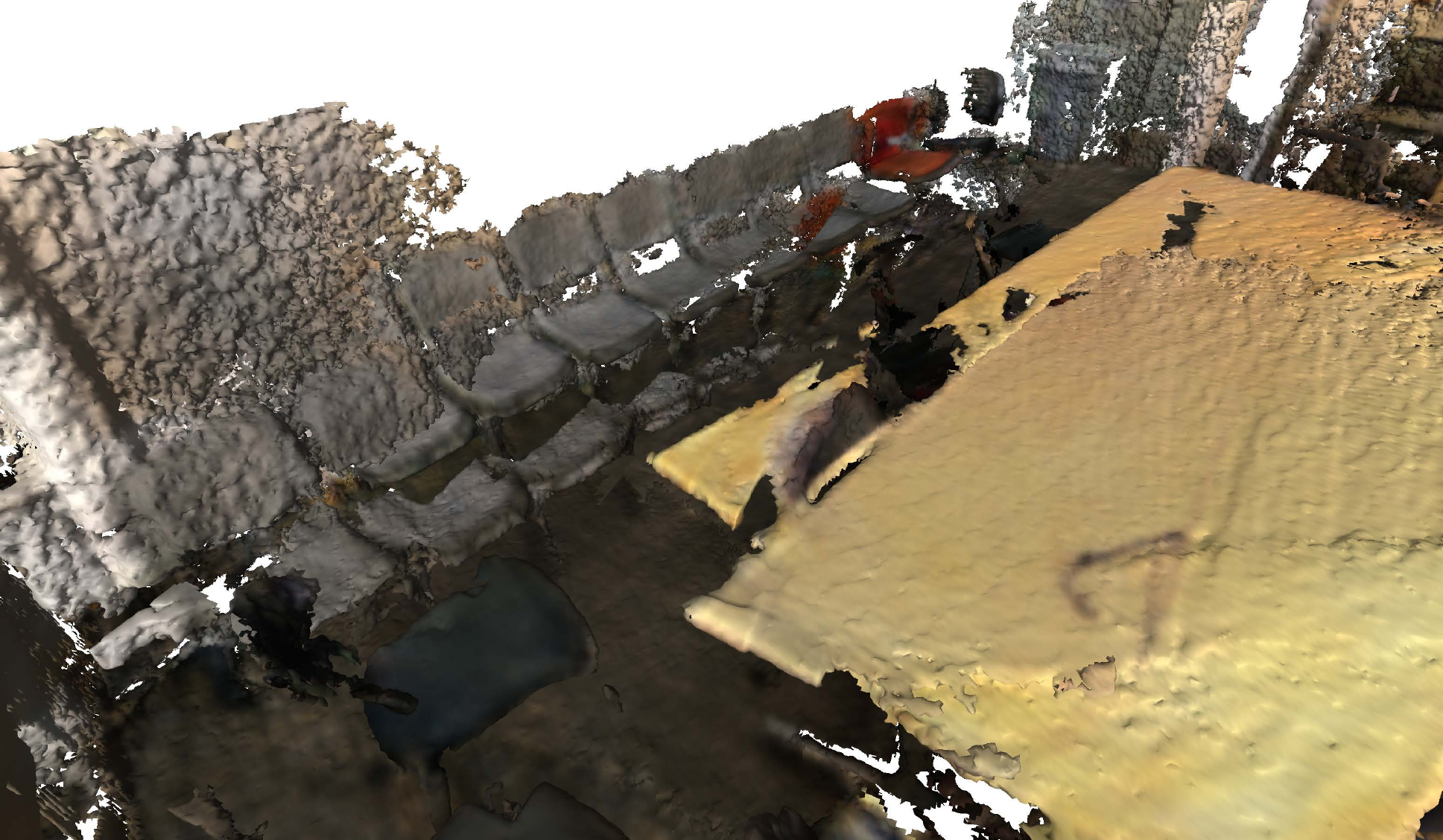} &
        \includegraphics[valign=c,width=\wratio\textwidth]{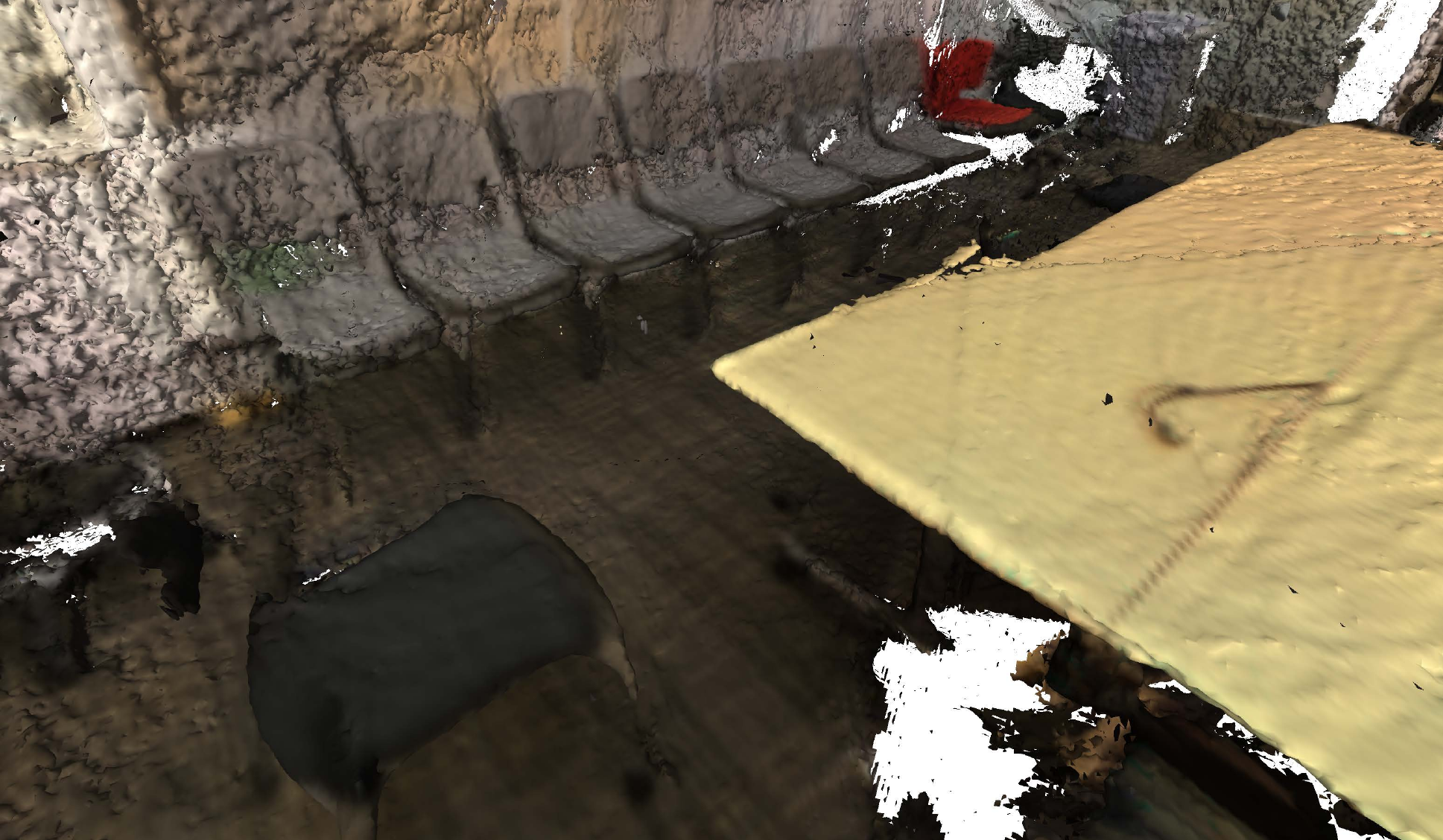} &
        \includegraphics[valign=c,width=\wratio\textwidth]{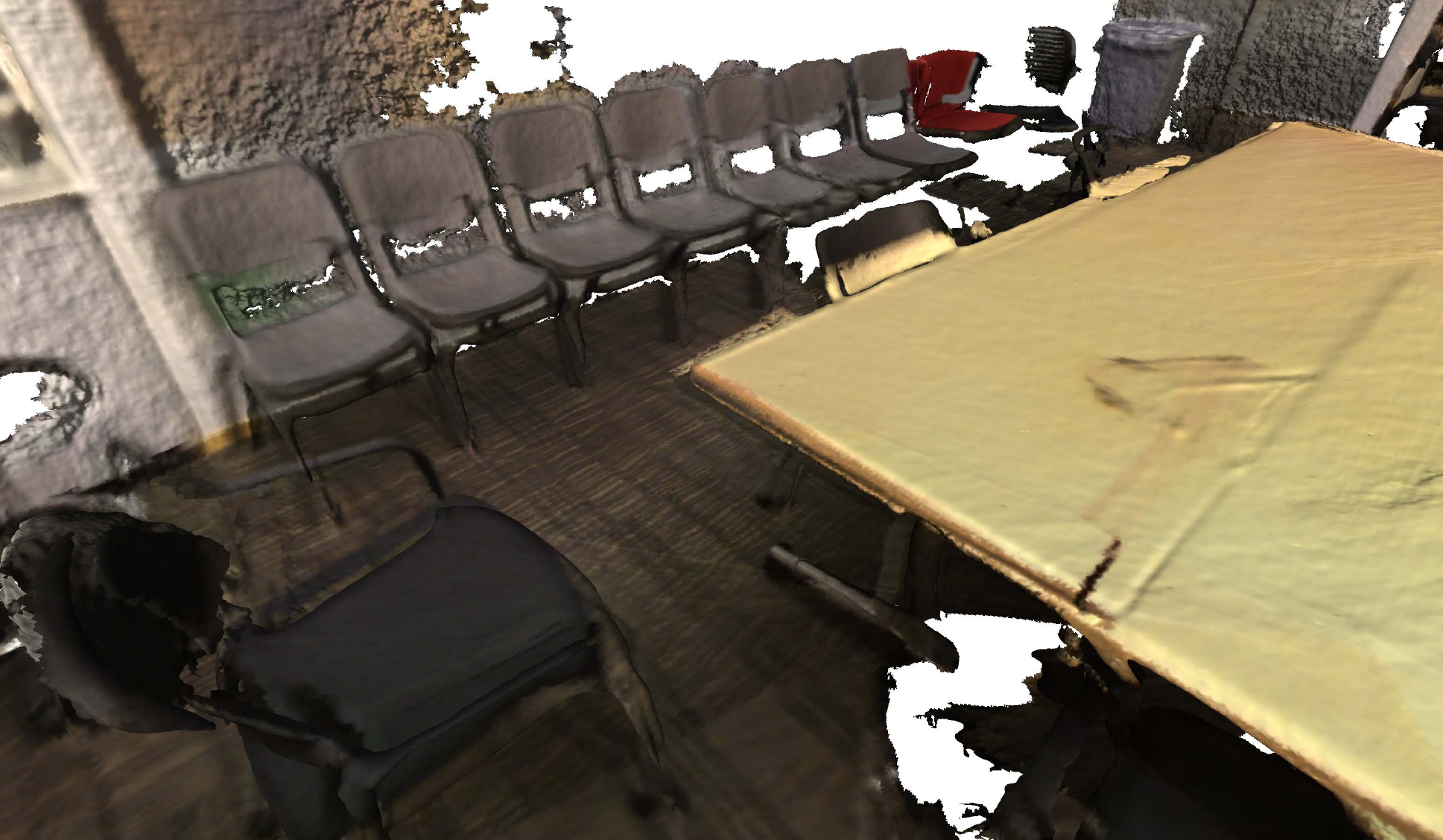}
        \\[22pt]
        \rotatebox[origin=c]{90}{\texttt{desk1}} & 
        \includegraphics[valign=c,width=\wratio\textwidth]{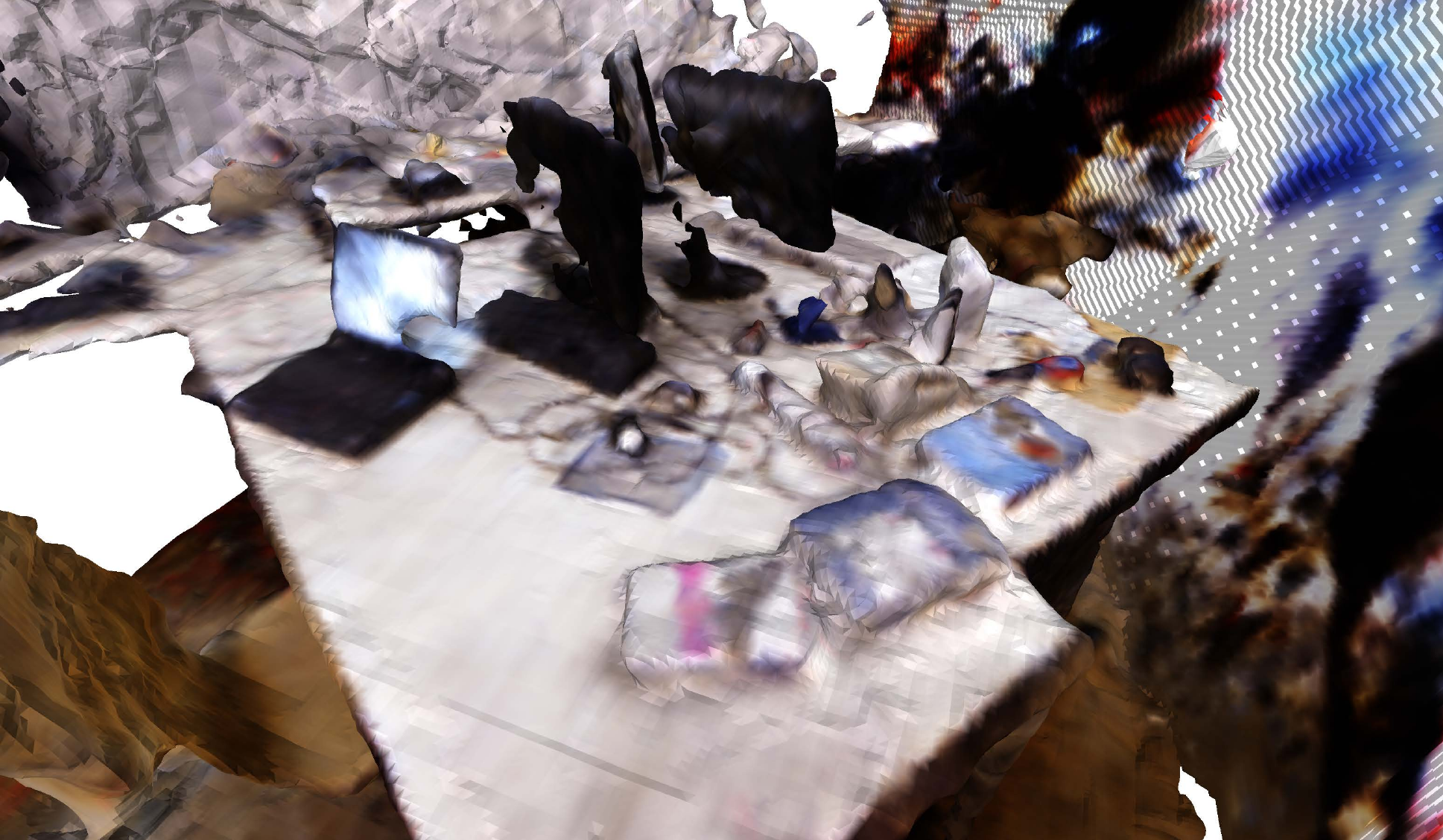} &
        \includegraphics[valign=c,width=\wratio\textwidth]{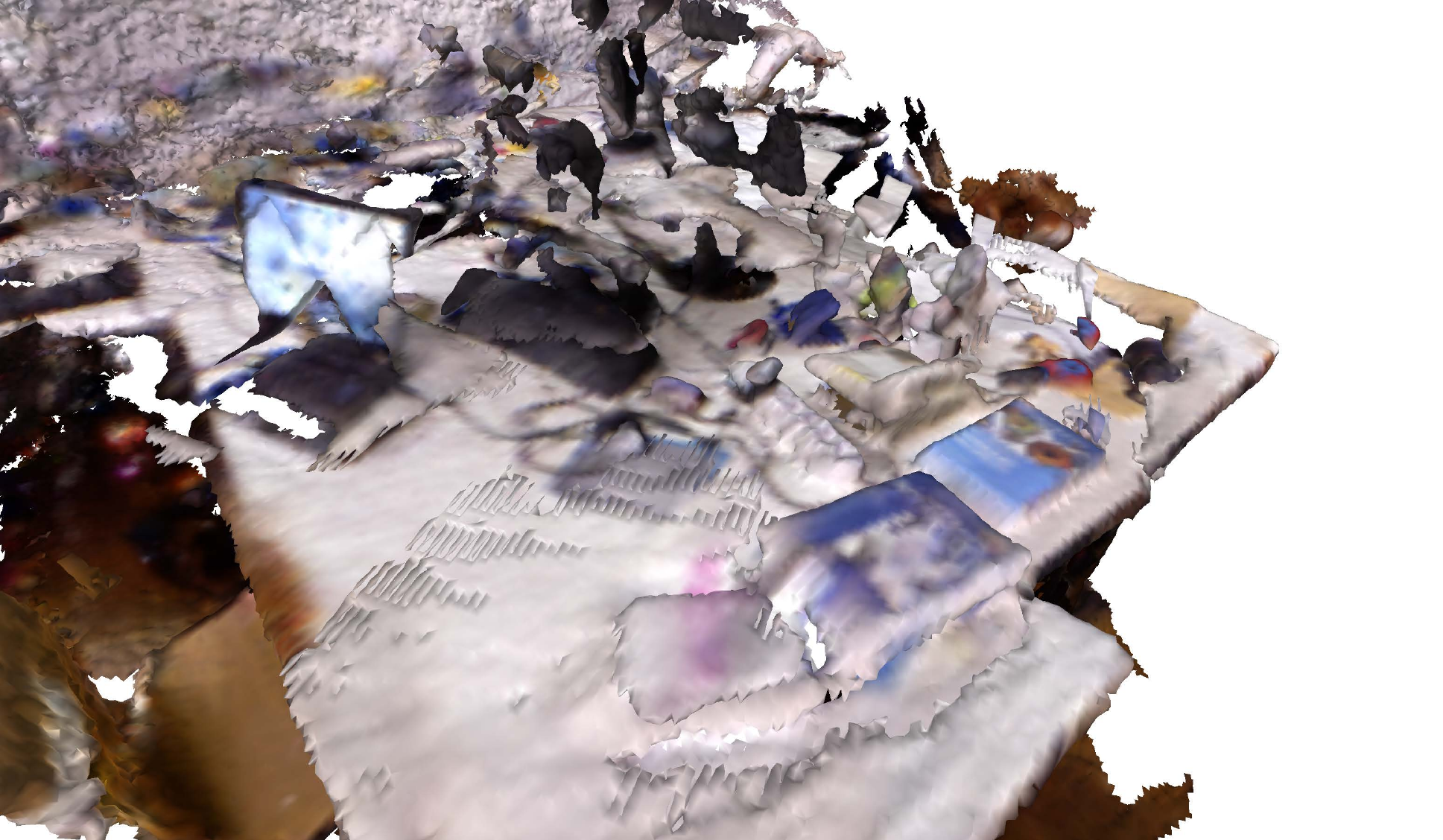} &
        \includegraphics[valign=c,width=\wratio\textwidth]{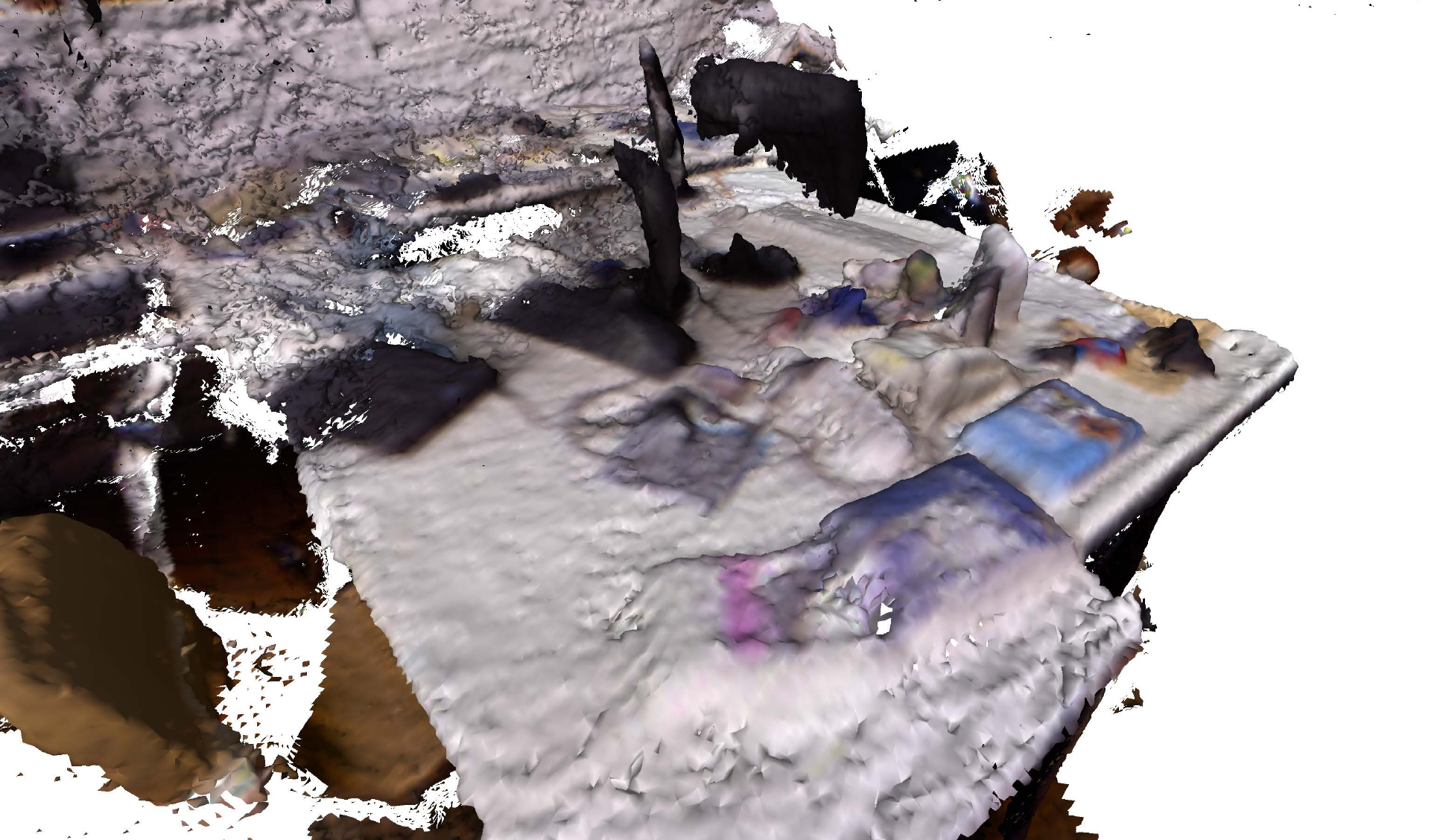} &
        \includegraphics[valign=c,width=\wratio\textwidth]{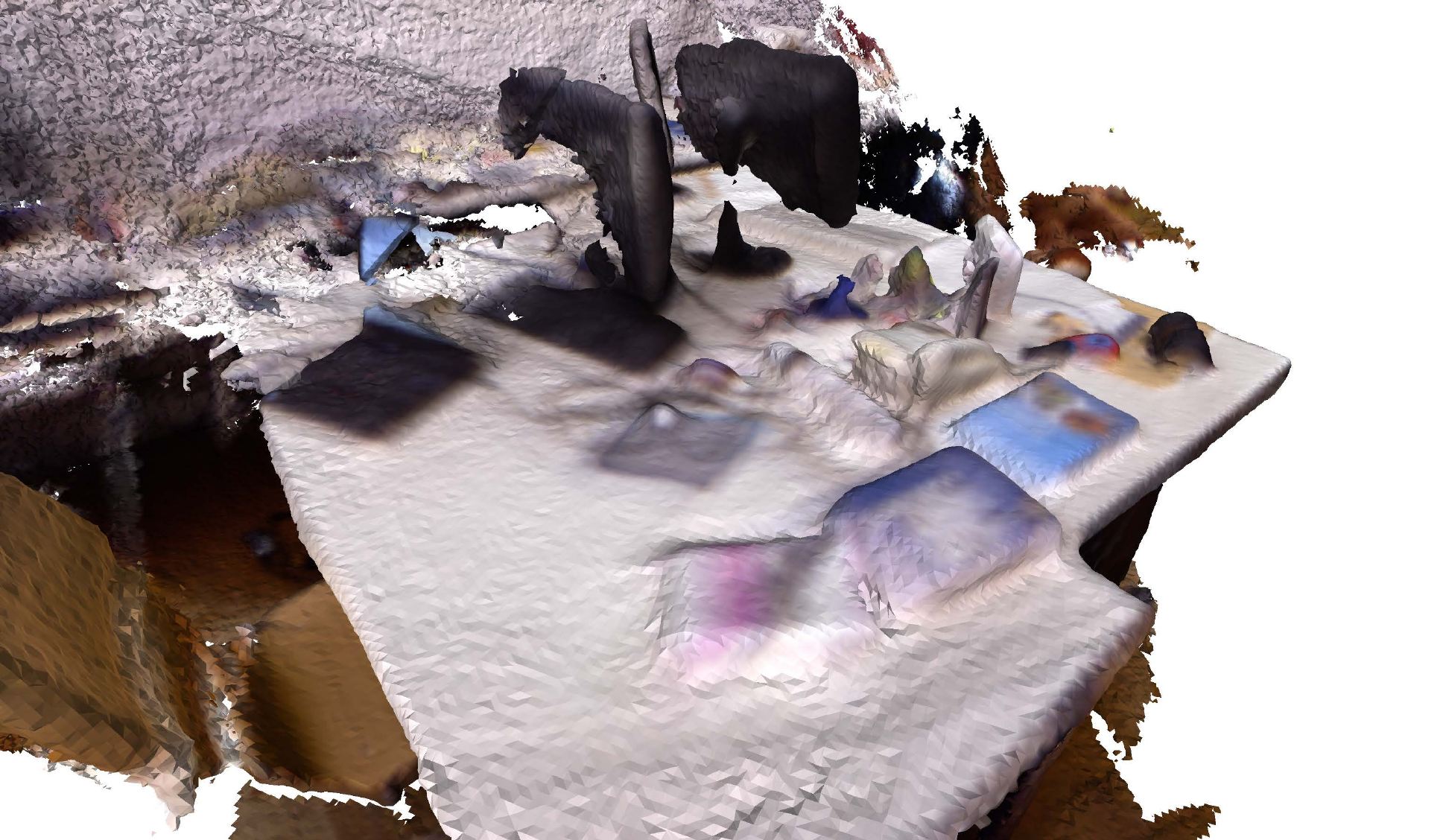} 
        \\[22pt]
        \rotatebox[origin=c]{90}{\texttt{office}} & 
        \includegraphics[valign=c,width=\wratio\textwidth]{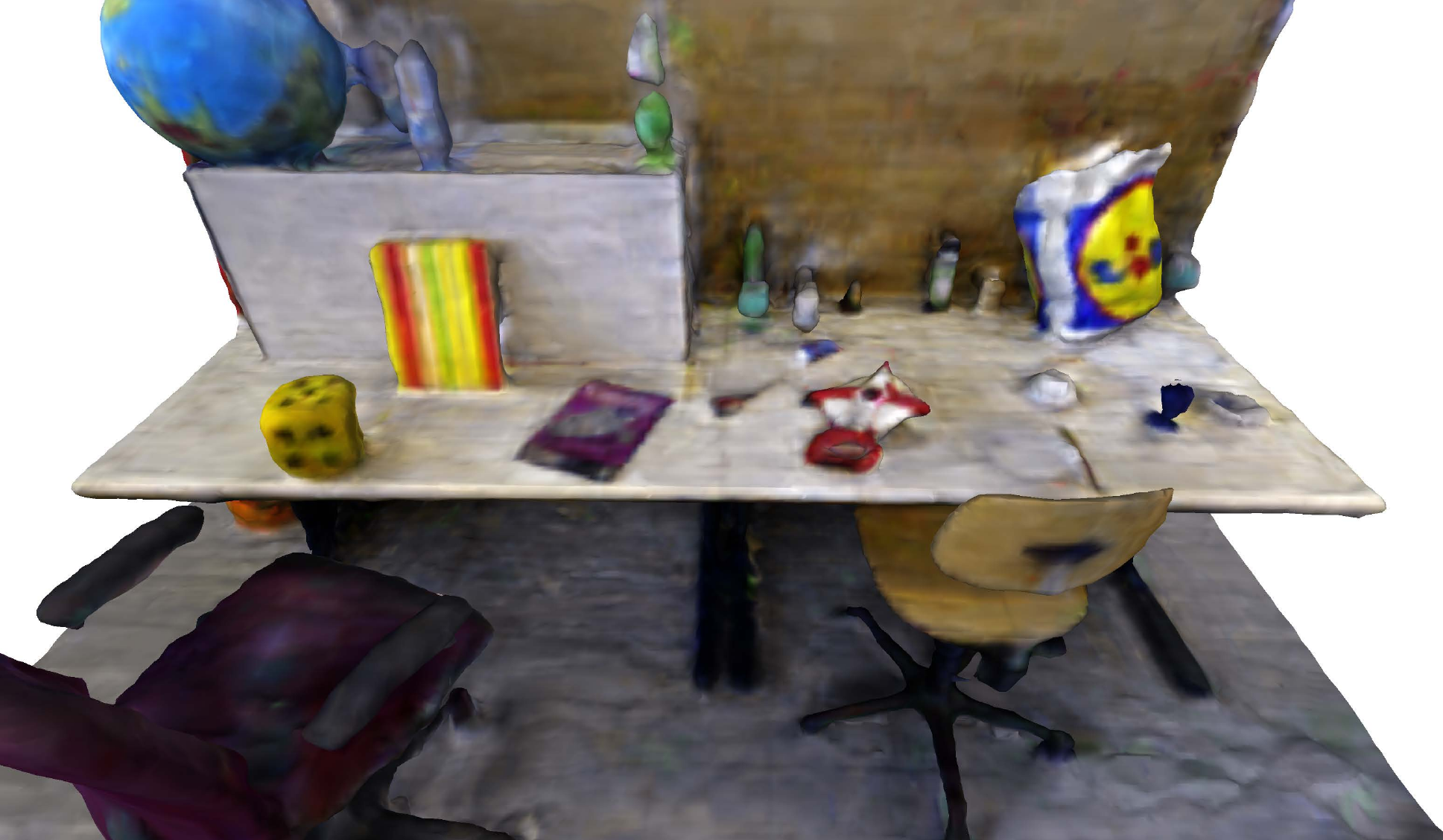} &
        \includegraphics[valign=c,width=\wratio\textwidth]{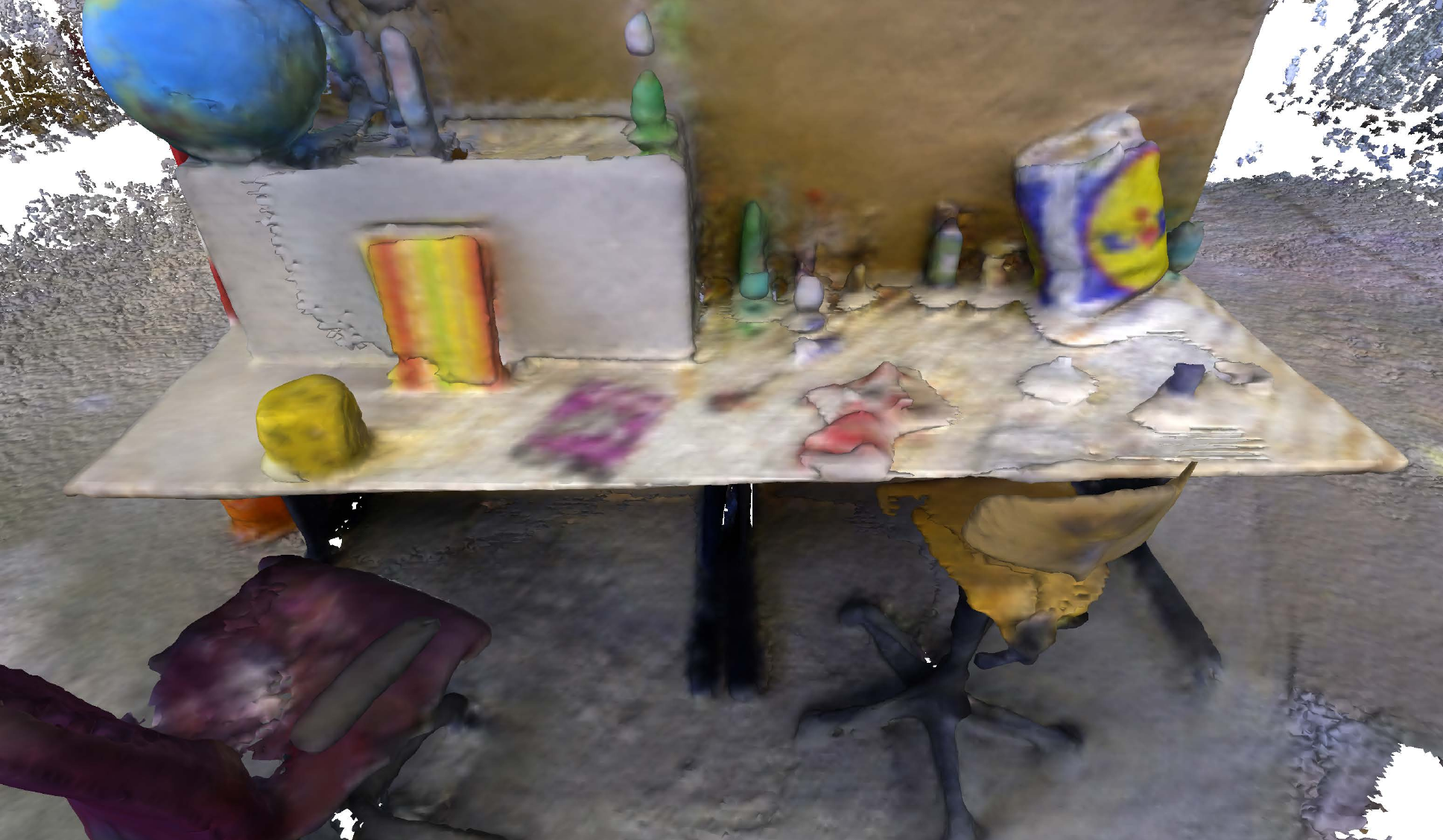} &
        \includegraphics[valign=c,width=\wratio\textwidth]{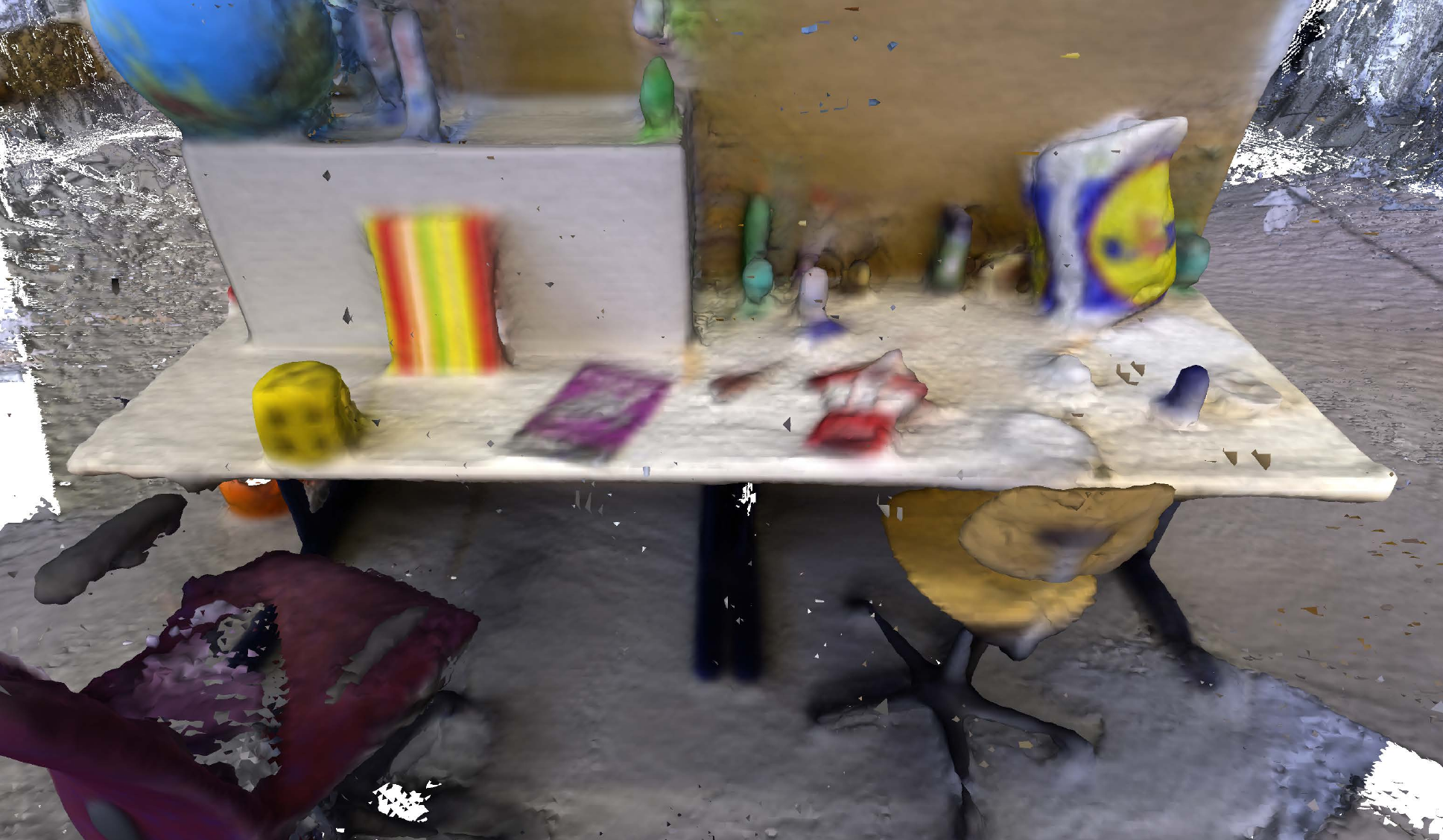} &
        \includegraphics[valign=c,width=\wratio\textwidth]{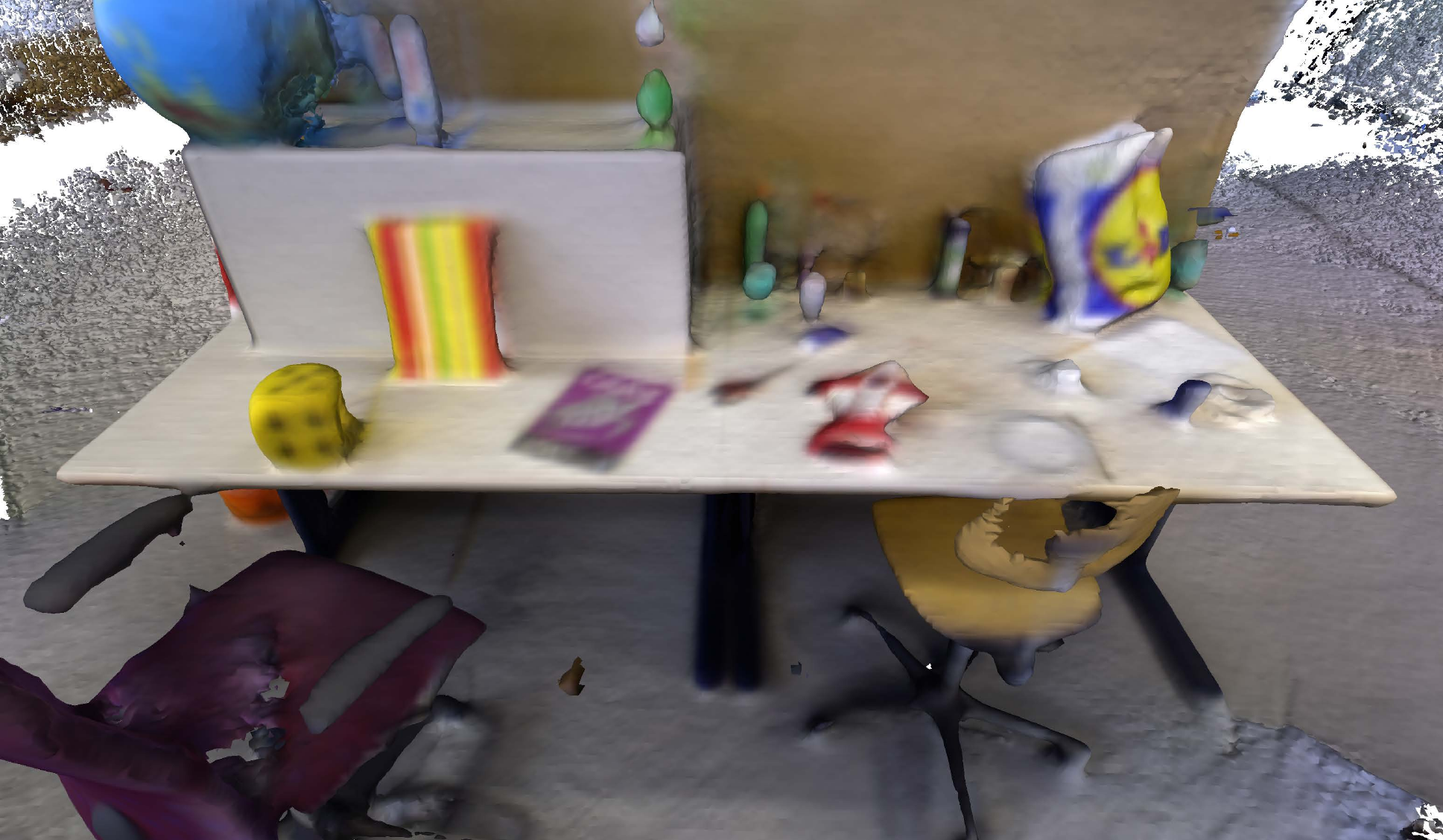}
        \\
        
    \end{tabular}

    \caption{\textbf{Qualitative Mesh-based Comparison on ScanNet\cite{Dai2017ScanNet} and TUM-RGBD\cite{stuhmer2010real} datasets.} For TUM-RGBD, the ground truth is obtained by TSDF fusion. NICE-SLAM\cite{zhu2022nice} shows over-smoothed surfaces. Point-SLAM\cite{sandstrom2023point} has duplicated geometry. ESLAM\cite{mahdi2022eslam} improves the reconstruction, while \ours is moderately better in recovering geometric details, see the chairs in \texttt{scene\_0059} for example.}
\label{fig:real_mesh}
\vspace{1cm}
\end{figure*}

\boldparagraph{Qualitative Reconstruction Results.}
\cref{fig:replica_mesh} shows reconstructed mesh on Replica dataset with a normal map shader to highlight the difference. \cref{fig:real_mesh} compares colored mesh on ScanNet~\cite{Dai2017ScanNet} and TUM-RGBD scenes. \ours can recover more geometric and color details in real-world reconstructions.
\label{par:supp_recon}

\boldparagraph{Novel View Synthesis.}
In \cref{tab:eval_nvs} we report the novel view synthesis results on the selected Scannet++~\cite{yeshwanth2023scannet++} scenes. The evaluated novel views in this dataset are not sampled from the input stream, but held-out views, which can better assess the extrapolation capability of the method. \ours demonstrates clear advantage and outperforms concurrent work~\cite{keetha2023splatam} by an average of 3.6 dB in PSNR. Qualitative rendering results are provided in \cref{fig:nvs_render}.
\label{par:supp_nvs}
\begin{table}[!htb]
\centering
\scriptsize
\renewcommand{\arraystretch}{1.5}

\begin{tabularx}{0.9\textwidth}{lYYYYYY}
\toprule
Method & \texttt{b20a261fdf} & \texttt{8b5caf3398} & \texttt{fb05e13ad1} & \texttt{2e74812d00} & \texttt{281bc17764} & Average \\
\midrule
ESLAM~\cite{mahdi2022eslam}& 13.63	&11.86	&11.83	&\nd10.59	&10.64	&11.71 \\
SplaTAM~\cite{keetha2023splatam} & \nd23.95 & \nd22.66 & \nd13.95 & 8.47 & \nd20.06 & \nd17.82 \\
\ours & \fs25.92 & \fs24.49 & \fs16.36 & \fs18.56 & \fs22.04 & \fs21.47 \\

\bottomrule
\end{tabularx}

\caption{\textbf{Novel View Synthesis Performance on ScanNet++ dataset~\cite{yeshwanth2023scannet++}} (PSNR $\uparrow$ [dB]). \ours demonstrates a clear advantage, outperforming concurrent work~\cite{keetha2023splatam} by an average of 3.6 dB in PSNR for held-out views. Our calculation includes all pixels, regardless of whether they have valid depth input.}
\label{tab:eval_nvs}
\end{table}

\begin{figure*}[!htb] \centering
    \newcommand{\wratio}{0.24}
    \setlength{\tabcolsep}{0.5pt}
    \renewcommand{\arraystretch}{1}
    \begin{tabular}{lllll}
    & \multicolumn{1}{c}{ESLAM~\cite{mahdi2022eslam}}
    & \multicolumn{1}{c}{SplaTAM~\cite{keetha2023splatam}}
    & \multicolumn{1}{c}{\ours}
    & \multicolumn{1}{c}{Ground Truth}
    \\
        \rotatebox[origin=c]{90}{\texttt{281bc17764}} & 
        \includegraphics[valign=c,width=\wratio\textwidth]{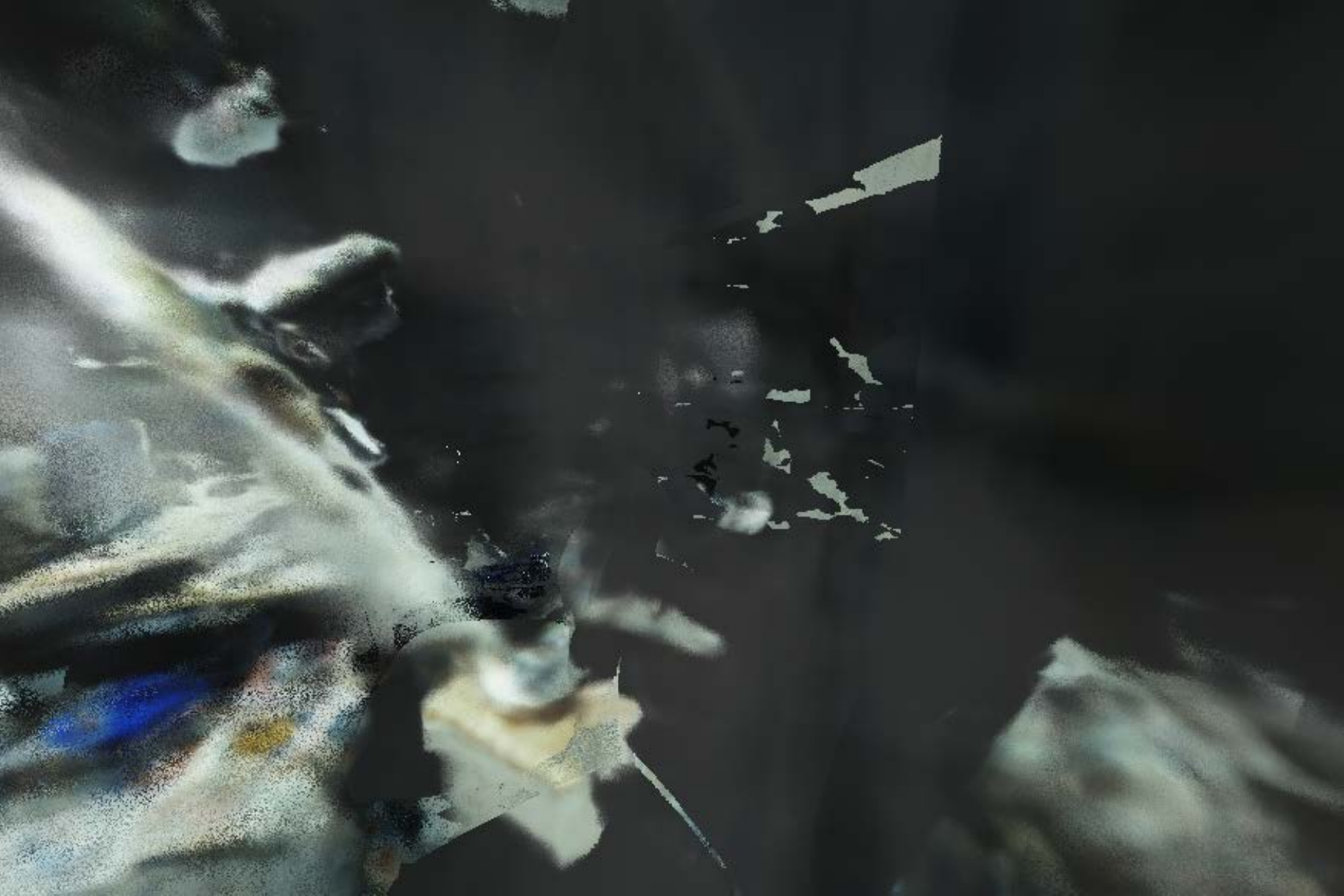} &
        \includegraphics[valign=c,width=\wratio\textwidth]{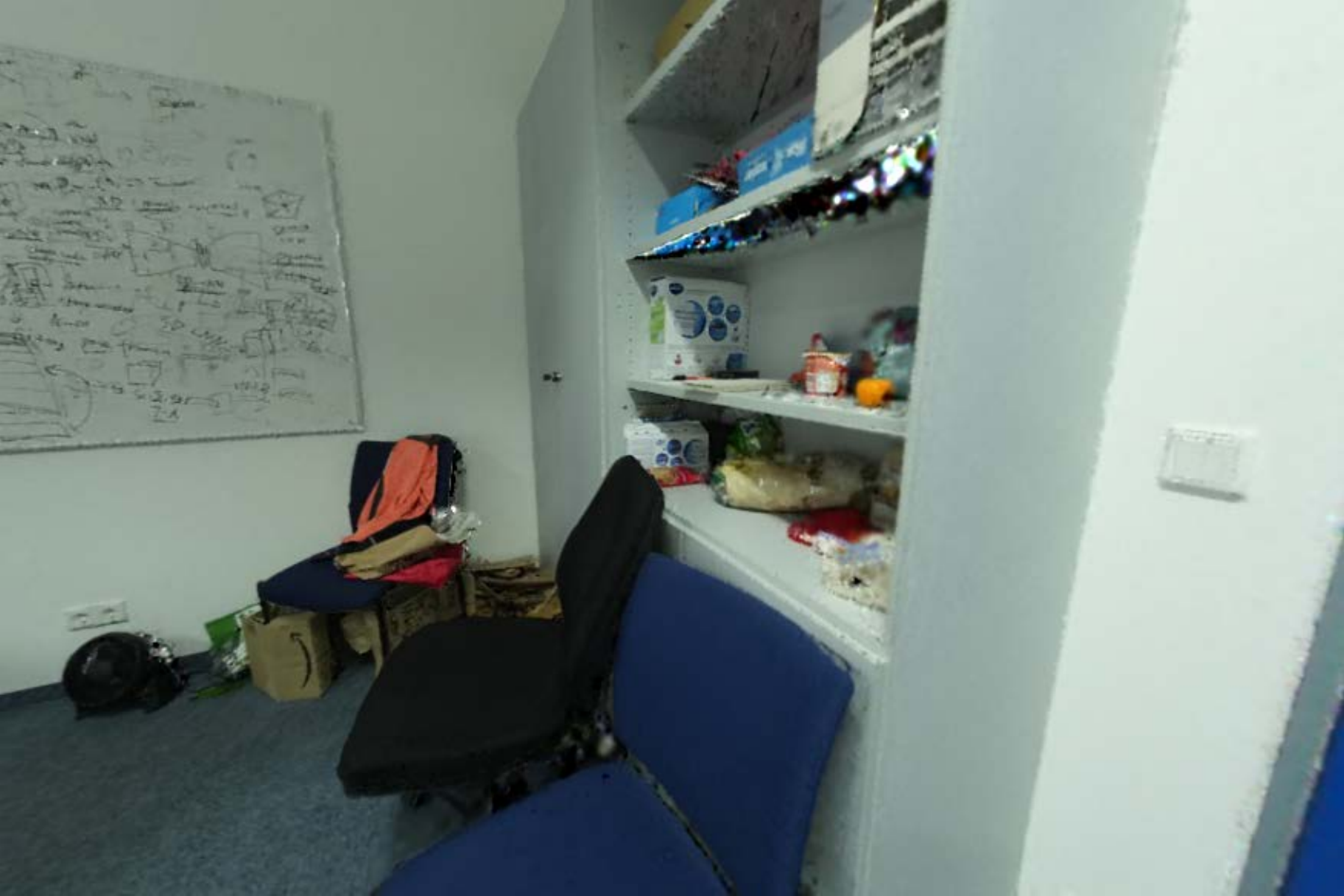} &
        \includegraphics[valign=c,width=\wratio\textwidth]{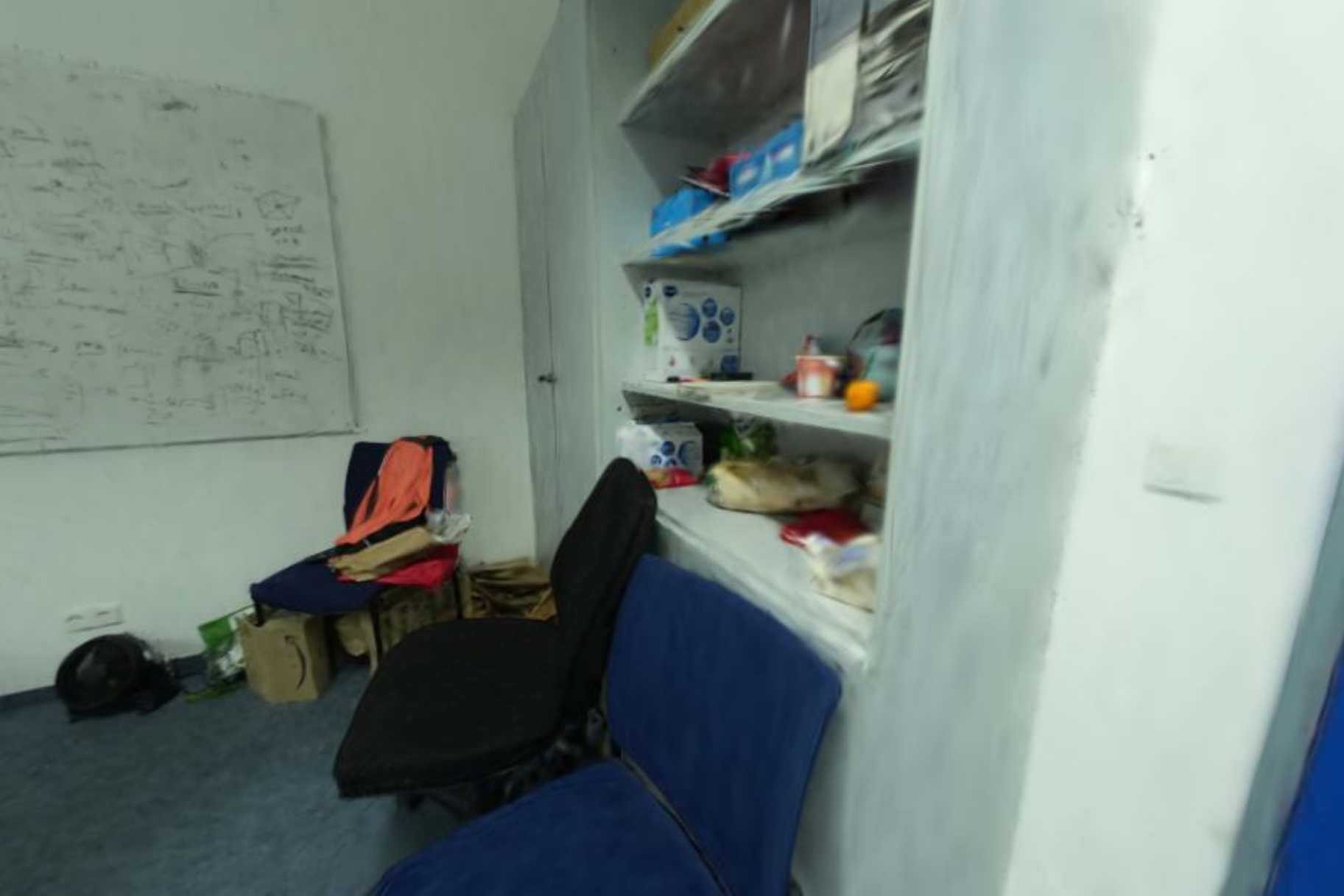} &
        \includegraphics[valign=c,width=\wratio\textwidth]{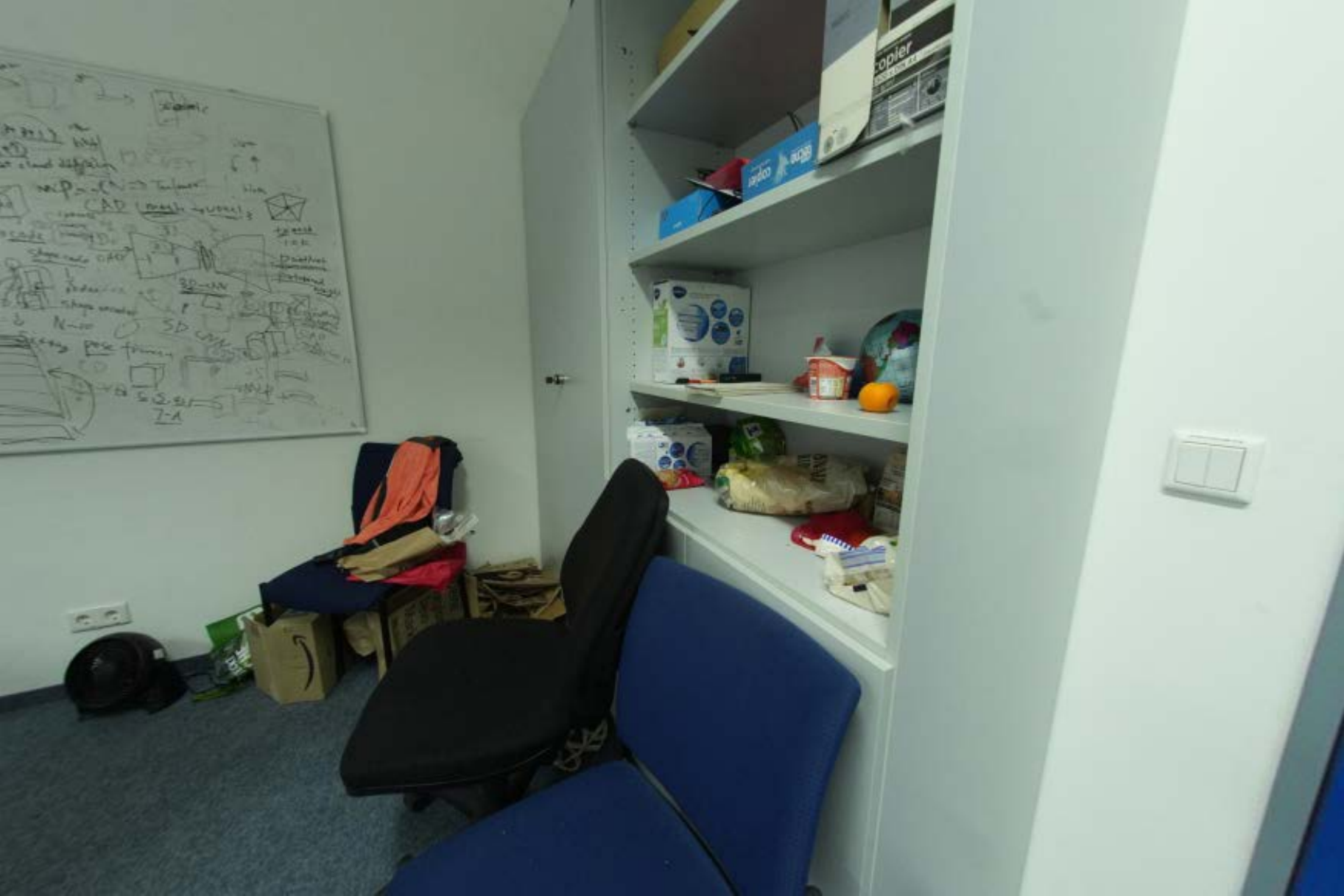} 
        \\[22pt]
        \rotatebox[origin=c]{90}{\texttt{8b5caf3398}} & 
        \includegraphics[valign=c,width=\wratio\textwidth]{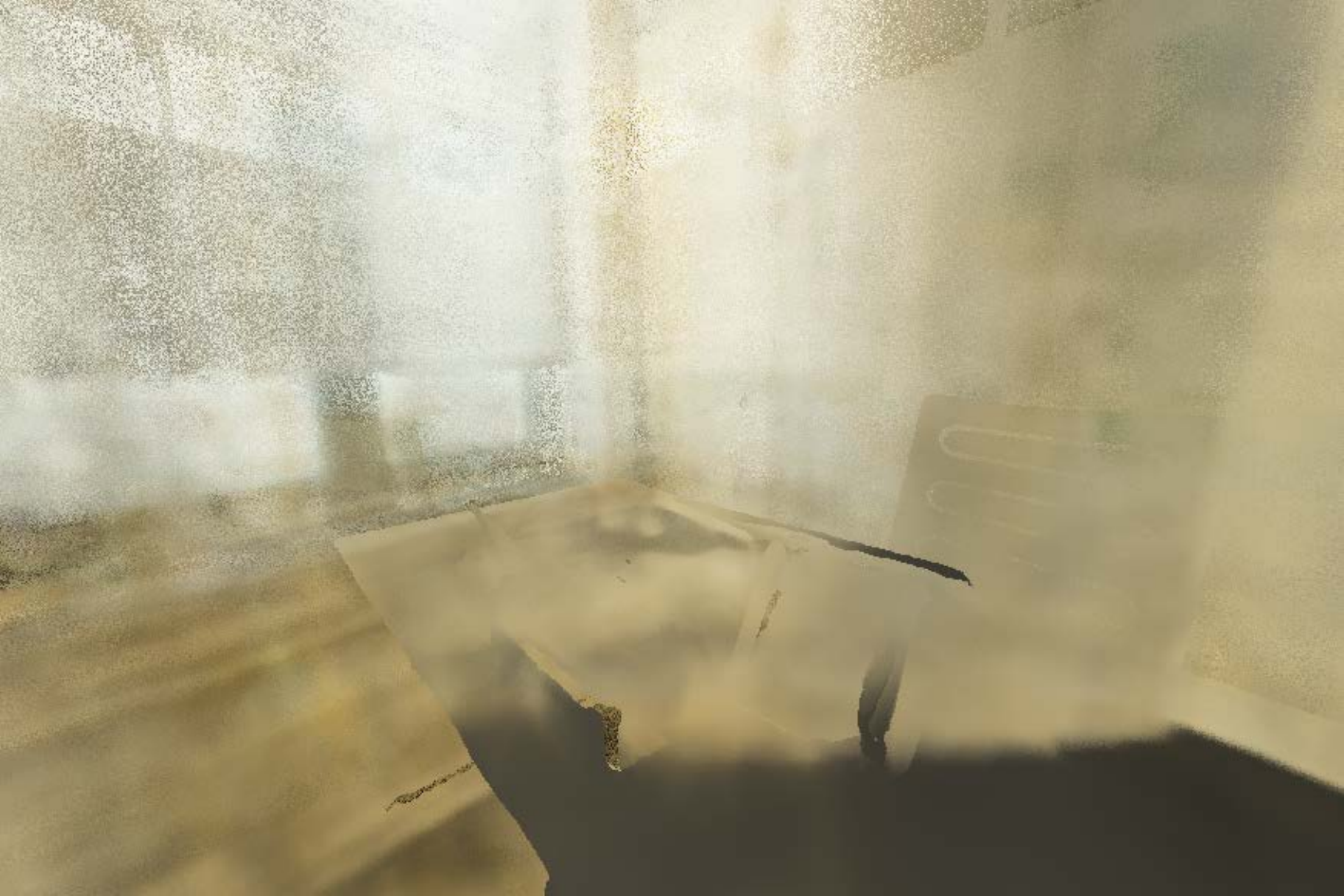} &
        \includegraphics[valign=c,width=\wratio\textwidth]{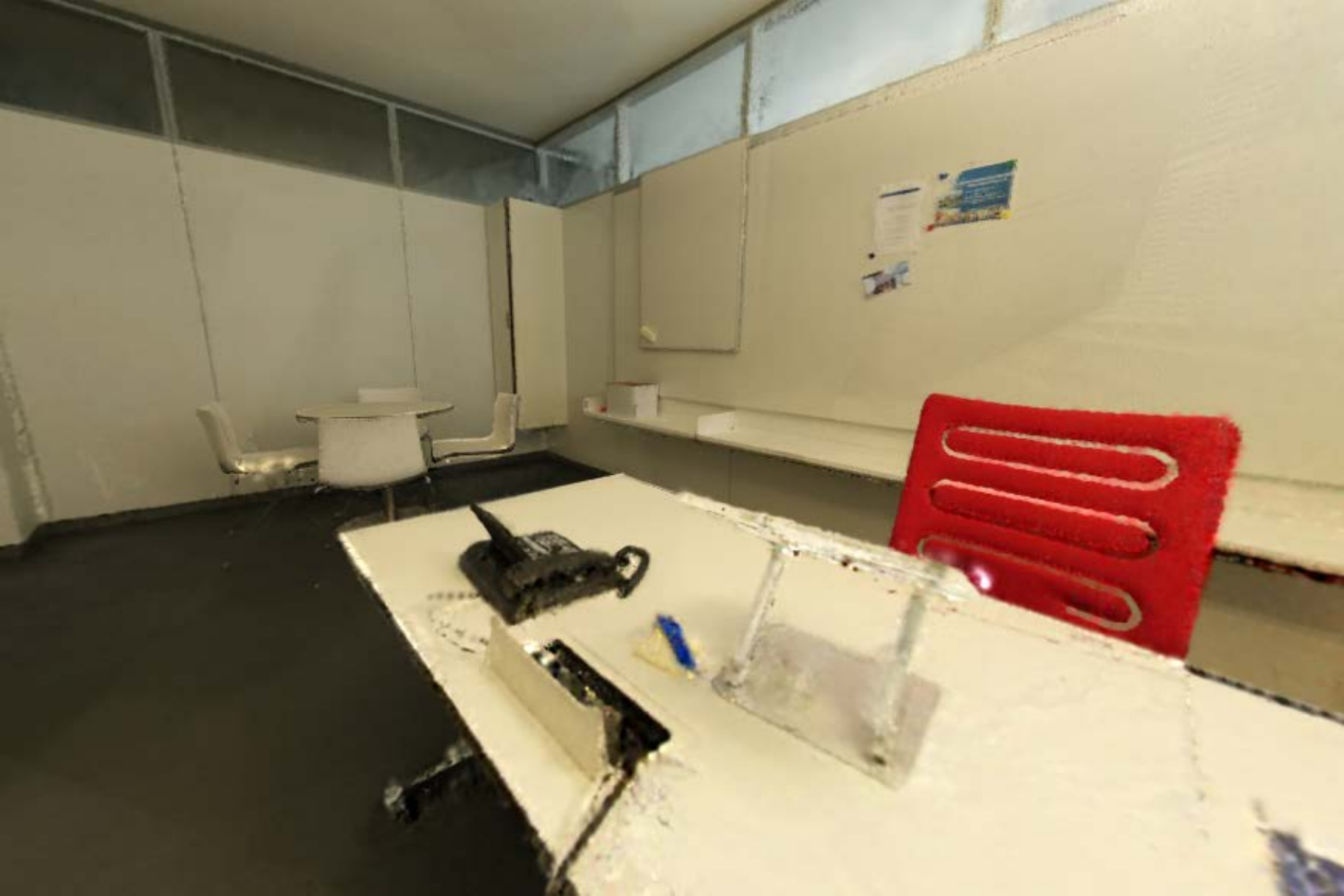} &
        \includegraphics[valign=c,width=\wratio\textwidth]{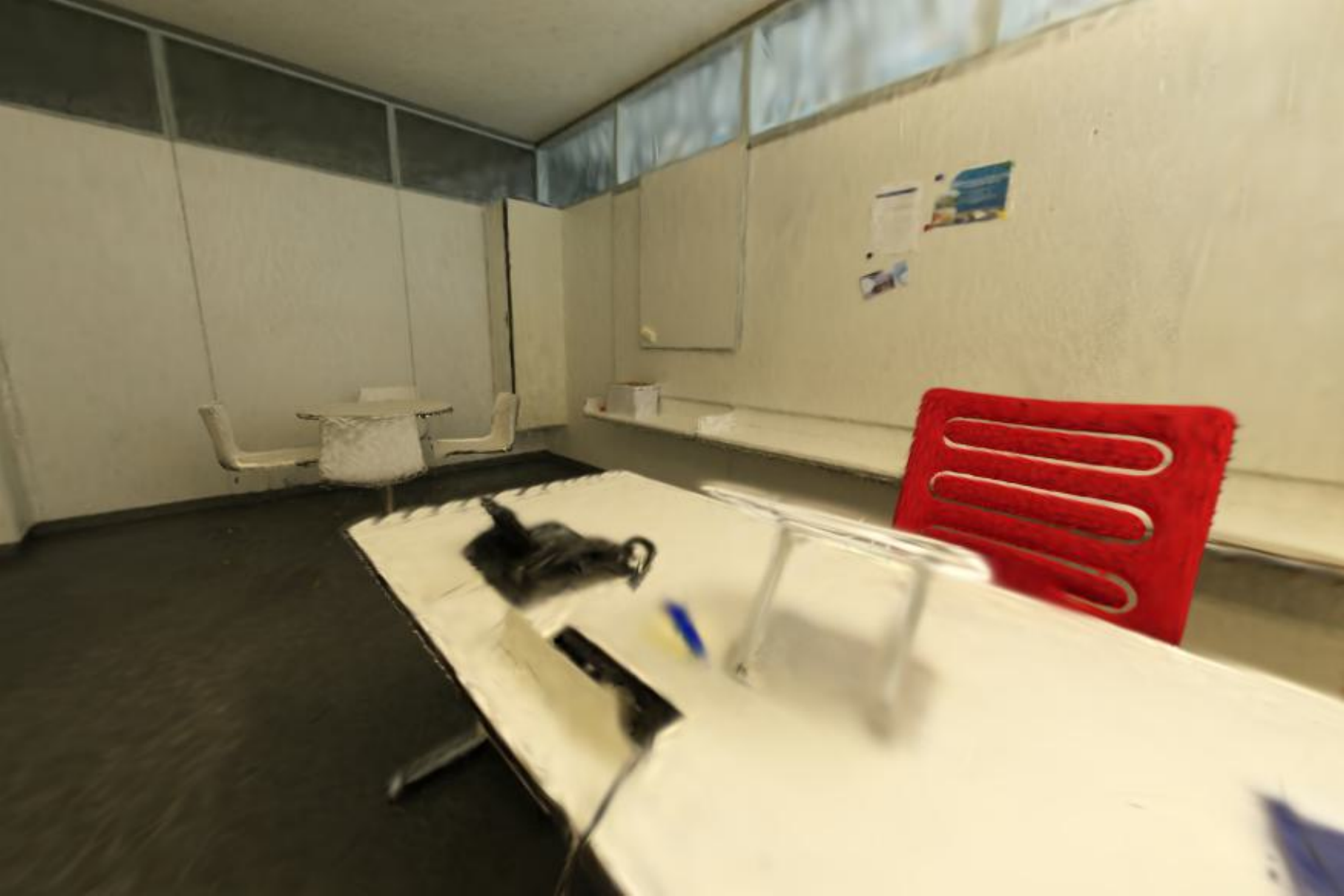} & \includegraphics[valign=c,width=\wratio\textwidth]{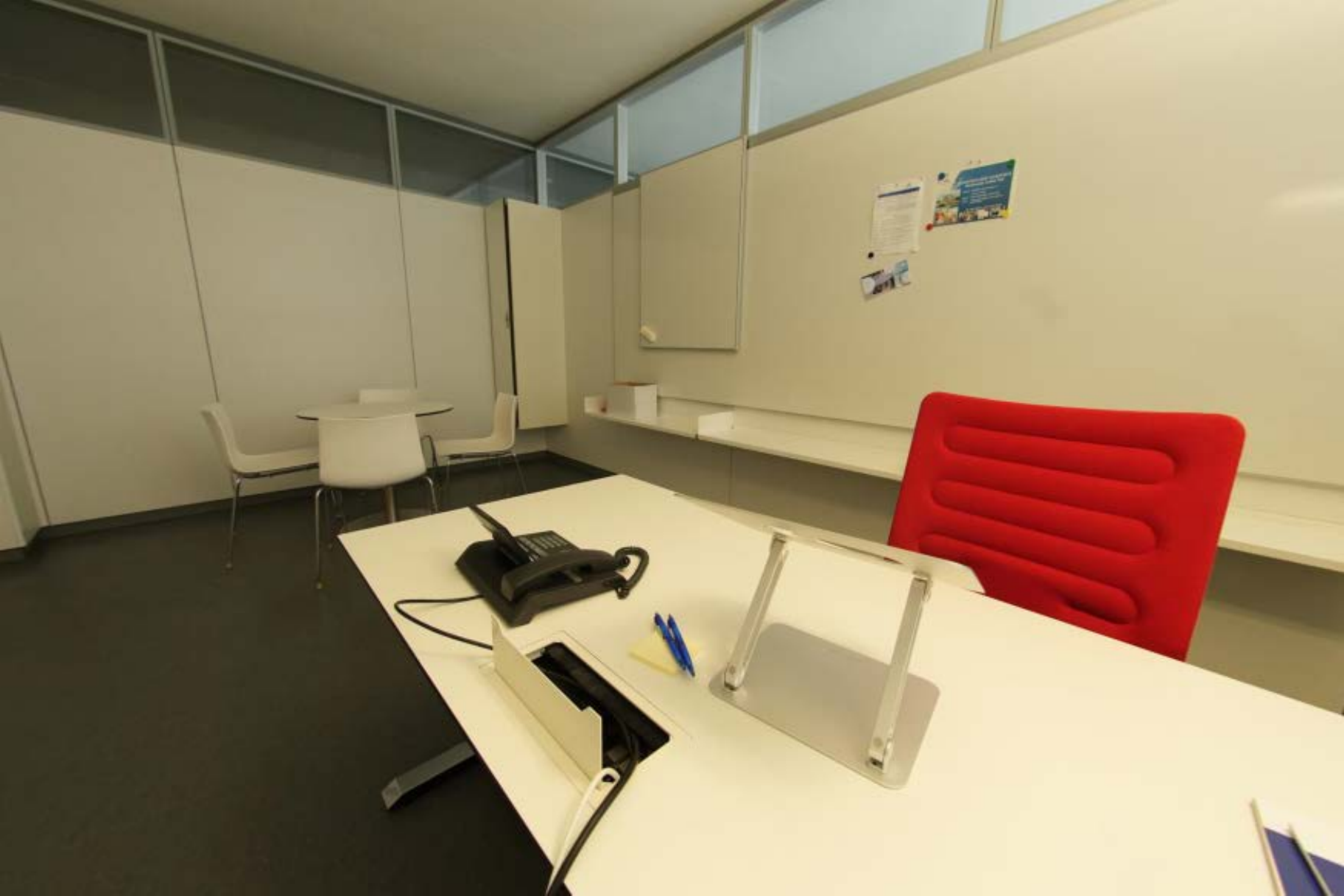} 
    \end{tabular}
    \caption{\textbf{Qualitative Novel View Synthesis Comparison on the ScanNet++ dataset~\cite{yeshwanth2023scannet++}}. \ours renders least artifacts at held-out views.}
\label{fig:nvs_render}
\end{figure*}

\bibliographystyle{splncs04}
\bibliography{main}

\end{document}